\documentclass{article}

\PassOptionsToPackage{numbers, compress}{natbib}

\usepackage[preprint]{neurips_2026}

\usepackage{subfigure}
\usepackage[utf8]{inputenc} 
\usepackage[T1]{fontenc}    
\usepackage{hyperref}       
\usepackage{url}            
\usepackage{booktabs}       
\usepackage{amsfonts}       
\usepackage{nicefrac}       
\usepackage{microtype}
\usepackage{xcolor}   

\usepackage[most]{tcolorbox}
\usepackage{needspace}
\usepackage{xcolor}
\usepackage{pifont}
\usepackage{fvextra}
\usepackage{inconsolata}
\usepackage{titlesec}

\titlespacing*{\paragraph}{0pt}{2pt}{4pt}
\newcommand{\CaseHeading}[2]{%
  \par\noindent{\bfseries #1\enspace\textit{#2}}\!
  \par\vspace{1pt}\nobreak
}

\newtcolorbox{casebox}[1]{
  enhanced,
  breakable,
  colback=gray!6,
  colframe=gray!75!black,
  coltitle=white,
  colbacktitle=gray!75!black,
  title=\textbf{#1},
  fonttitle=\bfseries\large,
  boxrule=0.8pt,
  titlerule=0pt,
  arc=1pt,
  outer arc=1pt,
  left=7pt,
  right=7pt,
  top=6pt,
  bottom=6pt,
  toptitle=4pt,
  bottomtitle=4pt,
  boxsep=0pt,
  before skip=6pt,
  after skip=8pt
}

\newenvironment{casecontent}
  {%
    \small
    \setlength{\parindent}{0pt}
    \setlength{\parskip}{1.5pt}
  }
  {}

\newcommand{\CaseField}[1]{%
  \par\vspace{2pt}\noindent\textbf{#1:}\enspace
}

\DefineVerbatimEnvironment{casecode}{Verbatim}{
  fontsize=\small,
  fontfamily=tt,
  breaklines=true,
  breakanywhere=true,
  commandchars=\\\{\},
  xleftmargin=0pt,
  frame=none
}

\usepackage[table]{xcolor}

\usepackage{amsmath}
\usepackage{graphicx}
\usepackage{wrapfig}
\usepackage{caption}
\usepackage{float}
\usepackage{multirow}
\usepackage{pifont}
\usepackage[normalem]{ulem}
\usepackage{url}
\usepackage{enumitem}

\usepackage{booktabs}
\usepackage{multirow}
\usepackage[table]{xcolor}
\usepackage{graphicx}

\title{MOSAIC: Modular Orchestration for Structured Agentic Intelligence and Composition}

\author{%
  Yifan Bao$^{1,}$\thanks{These authors contributed equally to this work.} \quad
  Xinyu Xi$^{1,}$\footnotemark[1] \quad
  Xinyu Liu$^{1,}$\footnotemark[1] \quad
  Wen Ge$^{2}$ \quad
  Lei Jiang$^{2,5}$ \quad
  Kevin Zhang$^{3}$ \quad \\
  \textbf{Raad Khraishi}$^{4}$ \quad
  \textbf{Yihao Ang}$^{1,}$\thanks{These authors are corresponding authors.} \quad
  \textbf{Anthony K. H. Tung}$^{1,}$\footnotemark[2] \quad
  \textbf{Lukasz Szpruch}$^{3,}$\footnotemark[2] \quad
  \textbf{Hao Ni}$^{2,}$\footnotemark[2] \quad \\
  $^1$Department of Computer Science, National University of Singapore \\
  $^2$University College London \\
  $^3$University of Edinburgh \\
  $^4$ NatWest AI Research\\
  $^5$Alan Turing Institute\\
  \texttt{\{yifan\_bao, yihao\_ang, atung\}@comp.nus.edu.sg}, \\
  \texttt{\{xinyu\_xi, e1519740\}@u.nus.edu}, \\
  \texttt{\{wen.ge.25, h.ni\}@ucl.ac.uk}, \\
  \texttt{ljiang@turing.ac.uk}, \texttt{K.Zhang-60@sms.ed.ac.uk}, \\
  \texttt{Raad.Khraishi@natwest.com}, 
  \texttt{L.Szpruch@ed.ac.uk}
}

\begin{document}

\maketitle

\begin{abstract}
\label{abs}

Automated data science is a structured model-selection problem. A solution must choose data transformations, feature representations, architecture, training procedure, evaluation protocol, and refinement strategy for a task. AutoML systems automate parts of this process, but typically search within predefined pipeline, model, and hyperparameter spaces. LLM-based agents offer greater flexibility through retrieval, code generation, and execution feedback, yet their modelling decisions are often unstructured, difficult to verify, and hard to reuse. We introduce \textsc{MOSAIC} (Modular Orchestration for Structured Agentic Intelligence and Composition), a structured agentic framework for memory-grounded model selection and workflow construction. Given a task and dataset, \textsc{MOSAIC} builds a semantic task profile, retrieves prior cases and source-code modules, and constructs a blueprint: an intermediate representation specifying selected modelling components, composition, interface constraints, and execution requirements. This blueprint turns model selection into a staged, context-grounded search and grounds LLM-based code generation in retrieved evidence rather than unconstrained synthesis. Candidate models are validated by execution and refined using diagnostic feedback, training traces, task metrics, and a failure-aware reinforcement learning policy. We instantiate \textsc{MOSAIC} on financial time-series forecasting and generation, where models must satisfy predictive accuracy, distributional fidelity, execution reliability, and downstream financial criteria such as risk and tail behaviour. Experiments against AutoML and agentic baselines show that \textsc{MOSAIC} improves task performance, execution success, and decision traceability, demonstrating the value of treating automated data science as structured, reusable, and execution-grounded model selection.

\end{abstract}

\section{Introduction}
\label{sec:intro}
Data science modelling~\citep{domingos} requires data understanding, preprocessing, feature construction, model design, implementation, evaluation, and iterative refinement. These decisions are tightly coupled: modelling assumptions depend on the data, implementations depend on architectural choices, and effective refinements often require coordinated changes across the training procedure, model structure, and evaluation protocol.

AutoML systems automate important parts of this process, but typically do so by searching over predefined pipelines, model classes, and hyperparameter spaces~\citep{feurer2022auto, thornton2013auto}. LLM-based agents offer a more flexible alternative. They can retrieve examples, generate code, run programs, and react to execution feedback~\citep{yao2022react, wang2023voyager, lv2024codeact}. However, this flexibility is often realised through free-form code generation. Existing data-science agents~\citep{guo2024ds, baek2024researchagent, tsagent} typically lack both a structured memory of prior modelling work and a persistent intermediate representation of the workflow being constructed. As a result, they often behave as adaptive code generators rather than agents that systematically reuse, compose, verify, and improve modelling knowledge.

The central idea of \textsc{MOSAIC} is to make modelling memory explicit and operational. Prior tasks provide task-level memory~\citep{tsagent, shinn2023reflexion}: which modelling strategies worked for related datasets, objectives, and constraints. Code repositories provide implementation-level memory: reusable modules for data processing, feature engineering, model construction, training, evaluation, and refinement~\citep{novikov2025alphaevolve, ellis2023dreamcoder}. \textsc{MOSAIC} extracts, annotates, and retrieves such modules for a new task, and composes them through a structured intermediate representation.

This intermediate representation is a \emph{blueprint}: a workflow-level object that sits between the task description and executable code. The blueprint specifies the required modelling capabilities, selected modules, composition order, dimensional compatibility constraints, and execution constraints. Code generation is therefore grounded in retrieved cases, extracted code modules, and an explicit workflow plan, rather than left to unconstrained LLM synthesis. Generated candidates are then executed, diagnosed, and refined using errors, logs, validation metrics, and training dynamics.

A second challenge is that model improvement is often a long-horizon process. Greedy edit-and-test loops can miss refinements whose benefits only appear after subsequent changes, such as introducing normalisation before changing the learning-rate schedule or modifying the model architecture before adjusting the loss. At the same time, learning a policy that edits code directly is difficult: the space of possible edits is extremely large, and useful feedback is only observed after the edited program is executed. \textsc{MOSAIC} therefore separates refinement decisions from code realisation. A learned policy selects structured high-level refinement actions, while a frozen LLM executor translates these actions into concrete code edits. The policy is trained offline from refinement trajectories and uses failure-aware trajectory branching, rollback, and invalid-action masking to convert execution failures and performance-degrading edits into reusable supervision.

We evaluate \textsc{MOSAIC} on financial time-series forecasting and generation. This is a demanding testbed because models must perform well under temporal validation, preserve statistical and dependence structure, and satisfy downstream financial metrics such as risk and tail behaviour. It therefore tests whether memory-grounded workflow construction improves not only task performance, but also execution reliability and decision traceability.
Our contributions are:
\begin{itemize}[nolistsep,leftmargin=10pt]

    \item We formulate automated data science as \emph{memory-grounded modular workflow construction}, where task solutions are built by retrieving, composing, executing, and refining reusable modelling capabilities.

    \item We introduce a \emph{blueprint-based, repository-grounded synthesis framework} that retrieves prior cases and source-code modules, extracts reusable components, composes them into executable workflow blueprints, and generates task-adapted implementations.

    \item We formulate model refinement as a \emph{failure-aware offline reinforcement learning} problem, where execution feedback, rollback, trajectory branching, and invalid-action masking are used to learn long-horizon refinement policies for LLM-mediated code editing.

    \item We evaluate \textsc{MOSAIC} on financial time-series forecasting and generation, showing improvements in predictive performance, distributional fidelity, execution reliability, and decision traceability relative to AutoML and agentic baselines.

\end{itemize}
\section{Related Work}
\label{sec-related}

\paragraph{AutoML and Neural Architecture Search.}
AutoML systems such as AutoGluon~\citep{agtimeseries}, Auto-sklearn~\citep{feurer2022auto}, and Optuna~\citep{akiba2019optuna} automate model selection and hyperparameter tuning, while neural architecture search methods explore predefined structural spaces~\citep{zoph2016neural, elsken2019neural, liu2018darts}. However, these approaches operate over fixed search spaces and do not reason over task descriptions, retrieve prior cases, or synthesize new architectures from reusable modules. \textsc{MOSAIC} differs by treating model construction as a repository-grounded composition problem: it retrieves existing code, extracts modules, constructs a blueprint, and synthesizes executable models beyond fixed model-bank selection.

\paragraph{LLM-Based Agentic Systems.}
ReAct-style agents~\citep{yao2022react} interleave reasoning and action for interactive decision-making, while DS-Agent~\citep{guo2024ds} and ResearchAgent~\citep{baek2024researchagent} extend this paradigm to data science workflows through case-based reasoning and iterative self-critique. Broader auto-research systems, including AI Scientist~\citep{lu2024ai} and Agent Laboratory~\citep{schmidgall2025agent}, automate larger parts of the research pipeline. Recent work also studies agentic component selection as optimisation over capability, cost, and compatibility~\citep{yuan2025automated}. However, existing agentic systems often lack task-specific formalisation of modelling modules, execution constraints, and reusable workflow structure. \textsc{MOSAIC} addresses this gap by imposing an explicit module-level workflow representation for financial time-series modelling.

\paragraph{LLM Code Generation and Program Refinement.}
Recent methods explore diverse strategies for optimising solutions in code space: AlphaEvolve~\citep{novikov2025alphaevolve} applies evolutionary search with LLM-guided mutations, AIDE~\citep{jiang2025aide} models generation as tree search with drafting, debugging, and improving operators, EffiLearner~\citep{huang2024effilearner} drives iterative refinement through runtime execution feedback, and LLM4EFFI~\citep{ye2025llm4effi} separates high-level algorithmic reasoning from code synthesis. These methods optimise code but do not explicitly decompose the modelling problem into reusable time-series modules, with blueprint-level architectural constraints. \textsc{MOSAIC} instead grounds synthesis in extracted module representations, architecture-family-aware composition plans, and execution-based validation.

\paragraph{Reinforcement Learning for Sequential Decision-Making.}

Reinforcement Learning (RL) has been combined with LLMs by directly optimising LLM policies~\citep{ouyang2022training}, using LLMs as high-level planners~\citep{wang2023voyager}, or using LLMs to design reward functions~\citep{ma2023eureka}. Our setting is different: a learned refinement policy selects structured high-level actions, while a frozen LLM executor translates these actions into concrete code edits. We build on offline RL, particularly Implicit Q-Learning~\citep{kostrikov2021offline}, as well as invalid-action masking~\citep{huang2020closer} and safe RL shielding~\citep{alshiekh2018safe}. \textsc{MOSAIC} adapts these ideas to LLM-mediated code refinement, where execution failures and performance-based rollbacks require construction of failure-aware trajectories and deployment-time action constraints.

\begin{figure}[t]
    \centering
    \includegraphics[width=\linewidth]
    {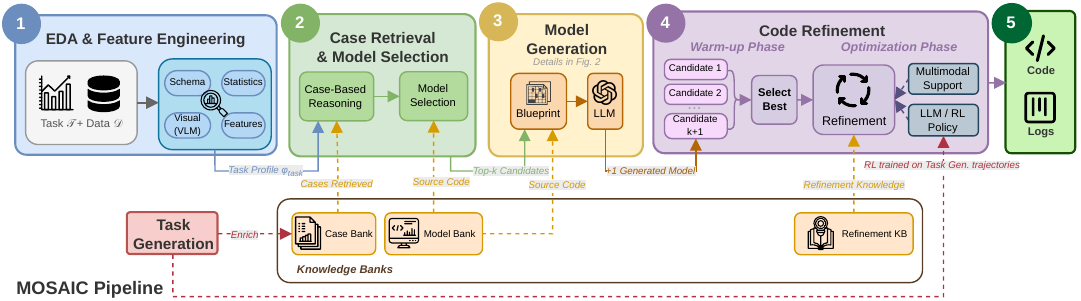}
    \vspace{-1.0em}
    \caption{Overview of the MOSAIC pipeline.}
    \label{fig:mosaic_pipeline}
    \vspace{-2.0em}
\end{figure}

\section{Overview of MOSAIC}
\label{sec:overview}

\subsection{Problem Formulation}
\label{subsec:problem}

The goal is to build a statistical model that is suitable for the task at hand. This requires to choose an appropriate model class, specify its architecture, adapt it using domain-specific modelling and optimisation heuristics, select a training procedure, tune hyperparameters, and impose suitable regularisation. These choices are interdependent: the admissible model class depends on the data and task, the training method depends on the architecture, and the evaluation protocol depends on the intended downstream use.

We therefore formulate automated data modelling as a workflow construction problem. Rather than searching directly over parameters of a single model family, \textsc{MOSAIC} searches over a structured space of executable modelling workflows. Each workflow specifies how data are processed, which features are constructed, which model class and architecture are selected, how the model is trained and regularised, how domain-specific refinements are applied, and how the final output is evaluated.

\paragraph{Task specification.}
A  modelling task is specified as
$    \mathcal{T} = (\mathcal{W}, \mathcal{D}, \mathcal{L}),$
where $\mathcal{W}$ is the natural-language task description, $\mathcal{D}$ is the dataset together with its train/validation/test split, and $\mathcal{L}$ is the evaluation protocol. The evaluation protocol may include predictive losses, distributional metrics, and downstream financial criteria such as risk, tail behaviour, or portfolio performance. (See details in \ref{app:taskgen} and examples of tasks in Appendix \ref{app:case_study})

\paragraph{Knowledge banks.}
\textsc{MOSAIC} assumes access to external knowledge resources
$
    \mathcal{K}
    =
    (
    \mathcal{E}_{\mathrm{case}},
    \mathcal{E}_{\mathrm{code}},
    \mathcal{E}_{\mathrm{refine}}
    )$,where $\mathcal{E}_{\mathrm{case}}$ is a case bank of prior task--solution pairs, $\mathcal{E}_{\mathrm{code}}$ is a repository of reusable model architectures, training routines, feature transformations, and evaluation components, and $\mathcal{E}_{\mathrm{refine}}$ is a refinement knowledge bank containing domain-specific optimisation heuristics. These knowledge banks define the reusable modelling capabilities from which new workflows can be constructed (see Appendix \ref{app:knowledge_banks}). 

\paragraph{Modelling objective.}
Given a task $\mathcal{T}$ and knowledge banks $\mathcal{K}$, the objective is to select and execute a modelling solution $m$ that performs well under the task-specific evaluation protocol. The solution $m$ includes the modelling choices needed to obtain a trained model, such as the model class, architecture, training method, hyperparameters, regularisation, and domain-specific optimisation heuristics.

Let $P_{\mathcal{T}}$ denote the task distribution, and let $\mathcal{U}(\mathcal{T},m)$ denote the loss obtained by executing modelling solution $m$ on task $\mathcal{T}$. The objective is
\begin{eqnarray}\label{eqn:obj_fun}
m^\star    \in    \arg\min_{m \in \mathcal{A}(\mathcal{T},\mathcal{K})}    \mathbb{E}_{\mathcal{T} \sim P_{\mathcal{T}}}    \left[        \mathcal{U}(\mathcal{T},m)    \right].
\end{eqnarray}
This formulation makes explicit that the search is not only over the parameters of a fixed model, but over admissible modelling solutions selected for the task at hand. \textsc{MOSAIC} operationalises this search by representing such solutions as structured executable workflows, introduced next.

\subsection{System Pipeline}
\label{subsec:system}

Figure~\ref{fig:mosaic_pipeline} illustrates \textsc{MOSAIC} as a staged search procedure for a modelling solution 
$m \in \mathcal{A}(\mathcal{T},\mathcal{K})$. The system does not ask an LLM to generate the final model directly. Instead, it progressively constructs the context needed to make model search well posed: first from the task and data, then from retrieved cases, then from reusable code modules, and finally from execution feedback.

Given a task $\mathcal{T}=(\mathcal{W},\mathcal{D},\mathcal{L})$, \textsc{MOSAIC} first performs \textbf{semantic-aware exploratory data analysis (EDA)} and feature engineering augmented by VLM-based multimodal support (details in Appendix~\ref{app:eda}).
This produces a task profile
\[
    \phi_{\mathrm{task}}
    =
    (
    \phi_{\mathrm{schema}},
    \phi_{\mathrm{stat}},
    \phi_{\mathrm{visual}},
    \phi_{\mathrm{feat}}
    ),
\]
\begin{wrapfigure}{r}{0.45\textwidth}
\centering
\vspace{-1em}
\includegraphics[width=0.45\textwidth]{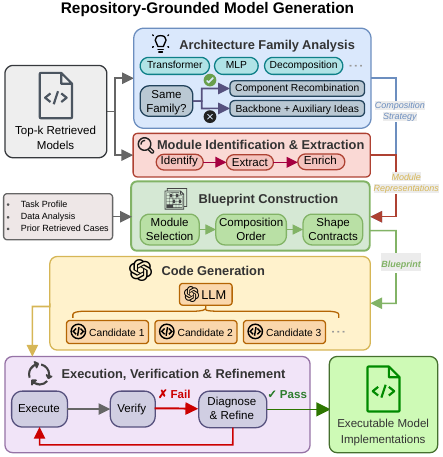}
\vspace{-1.5em}
\caption{Overview of the repository-grounded model generation pipeline.}
\label{fig:code_generation}
\vspace{-2em}
\end{wrapfigure} 
which summarises the dataset schema, statistical meta-features, visual time-series patterns, and engineered features. This profile acts as  
the first layer of context for model search. 
It is used to query the case bank $\mathcal{E}_{\mathrm{case}}$ and retrieve prior tasks with similar data characteristics, objectives, and evaluation requirements.
The retrieved cases, together with the model/code bank $\mathcal{E}_{\mathrm{code}}$, are then used to select the top-$k$ candidate models, denoted by
$m^{(k)} = (m_1,\ldots,m_k)$.

\textbf{Repository-grounded model generation}. This stage expands beyond direct model selection. As shown in Figure~\ref{fig:code_generation}, the repository-grounded model generation module uses the top-$k$ candidates as context for constructing a new candidate model. It first analyses the architectural families represented among the retrieved candidates, then extracts reusable modules from the corresponding source code, \emph{e.g.} embeddings, encoders, and forecasting heads. 
These modules are enriched with semantic annotations and shape information, allowing them to be composed subject to dimensional and functional compatibility constraints. Further details are provided in Appendix~\ref{app:codegen}.

From this information, \textsc{MOSAIC} constructs a blueprint $B$: an intermediate representation that specifies the relevant modelling components, their composition order, their input--output interfaces, and the execution constraints required for a valid implementation. The LLM is then used as an executor of this structured context rather than as an unconstrained code generator. Conditioned on the retrieved cases, extracted modules, and blueprint, it synthesises a new candidate model $m_{k+1}$. This generated model is added to the original top-$k$ candidates ($m^{(k)}=(m_1,\ldots,m_k)$), producing the candidate pool $
    m^{(k+1)} = (m^{(k)},m_{k+1}).$

This staged construction can be viewed as a factorisation of the search for the final modelling solution. Informally,
\[
\begin{aligned}
p(m \mid \mathcal{T},\mathcal{K})
&\approx
p(m \mid m^{(k+1)}, h, \mathcal{E}_{\mathrm{refine}})
\,p(m_{k+1} \mid B, m^{(k)}, \mathcal{E}_{\mathrm{code}})
\\
&\quad \times
p(B \mid \phi_{\mathrm{task}}, m^{(k)}, \mathcal{E}_{\mathrm{code}})
\,p(m^{(k)} \mid \phi_{\mathrm{task}}, \mathcal{E}_{\mathrm{case}}, \mathcal{E}_{\mathrm{code}})
\,p(\phi_{\mathrm{task}} \mid \mathcal{T}),
\end{aligned}
\]
where $h$ denotes the execution history, including logs, errors, validation metrics, training traces, and loss-curve visualisations. This factorisation should be read as a procedural description rather than as a calibrated probabilistic model: each stage conditions the next one, narrows the search space, and grounds LLM generation in task-specific evidence.

Finally, the $k+1$ candidates enter a two-phase refinement stage. In the warm-up phase, all candidates are executed and evaluated in parallel. The best-performing candidate is then refined in an optimisation phase using the refinement knowledge bank $\mathcal{E}_{\mathrm{refine}}$ and execution feedback. Refinement can be driven by the learned RL policy with LLM executor described in Section~\ref{sec:rl}. The resulting output is an executable modelling solution selected through retrieval, blueprint construction, LLM-based realisation, execution, and refinement.
\subsection{RL-Guided Long-Horizon Refinement}
\label{sec:rl}

After repository-grounded generation and execution verification, \textsc{MOSAIC} obtains a pool of executable candidate models. The refinement stage aims to improve the selected candidate through a sequence of structured edits. This is naturally a sequential decision problem: a locally neutral or even slightly degrading edit, such as adding normalisation, changing the learning-rate schedule, or modifying regularisation, may enable subsequent improvements. Conversely, an apparently plausible edit may lead to an unstable or poorly performing model and should not be repeatedly explored from the same state.

We formulate refinement as a finite-horizon decision process over executable models. At refinement step $t$, the state is
$s_t = (m_t,\mathcal{M}_t)$, where $m_t$ is the current executable model, including its code, architecture, training configuration, and hyperparameters, and $\mathcal{M}_t$ denotes the accompanying model diagnostics. These diagnostics include, validation loss induced by the evaluation $\mathcal{L}$, execution logs and any failure information observed so far. Thus, the state records not only the current model, but also the evidence needed to decide how it should be refined.

The action space consists of structured refinement strategies rather than raw code edits. An action $a_t \in \mathcal{A}_{\mathrm{ref}}$ may specify, for example, to add normalisation, modify dropout, change the learning-rate schedule, tune batch size, adjust regularisation, alter model capacity, or replace a modelling component. The action is therefore a high-level modelling decision. A frozen LLM executor then translates this structured action into concrete code changes, which are executed to obtain the next model and diagnostics.

This policy--executor interface can be written as the following factorisation:
\[
\begin{aligned}
p(s_{t+1}\mid s_t)
&=
\sum_{a_t,e_t}
p(s_{t+1}\mid e_t,s_t)
p(e_t\mid a_t,s_t)
\pi(a_t\mid s_t),
\end{aligned}
\]
where $p(a_t\mid s_t)$ is the refinement policy, $p(e_t\mid a_t,s_t)$ is the LLM executor that realises the selected strategy as a code edit $e_t$, and $p(s_{t+1}\mid e_t,s_t)$ is the execution environment obtained by running the edited program. In this way, the RL policy learns to choose refinement strategies, while the LLM remains responsible for implementing them.

The goal is to minimise the task-specific loss over a refinement horizon. Let $\mathcal{L}(s_t) := \mathcal{L}(\mathcal{T},m_t)
$
denote the loss obtained by executing the model $m_t$ on task $\mathcal{T}$. For a refinement policy $\pi$, we seek
\[
    \pi^\star
    \in
    \arg\min_{\pi}
    \mathbb{E}_{\pi}
    \left[
        \sum_{t=0}^{H-1}
        \gamma^t
        \left(\mathcal{L}(s_{t+1}) - \mathcal{L}(s_{t}) \right)
    \right],
\]
subject to the constraint that each visited model remains executable and satisfies the required interface and numerical-validity checks. Here the reward $r_t = - \left(  \mathcal{L}(s_{t+1}) - \mathcal{L}(s_{t}) \right)$ and $\gamma$ is the discount factor.

\paragraph{Offline trajectory construction.}
The refinement policy is trained offline. For each generated training task, \textsc{MOSAIC} instantiates an executable modelling program and explores alternative refinement paths using an LLM-based refinement agent. At each step, the agent proposes a structured refinement action, the LLM realises it as a concrete code edit, the edited program is executed, and the resulting loss and diagnostics are recorded. This produces trajectories of the form $(s_0,a_0, r_0, s_1,a_1,\ldots,s_H, r_H),$
where each state contains both the current model and its diagnostics.

A key difficulty is that LLM-mediated refinement produces many failed or degrading branches. \textsc{MOSAIC} therefore uses execution guardrails. A \emph{hard failure} occurs when the edit cannot be executed, for example because of a syntax error, shape mismatch, missing dependency, NaN loss, or timeout. In this case, the transition is stored as a terminal failed transition and the system rolls back to the last valid executable checkpoint. A \emph{soft failure} occurs when the edit is executable but causes sustained degradation in the validation loss. In this case, the system rolls back to the previous best checkpoint rather than continuing from the degraded model.

These rollbacks create a tree-structured refinement process rather than a single linear trajectory. To preserve valid learning trajectories, \textsc{MOSAIC} uses trajectory branching, as illustrated in  Figure \ref{fig:rl} and detailed in Appendix~\ref{app:rl}. 
Suppose the refinement path reaches a good checkpoint $s^\star$, then explores a sequence of edits that leads to degraded states and triggers a rollback. Instead of storing the raw path with a discontinuous transition back to $s^\star$, the rollout is split into two branches:
(1)
    $\text{failure branch: } s^\star \rightarrow \cdots \rightarrow s_{\mathrm{bad}}$, and
(2) $\text{continuation branch: } s^\star \rightarrow \cdots$.

The failure branch is retained as negative supervision, while the continuation branch records the subsequent valid search from the restored checkpoint. The action that initiated the failed branch is stored in a state-specific invalid-action mask $ \mathcal{I}_{\mathrm{invalid}}(s^\star)$.
Thus, the system does not merely discard bad refinements: it learns which actions should not be repeated from the same diagnostic state.

\paragraph{Offline learning and deployment.}
The branched trajectories define an offline dataset of states, structured refinement actions, losses, next states, terminal indicators, and invalid-action masks. We train a refinement policy on this dataset using offline RL, instantiated in our experiments with Implicit Q-Learning \citep{kostrikov2021offline}. The policy is trained to select actions that reduce long-horizon loss while avoiding actions that previously caused hard failures or soft reverts. During training, actions in $\mathcal{I}_{\mathrm{invalid}}(s)$ are masked for the corresponding state.

At deployment, the learned policy is frozen. For a new task, \textsc{MOSAIC} observes the current state $s_t=(m_t,\mathcal{M}_t)$ and selects the best available refinement action under a dynamic valid-action mask:
\[
    a_t
    \in
    \arg\min_{a \in \mathcal{A}_{\mathrm{valid}}(s_t)}
    Q(s_t,a),
\]
where $\mathcal{A}_{\mathrm{valid}}(s_t)$ excludes actions already known to fail from the current state. The selected action is realised by the LLM executor, the resulting code is executed, and the diagnostics are updated. If the edit triggers a hard failure or soft revert, \textsc{MOSAIC} restores the last valid checkpoint, adds the failed action to the dynamic mask, and queries the policy again. The RL module therefore provides long-horizon strategy selection, while the selected strategy is executed by the LLM for flexibility. 
\section{Experiments}
\label{sec:exp}
\subsection{Experimental Setup}

\paragraph{Datasets and Tasks.}
We evaluate \textsc{MOSAIC} on financial time-series forecasting and generation. Forecasting targets accurate prediction of future prices, while generation aims to synthesise realistic time-series that preserve key statistical and structural properties of the original data. Both tasks are further assessed through downstream financial metrics. 
We adopt three widely-used financial datasets \citep{ctbench, tsagent}: (1) Hourly closing prices for a set of 20 cryptocurrency trading pairs against USDT (Tether) in 2024 (\textbf{Crypto}); 
(2) High-frequency limit order book data sourced from the LOBSTER platform, covering NASDAQ-listed equities with millisecond-resolution event streams including order submissions, cancellations, and executions (\textbf{LOB}); and
(3) Daily closing prices for 10 major U.S. stocks from January 2020 to December 2024, covering business days only (\textbf{Stock}); details in Appendix~\ref{app:datasets}.

\paragraph{Baselines.}
At the system level, we benchmark against DS-Agent~\citep{guo2024ds}, ResearchAgent~\citep{baek2024researchagent},TS-Agent~\citep{tsagent} AutoGluon~\citep{agtimeseries} (forecasting), and Optuna~\citep{akiba2019optuna} (generation). 
At the module level, we compare Repository-Grounded Model Generation against AlphaEvolve~\citep{novikov2025alphaevolve}, AIDE~\citep{jiang2025aide}, EffiLearner~\citep{huang2024effilearner}, and LLM4EFFI~\citep{ye2025llm4effi}, and RL for Refinement against BC~\citep{pomerleau1988alvinn}, DQN~\citep{mnih2015human}, and an LLM-only policy, each integrated as a drop-in replacement. 
All agents use GPT-5.4~\citep{openai2025gpt5}, GPT-4o~\citep{openai2023gpt4}, Claude Opus 4~\citep{anthropic2025claude}, and Nova Pro~\citep{aws2025nova}.

\paragraph{Evaluation Protocol.}
We follow prior work~\citep{tsagent} and report standard forecasting metrics, including RMSE, MAE, MAPE, and sMAPE, as well as generation metrics, including Marginal, Correlation, Autocorrelation, and Covariance distances. We also report $\Delta$VaR and $\Delta$ES on Crypto, $\Delta$Sharpe for forecasting, and system-level Success Rate.
For model generation, we report Win Rate, the proportion of generated models selected as the final system output after joint refinement with retrieved candidates. For RL, we report Steps to Best Incumbent (abbreviated as Steps), measuring how many refinement steps are needed to reach the best performing candidate. We further ablate EDA, Feature Engineering, and Multimodal Support using the same metrics. All reported metrics are averaged over five independent runs to ensure statistical reliability.


\begin{figure}[H]
  \vspace{-1.0em}
  \centering
  \includegraphics[width=0.60\textwidth]{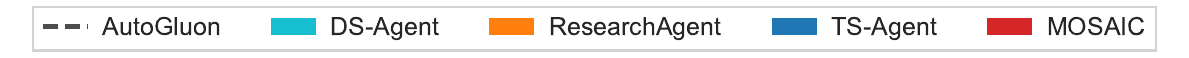} \\
  \includegraphics[width=\textwidth]{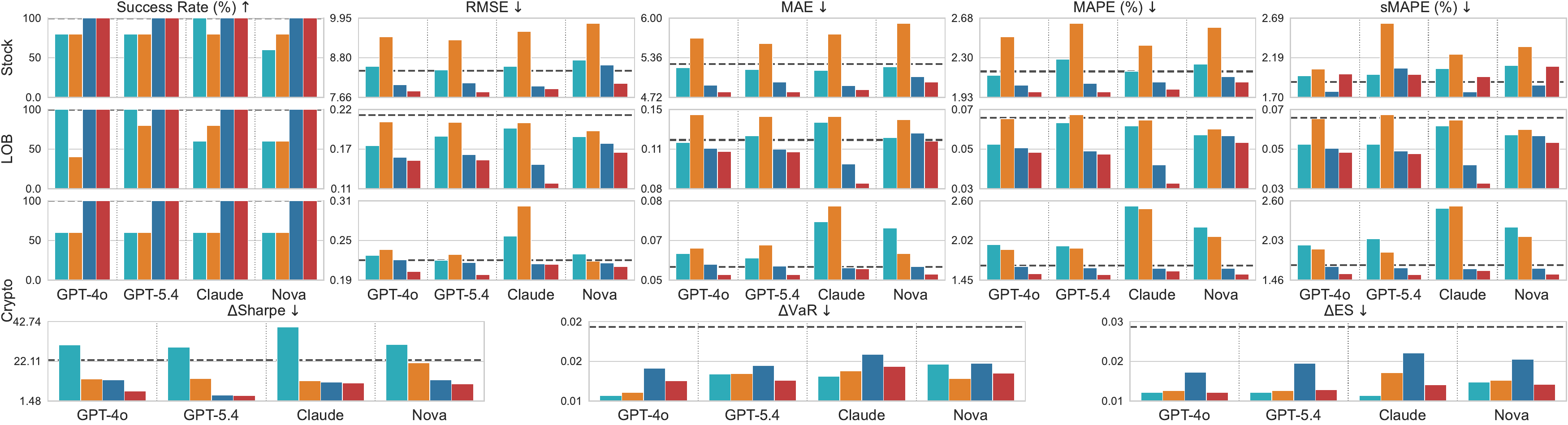} \\
  \vspace{-0.5em}
  \caption{Forecasting performance across datasets, metrics, and LLM backbones. Each metric is averaged over five runs.}%
  \label{fig:tsf_metrics}%
  \vspace{-1.5em}
\end{figure}

\paragraph{Module and Ablation Definitions.}
For clarity, we denote system variants as follows: \textbf{ModelGen (MOSAIC)} performs repository-grounded model generation without RL, using the LLM for refinement if candidates are valid; \textbf{IQL (MOSAIC)} applies RL-guided refinement independently of ModelGen; \textbf{EDA (MOSAIC)} removes EDA, Feature Engineering, and Multimodal Support; and \textbf{Full MOSAIC} runs the complete system, with refinement guided by RL when ModelGen does not yield a winning candidate, and otherwise by the LLM. More implementation details are in Appendix~\ref{app:implementation}.

\subsection{Overall Evaluation}
\paragraph{Time Series Forecasting.}
\begin{wrapfigure}{r}{0.52\textwidth}
\centering
\vspace{-1em}
\subfigure[Forecasting Ranking]{%
    \label{fig:tsf_radar}%
    \parbox[b]{0.48\linewidth}{%
        \centering
        \includegraphics[width=\linewidth, height=0.5cm, keepaspectratio=true]{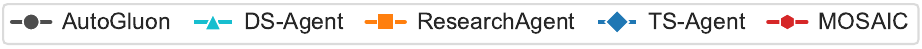} \\[2pt]
        \includegraphics[width=\linewidth]{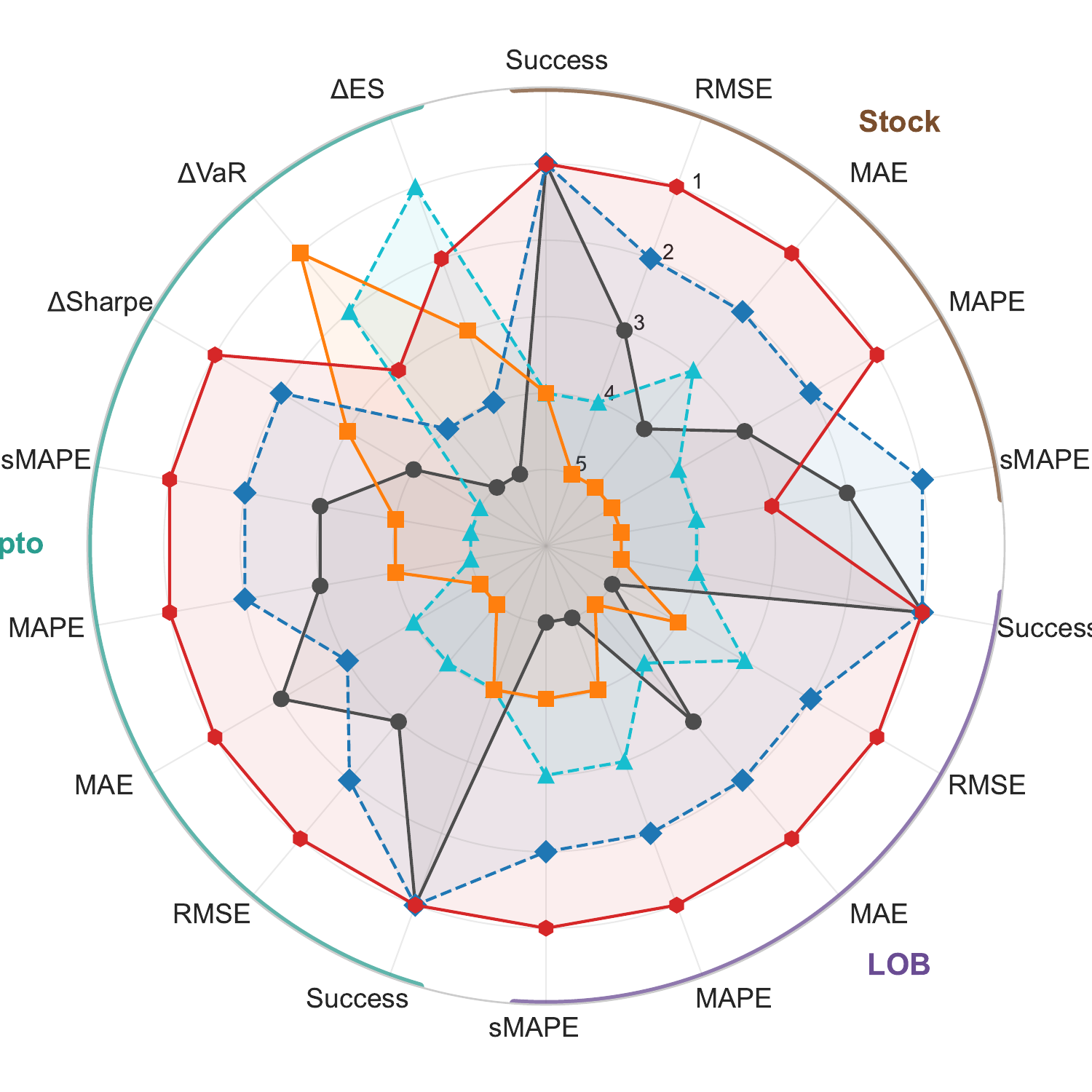}}}%
\hfill
\subfigure[Generation Ranking]{%
    \label{fig:tsg_radar}%
    \parbox[b]{0.48\linewidth}{%
        \centering
        \includegraphics[width=\linewidth, height=0.5cm, keepaspectratio=true]{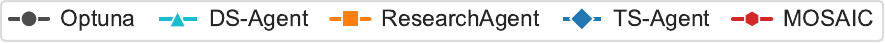} \\[2pt]
        \includegraphics[width=\linewidth]{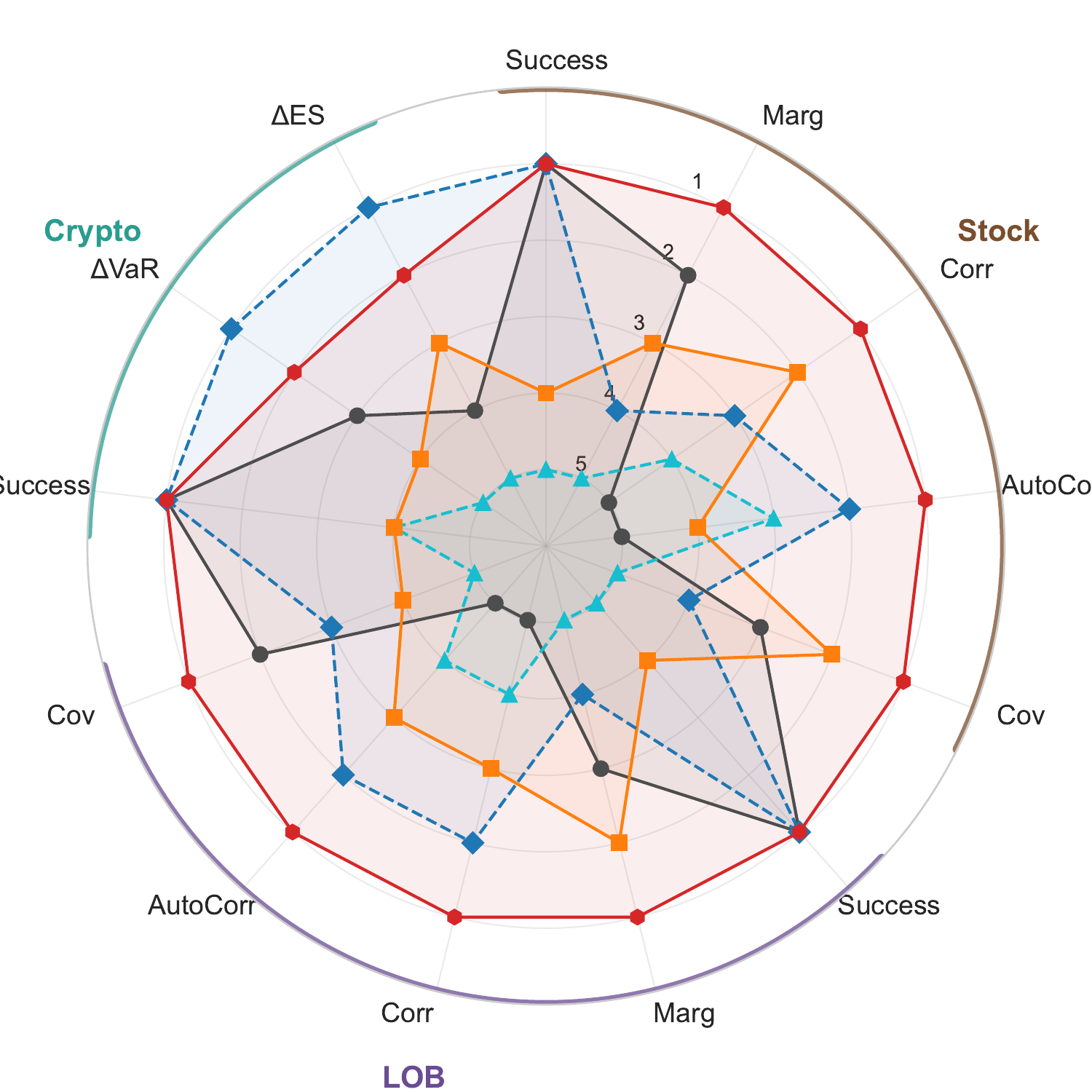}}}%
\vspace{-0.75em}
\caption{Radar charts for ranking.}
\label{fig:radar_wrap}
\vspace{-1.0em}
\end{wrapfigure}

\textsc{MOSAIC} consistently achieves the best results on forecasting tasks across three datasets and four LLM backbones, as supported by Figure~\ref{fig:tsf_metrics} and results in Appendix~\ref{app:tsf_detailed}.
 Averaged across backbones, RMSE is reduced by 8\% on LOB, 5\% on Crypto, and 3\% on Stock relative to TS-Agent, with individual backbones showing gains up to 19\% (LOB-Claude) and 8\% (Crypto-GPT-5.4). Compared to AutoGluon, average RMSE reductions reach 33\% on LOB. For risk-aware metrics on Crypto, average $\Delta$Sharpe is reduced by over 21\% relative to TS-Agent, with the GPT-4o backbone alone showing a 46\% reduction. Ranking profiles (Figure~\ref{fig:tsf_radar}) confirm this consistency, with the system forming the outermost contour on nearly all axes. The gains may result from the model generation module and the RL-based refinement policy, which learns non-myopic editing strategies from offline trajectories.

\paragraph{Time Series Generation.}
Figure~\ref{fig:tsg_radar} summarizes generation ranking profiles. \textsc{MOSAIC} dominates on LOB and Stock, reducing average Marginal distance by over 32\% and Correlation by 27\% compared to TS-Agent on LOB, and Marginal by over 23\% on Stock. These gains are driven by the model generation module's ability to synthesise novel generative architectures tailored to each dataset's distributional characteristics. On Crypto, it outperforms DS-Agent, ResearchAgent, and Optuna on both $\Delta$VaR and $\Delta$ES, though TS-Agent achieves lower tail-risk distances on this dataset. \textsc{MOSAIC} also maintains full coverage across all dataset--backbone pairs, unlike DS-Agent and ResearchAgent, whose success rates range from 20\% to 100\%. The enhanced reliability confirms the benefit of our RL approach on the generation tasks. Detailed tables are provided in Appendix~\ref{app:tsg_detailed}.


\subsection{Modular Analysis}

\begin{figure}[H]
  \centering
  \vspace{-0.5em}
  \includegraphics[width=0.6\linewidth]{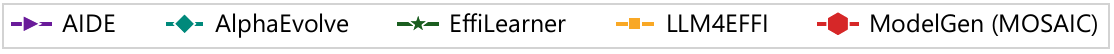} \\
  \vspace{-0.5em}
  \subfigure[Stock]{%
    \label{fig:tsf_cg_radar_stock}%
    \includegraphics[width=0.185\linewidth]{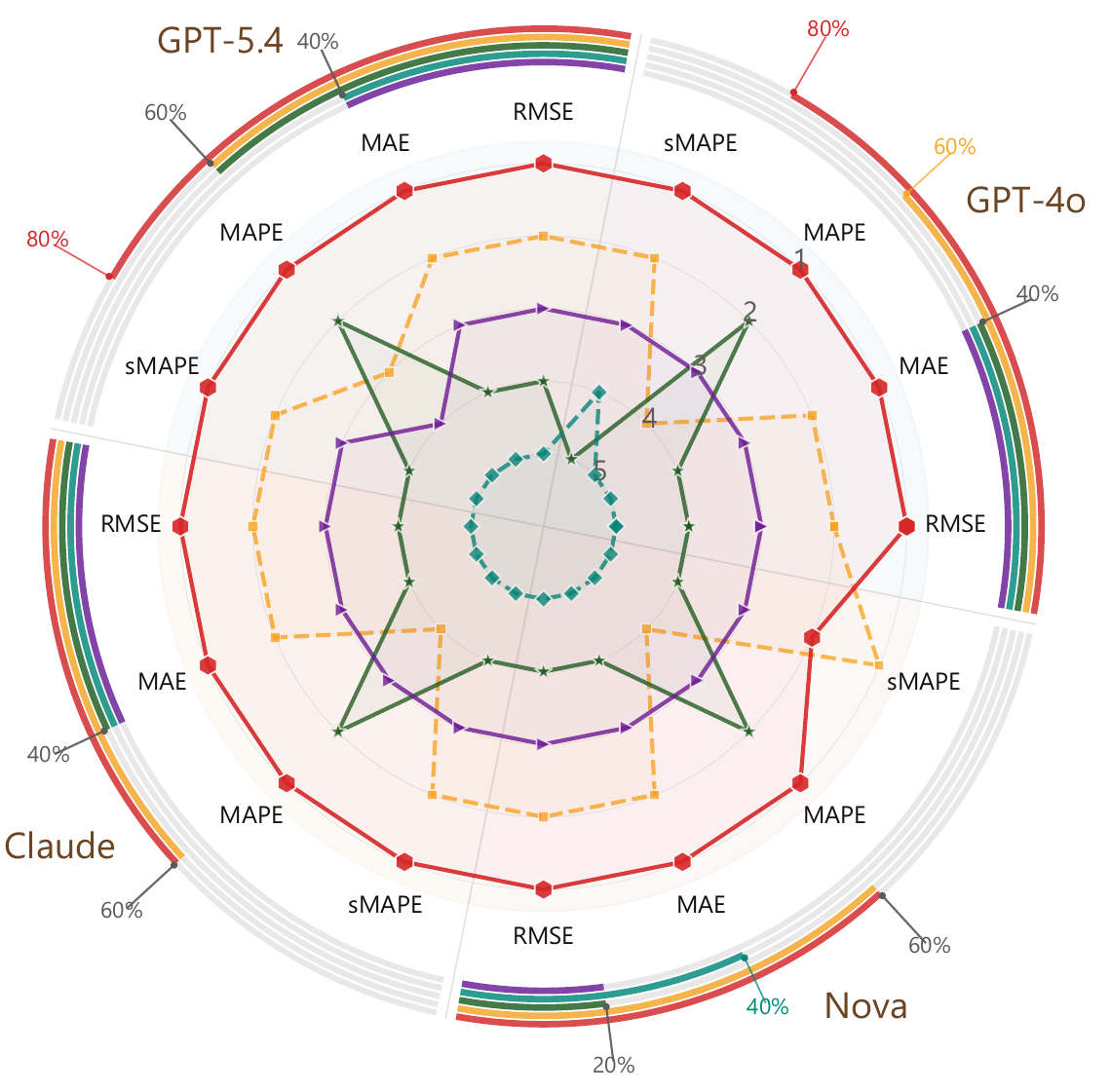}}
  \subfigure[LOB]{%
    \label{fig:tsf_cg_radar_lob}%
    \includegraphics[width=0.185\linewidth]{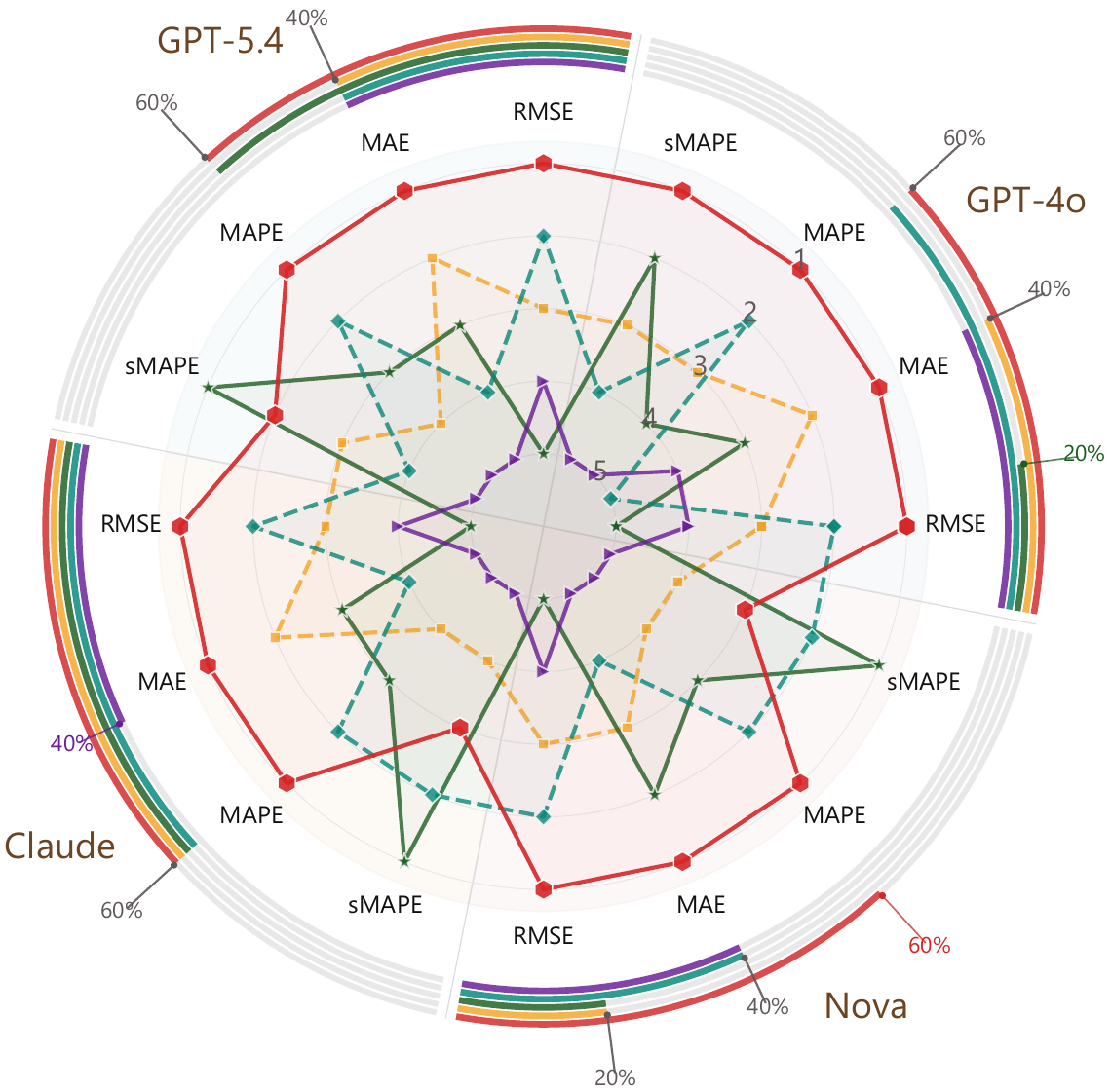}}
  \subfigure[Crypto]{%
    \label{fig:tsf_cg_radar_crypto}%
    \includegraphics[width=0.185\linewidth]{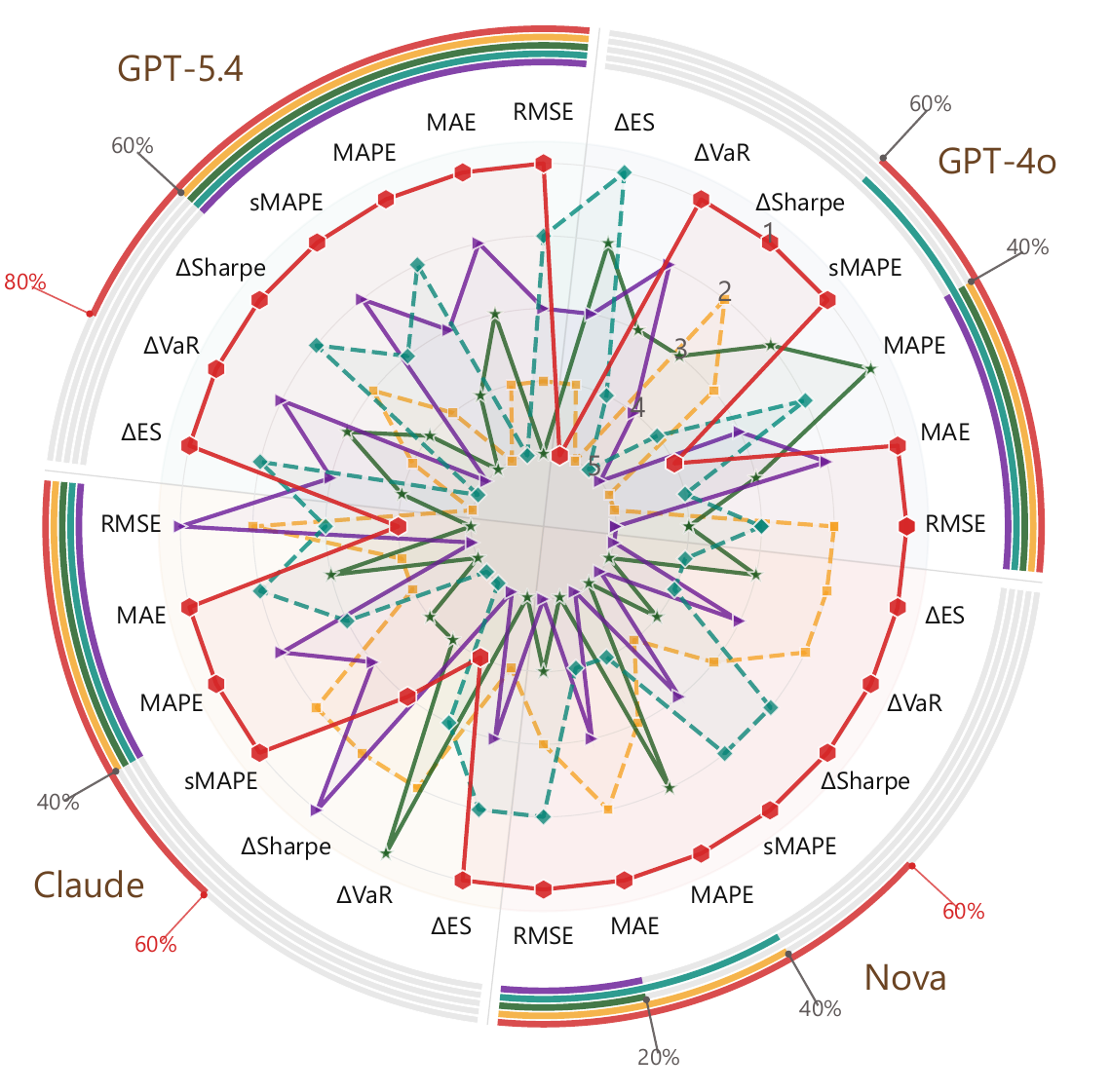}}
  \subfigure[Forecasting Avg.]{%
    \label{fig:tsf_cg_rank}%
    \includegraphics[width=0.210\linewidth]{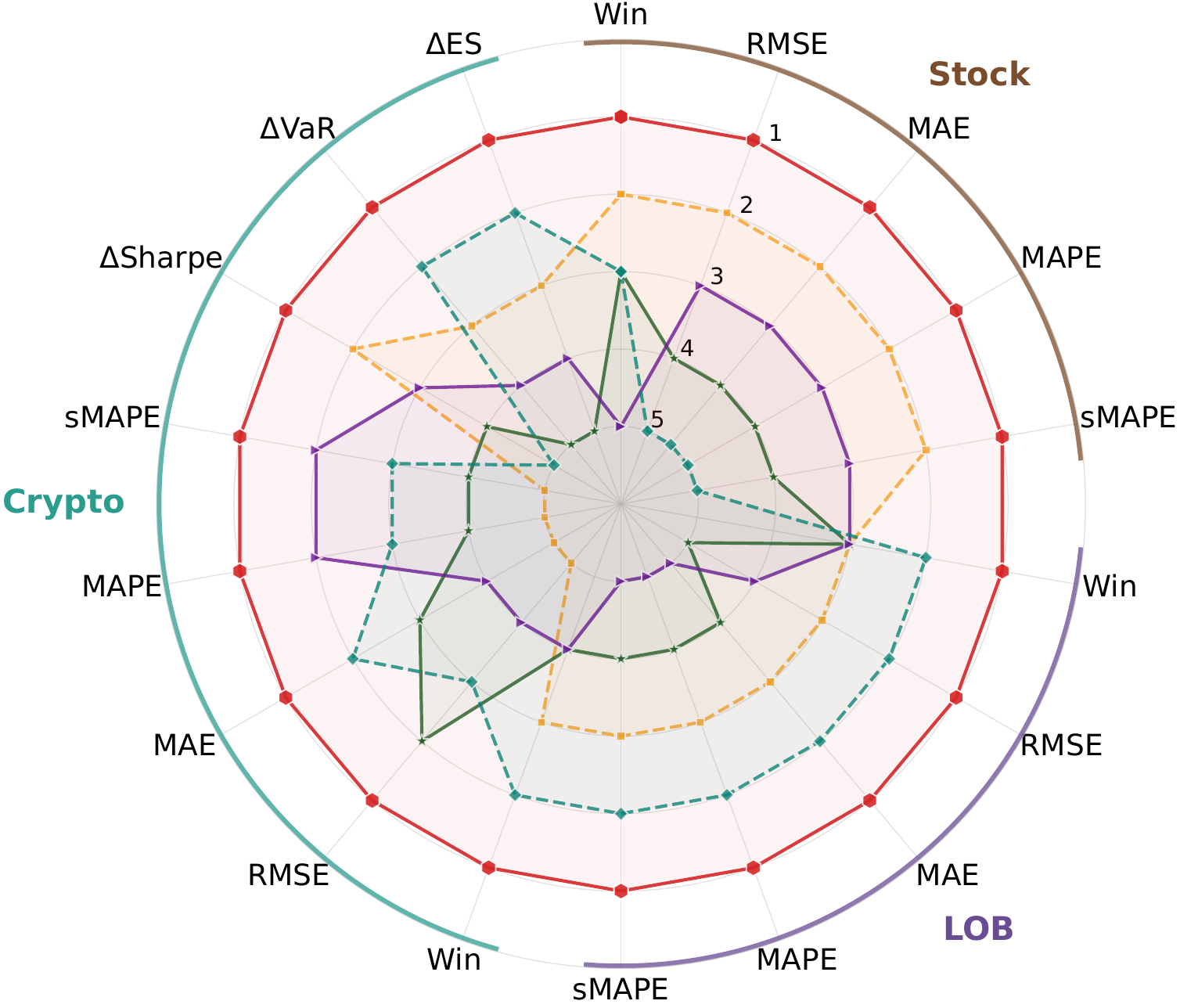}}
  \subfigure[Generation Avg.]{%
    \label{fig:tsg_cg_rank}%
    \includegraphics[width=0.195\linewidth]{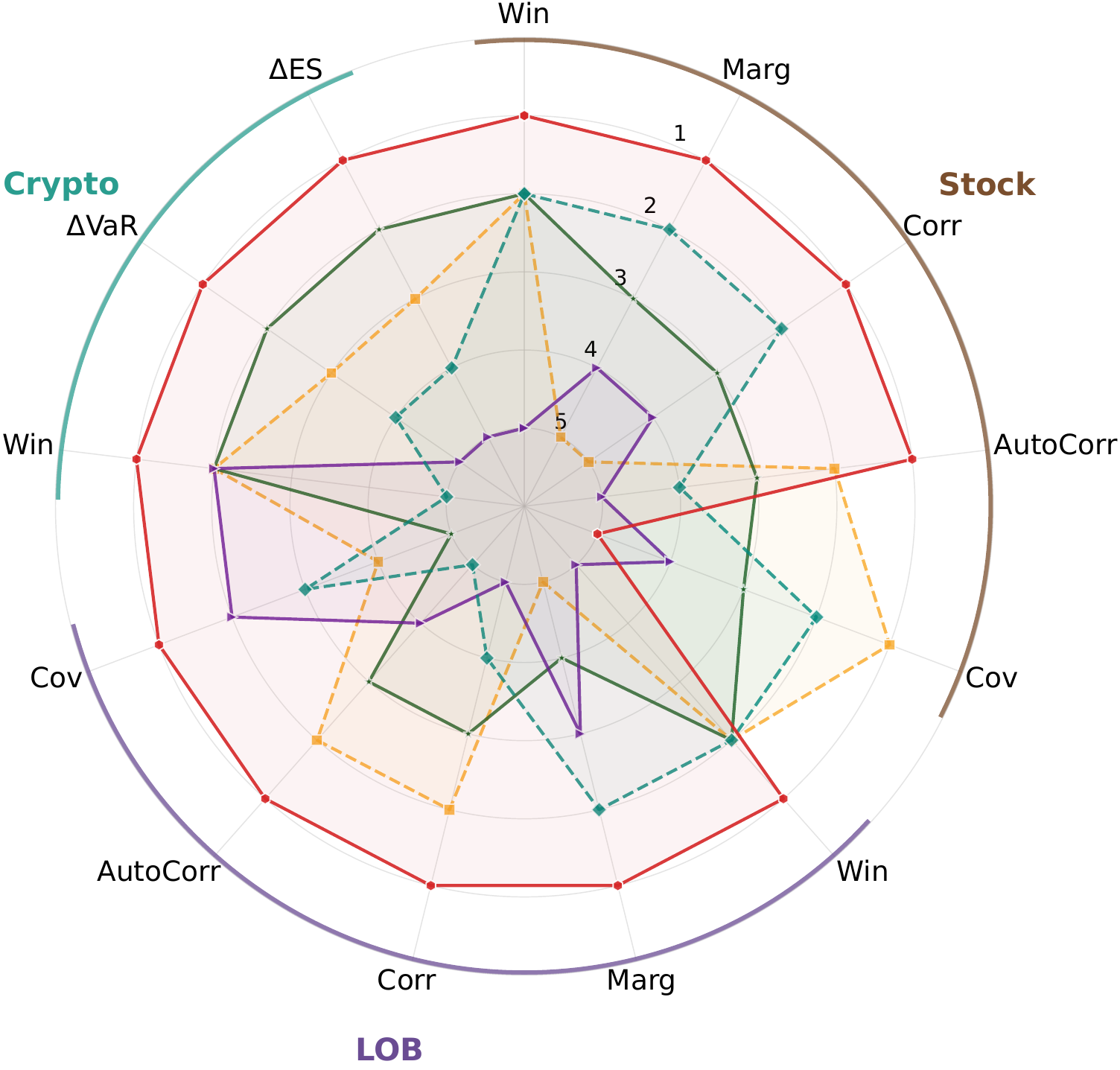}}
  \vspace{-0.75em}%
\caption{Model generation module comparison. (a)--(c): per-dataset Win Rate (outermost ring) and metric performance broken down by LLM backbone. (d)--(e): ranking profiles averaged across backbones for forecasting and generation respectively.}
\label{fig:cg_comparison}
  \vspace{-1.5em}
\end{figure}

\paragraph{Repository-Grounded Model Generation.}
Figure~\ref{fig:cg_comparison} shows the model generation module achieves the highest average Win Rate on both forecasting (65\% vs.\ 48\%) and generation (63\% vs.\ 42\%), with the best financial metrics on Crypto. These advantages stem from the architecture-family-aware blueprint and the validation loop that filters faulty implementations. More details are in Appendix~\ref{app:cg_detailed}.

\begin{figure}[H]
  \centering
  \vspace{-0.5em}
  \includegraphics[width=0.35\textwidth]{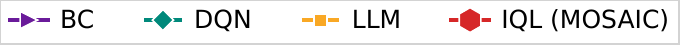} \\
  \vspace{-0.5em}
  \subfigure[Stock]{%
    \label{fig:tsf_rl_radar_stock}%
    \includegraphics[width=0.192\linewidth]{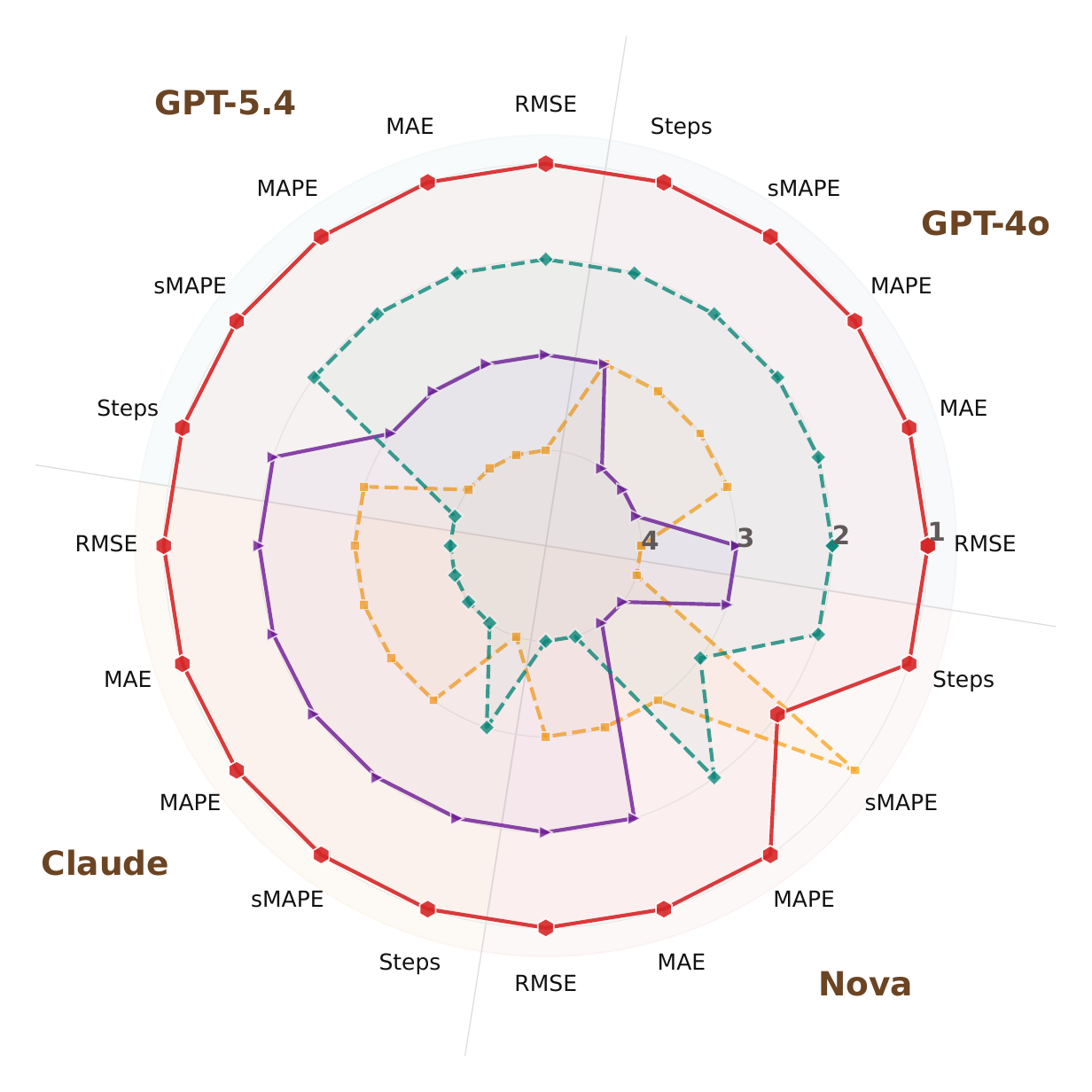}}
  \subfigure[LOB]{%
    \label{fig:tsf_rl_radar_lob}%
    \includegraphics[width=0.192\textwidth]{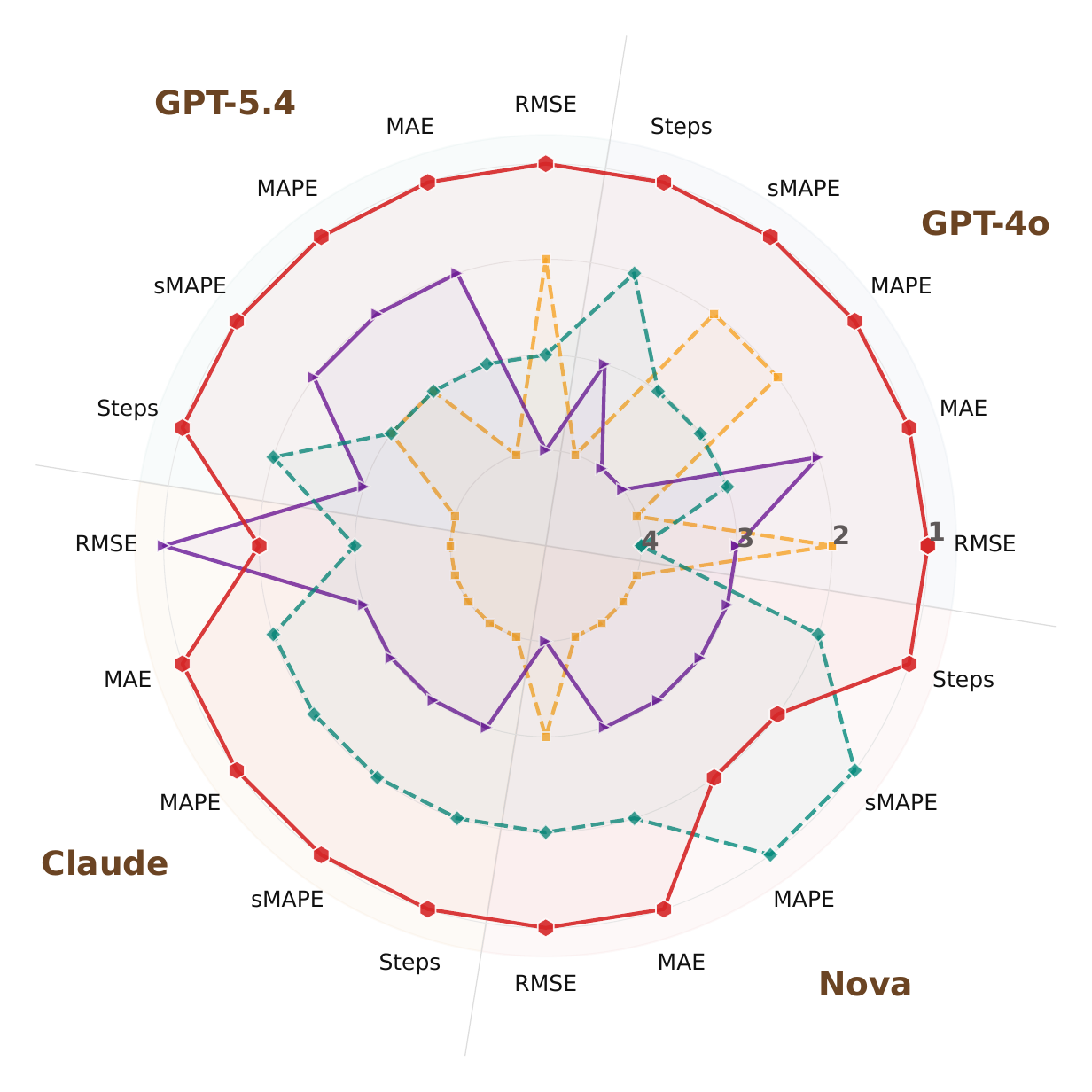}}
  \subfigure[Crypto]{%
    \label{fig:tsf_rl_radar_crypto}%
    \includegraphics[width=0.192\textwidth]{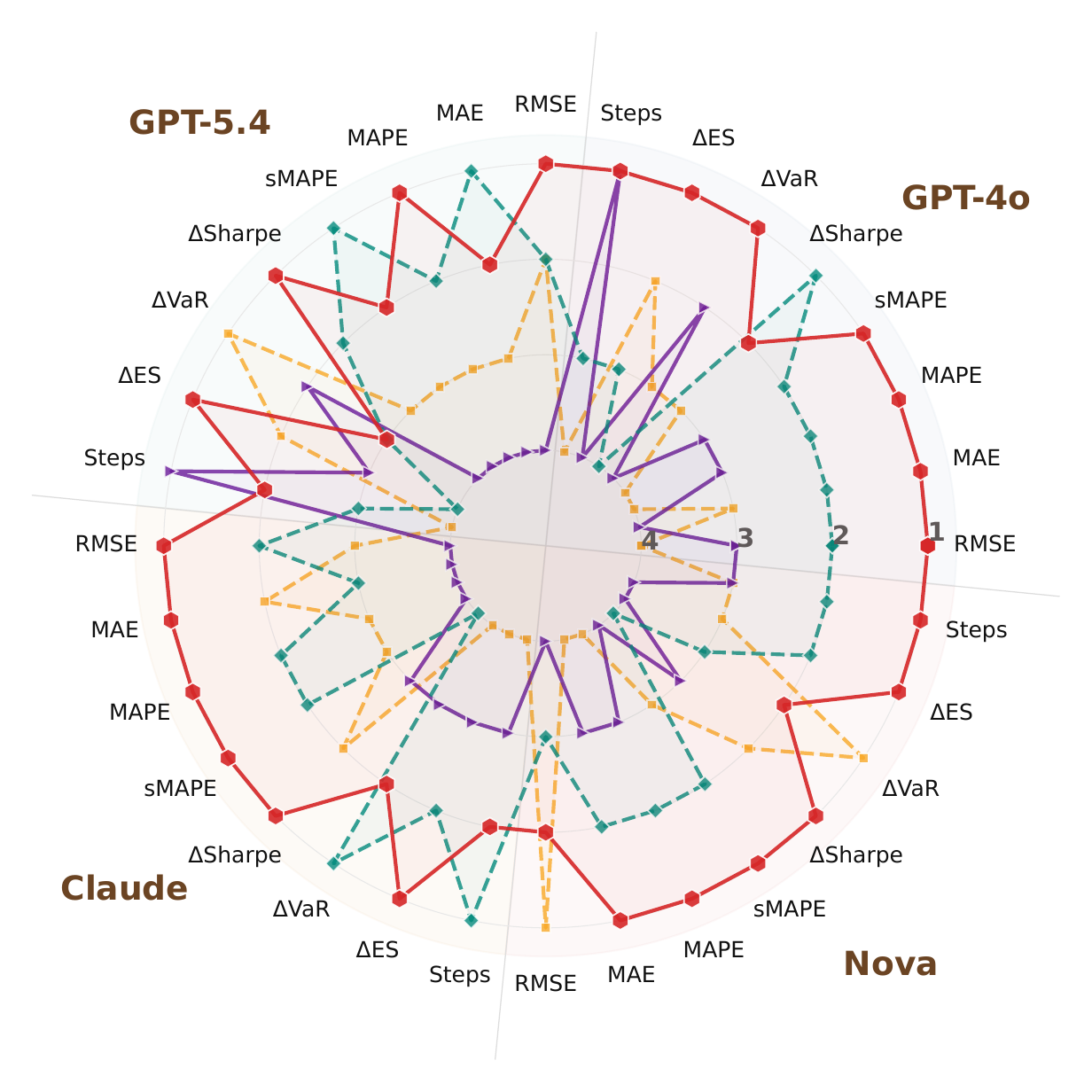}}
  \subfigure[Forecasting Avg.]{%
    \label{fig:tsf_rl_rank}%
    \includegraphics[width=0.192\textwidth]{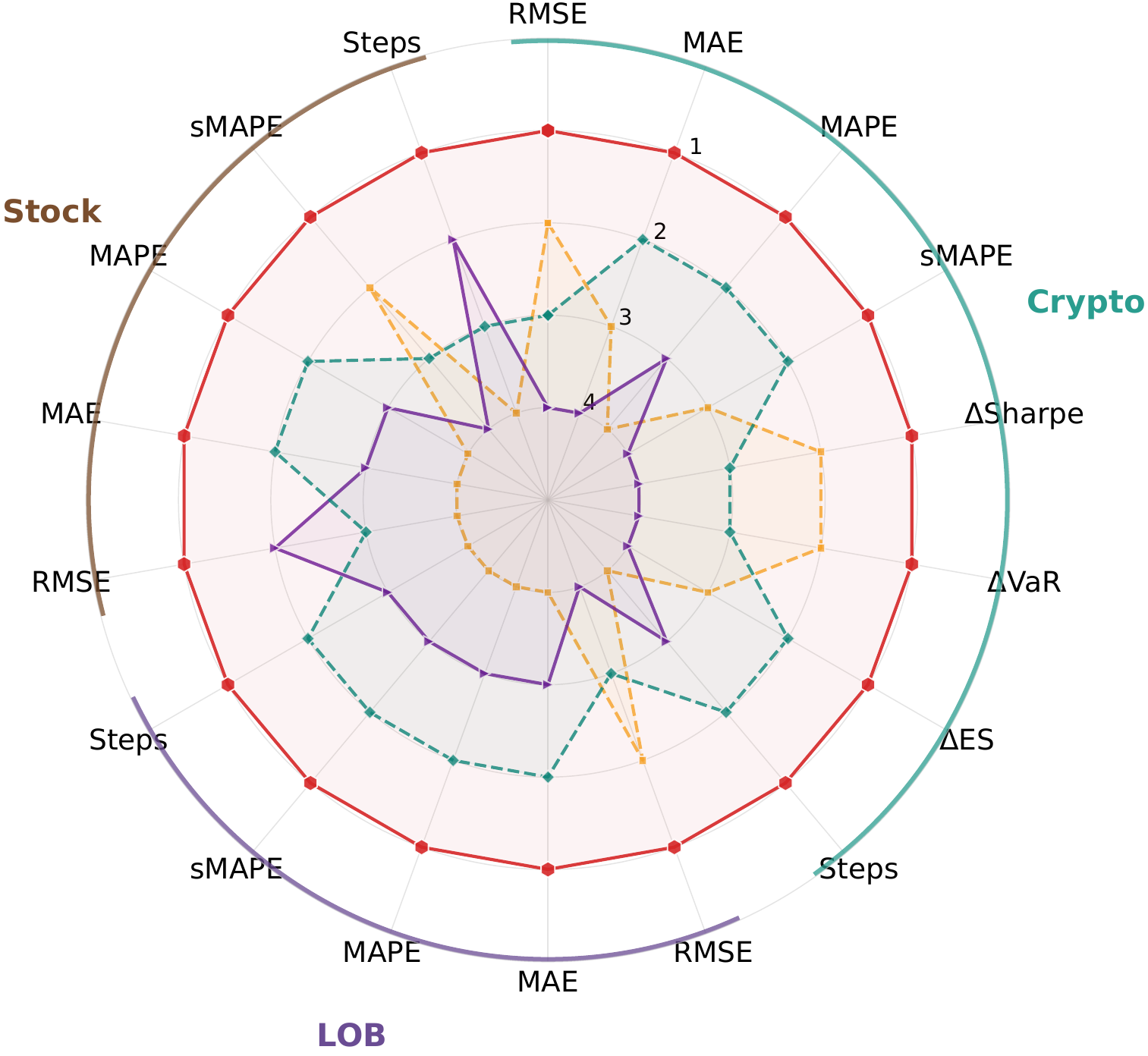}}
  \subfigure[Generation Avg.]{%
    \label{fig:tsg_rl_rank}%
    \includegraphics[width=0.192\textwidth]{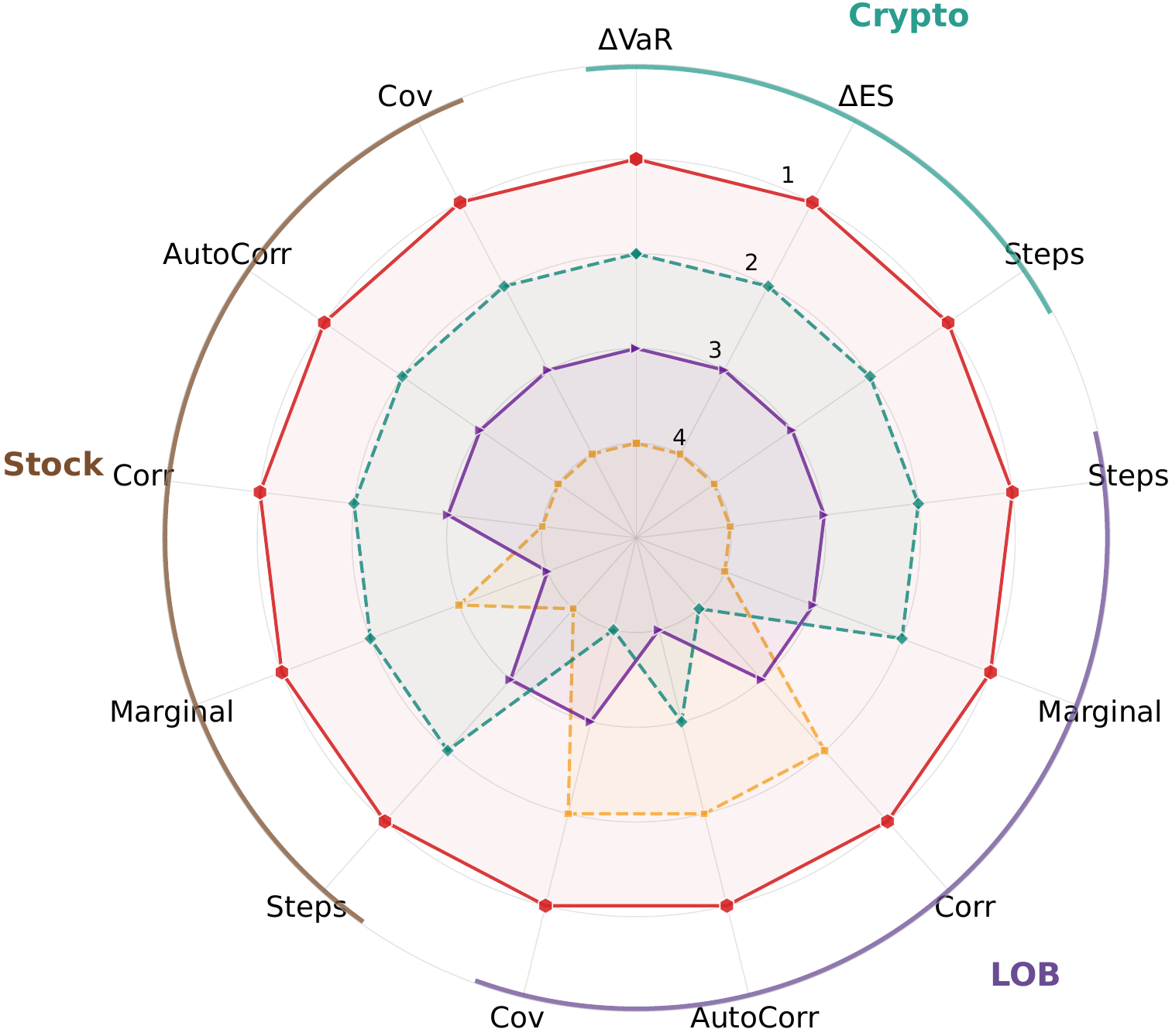}}
  \vspace{-0.75em}%
\caption{RL module comparison. (a)--(c): per-dataset metric performance broken down by LLM backbone. (d)--(e): ranking profiles averaged across backbones for forecasting and generation. Steps denotes Steps to Best Incumbent.}
\label{fig:rl_comparison}
  \vspace{-1.5em}
\end{figure}

\paragraph{Reinforcement Learning for Refinement.}
Figure~\ref{fig:rl_comparison} shows that IQL-based refinement wins 83\% of metric comparisons across datasets and LLM backbones, while converging in fewer steps than LLM-only refinement (8.7 vs.\ 10.0 on average). BC and DQN are competitive on isolated metrics but less consistent, suggesting that offline RL captures more reliable refinement strategies than purely prompt-driven or simpler RL alternatives. More details are in Appendix~\ref{app:rl_comparison}.

\subsection{Ablation Studies}

\paragraph{EDA, Feature Engineering, and Multimodal Support.}
Table~\ref{tab:eda_ablation} ablates the EDA and multimodal module, with TS-Agent as a lower bound. Removing these components leads to consistent degradation of up to 55\% across generation and financial metrics, confirming that statistical and visual priors are essential for downstream performance. Per-LLM breakdowns are provided in Appendix~\ref{app:eda_ablation}.

\begin{table*}[h]
  \centering
  \renewcommand{\arraystretch}{0.5}
  \caption{Ablation study of the EDA and multimodal module. All metrics are averaged across four LLM backbones. Subscripts F and G denote forecasting and generation respectively. Best results per dataset are \textbf{bolded}. -- indicates metric not applicable.}
  \label{tab:eda_ablation}
  \vspace{-0.5em}
  \resizebox{\textwidth}{!}{%
    \begin{tabular}{cl cccc cccc ccccc}
    \toprule
    \rowcolor[HTML]{FFF2CC}
    & & \multicolumn{4}{c}{\textbf{Forecasting}}
      & \multicolumn{4}{c}{\textbf{Generation}}
      & \multicolumn{5}{c}{\textbf{Financial (Crypto only)}} \\
    \cmidrule(lr){3-6}\cmidrule(lr){7-10}\cmidrule(lr){11-15}
    \rowcolor[HTML]{FFF2CC}
 \multirow{-3.5}{*}{\textbf{Dataset}} & \multirow{-3.5}{*}{\textbf{Variant}}
      & RMSE$\downarrow$ & MAE$\downarrow$ & MAPE$\downarrow$ & sMAPE$\downarrow$
      & Marg$\downarrow$ & Corr$\downarrow$ & ACorr$\downarrow$ & Cov$\downarrow$
      & $\Delta$Sharpe$_\text{F}$$\downarrow$ & $\Delta$VaR$_\text{F}$$\downarrow$ & $\Delta$ES$_\text{F}$$\downarrow$ & $\Delta$VaR$_\text{G}$$\downarrow$ & $\Delta$ES$_\text{G}$$\downarrow$ \\
    \midrule
    \rowcolor[HTML]{D9EAD3}
    \cellcolor{white} \multirow{5}{*}{Stock}
      &  MOSAIC & \textbf{7.907} & \textbf{4.853} & \textbf{2.011} & 2.004 & \textbf{0.387} & \textbf{0.356} & \textbf{0.085} & \textbf{0.892} & -- & -- & -- & -- & -- \\
      & EDA (MOSAIC) & 7.958 & 4.912 & 2.031 & 2.038 & 0.477 & 0.409 & 0.321 & 0.903 & -- & -- & -- & -- & -- \\
      & TS-Agent                  & 8.167 & 4.957 & 2.076 & \textbf{1.862} & 0.507 & 0.399 & 0.314 & 0.922 & -- & -- & -- & -- & -- \\
    \midrule
    \rowcolor[HTML]{D9EAD3}
    \cellcolor{white} \multirow{5}{*}{LOB}
      &  MOSAIC & \textbf{0.144} & \textbf{0.107} & \textbf{0.048} & \textbf{0.048} & \textbf{0.587} & \textbf{0.216} & \textbf{0.217} & \textbf{0.184} & -- & -- & -- & -- & -- \\
      & EDA (MOSAIC) & 0.158 & 0.112 & 0.050 & 0.050 & 0.911 & 0.457 & 0.627 & 0.289 & -- & -- & -- & -- & -- \\
      & TS-Agent                  & 0.157 & 0.115 & 0.052 & 0.052 & 0.870 & 0.296 & 0.295 & 0.255 & -- & -- & -- & -- & -- \\
    \midrule
    \rowcolor[HTML]{D9EAD3}
    \cellcolor{white} \multirow{5}{*}{Crypto}
      & MOSAIC & \textbf{0.210} & \textbf{0.052} & \textbf{1.548} & \textbf{1.553} & -- & -- & -- & -- & \textbf{7.970} & \textbf{0.013} & \textbf{0.015} & 0.075 & 0.095 \\
      & EDA (MOSAIC) & 0.216 & 0.054 & 1.607 & 1.608 & -- & -- & -- & -- & 11.188 & 0.067 & 0.094 & 0.088 & 0.122 \\
      & TS-Agent & 0.220 & 0.055 & 1.628 & 1.629 & -- & -- & -- & -- & 10.169 & 0.015 & 0.022 & \textbf{0.072} & \textbf{0.093} \\
    \bottomrule
    \end{tabular}
    \vspace{-1.0em}
  }
\end{table*}

\paragraph{Repository-Grounded Model Generation.}
Table~\ref{tab:cg_ablation} ablates the model generation module. Removing any single component leads to consistent degradation of up to 30\% across metrics, with Naive LLM performing worst. Win rates further confirm this, with the full ModelGen achieving up to 70\% across comparisons versus near-zero for ablated variants. Per-LLM breakdowns are in Appendix~\ref{app:cg_ablation}.

\begin{table*}[t]
  \centering
  \renewcommand{\arraystretch}{0.5}
  \caption{Ablation study of the model generation module. All metrics are averaged across four LLM backbones. Subscripts F and G denote forecasting and generation respectively. Best results per dataset are \textbf{bolded}. -- indicates metric not applicable.}
  \label{tab:cg_ablation}
  \vspace{-0.5em}
  \resizebox{\textwidth}{!}{%
    \begin{tabular}{cl cccc cccc ccccc cc}
    \toprule
    \rowcolor[HTML]{FFF2CC}
    & & \multicolumn{4}{c}{\textbf{Forecasting}} & \multicolumn{4}{c}{\textbf{Generation}} & \multicolumn{5}{c}{\textbf{Financial (Crypto only)}} & \multicolumn{2}{c}{\textbf{Win}} \\
    \cmidrule(lr){3-6}\cmidrule(lr){7-10}\cmidrule(lr){11-15}\cmidrule(lr){16-17}
    \rowcolor[HTML]{FFF2CC}
    \multirow{-3.5}{*}{\textbf{Dataset}} & \multirow{-3.5}{*}{\textbf{Variant}} & RMSE$\downarrow$ & MAE$\downarrow$ & MAPE$\downarrow$ & sMAPE$\downarrow$ & Marg$\downarrow$ & Corr$\downarrow$ & ACorr$\downarrow$ & Cov$\downarrow$ & $\Delta$Sharpe$_\text{F}$$\downarrow$ & $\Delta$VaR$_\text{F}$$\downarrow$ & $\Delta$ES$_\text{F}$$\downarrow$ & $\Delta$VaR$_\text{G}$$\downarrow$ & $\Delta$ES$_\text{G}$$\downarrow$ & F$\downarrow$ & G$\downarrow$ \\
    \midrule
    \rowcolor[HTML]{D9EAD3}
    \cellcolor{white} \multirow{5}{*}{Stock}
      &  ModelGen (MOSAIC)          & \textbf{7.841} & \textbf{4.829} & \textbf{2.002} & \textbf{1.990} & \textbf{0.355} & \textbf{0.359} & \textbf{0.083} & \textbf{0.905} & -- & -- & -- & -- & -- & \textbf{70} & \textbf{65} \\
      & w/o Mod.\ Und.  & 8.315 & 5.099 & 2.119 & 2.082 & 0.488 & 0.397 & 0.208 & 0.934 & -- & -- & -- & -- & -- & 20 & 10 \\
      & w/o Blueprint   & 8.434 & 5.133 & 2.140 & 2.100 & 0.502 & 0.401 & 0.310 & 0.934 & -- & -- & -- & -- & -- & 15 & 5 \\
      & w/o Verify      & 8.231 & 5.074 & 2.123 & 2.111 & 0.490 & 0.408 & 0.196 & 0.949 & -- & -- & -- & -- & -- & 15 & 15 \\
      & Naive LLM       & 8.279 & 5.032 & 2.111 & 2.155 & 0.509 & 0.403 & 0.301 & 0.935 & -- & -- & -- & -- & -- & 5 & 5 \\
    \midrule
    \rowcolor[HTML]{D9EAD3}
    \cellcolor{white} \multirow{5}{*}{LOB}
      & ModelGen (MOSAIC)           & \textbf{0.141} & \textbf{0.104} & \textbf{0.047} & \textbf{0.047} & \textbf{0.547} & \textbf{0.196} & \textbf{0.208} & \textbf{0.176} & -- & -- & -- & -- & -- & \textbf{60} & \textbf{65} \\
      & w/o Mod.\ Und.  & 0.154 & 0.113 & 0.051 & 0.051 & 0.753 & 0.252 & 0.261 & 0.227 & -- & -- & -- & -- & -- & 20 & 15 \\
      & w/o Blueprint   & 0.160 & 0.117 & 0.053 & 0.052 & 0.798 & 0.259 & 0.264 & 0.235 & -- & -- & -- & -- & -- & 5 & 5 \\
      & w/o Verify      & 0.159 & 0.116 & 0.053 & 0.052 & 0.843 & 0.297 & 0.292 & 0.249 & -- & -- & -- & -- & -- & 10 & 5 \\
      & Naive LLM       & 0.164 & 0.117 & 0.053 & 0.053 & 0.801 & 0.259 & 0.264 & 0.236 & -- & -- & -- & -- & -- & 5 & 5 \\
    \midrule
    \rowcolor[HTML]{D9EAD3}
    \cellcolor{white} \multirow{5}{*}{Crypto}
      & ModelGen (MOSAIC)           & \textbf{0.208} & \textbf{0.052} & \textbf{1.545} & \textbf{1.549} & -- & -- & -- & -- & \textbf{6.226} & \textbf{0.012} & \textbf{0.013} & 0.082 & 0.098 & \textbf{65} & \textbf{60} \\
      & w/o Mod.\ Und.  & 0.220 & 0.055 & 1.646 & 1.639 & -- & -- & -- & -- & 8.345 & 0.014 & 0.018 & 0.076 & 0.096 & 20 & 20 \\
      & w/o Blueprint   & 0.227 & 0.055 & 1.646 & 1.645 & -- & -- & -- & -- & 8.567 & 0.015 & 0.020 & 0.073 & \textbf{0.092} & 10 & 5 \\
      & w/o Verify      & 0.223 & 0.055 & 1.664 & 1.661 & -- & -- & -- & -- & 8.423 & 0.014 & 0.018 & 0.073 & 0.093 & 20 & 10 \\
      & Naive LLM       & 0.234 & 0.056 & 1.648 & 1.647 & -- & -- & -- & -- & 9.945 & 0.015 & 0.021 & \textbf{0.072} & 0.093 & 5 & 0 \\
    \bottomrule
    \end{tabular}
    \vspace{-1.0em}
  }
\end{table*}

\paragraph{Reinforcement Learning for Refinement.}

Table~\ref{tab:rl_ablation} ablates the RL module, where Soft Revert contributes most; its removal degrades performance by up to 27\% and increases refinement steps by roughly 20\%. Removing Trajectory Branching or Invalid Action Masking also hurts, though to a lesser extent. These results confirm that each RL component plays a distinct role in stable refinement. Per-LLM breakdowns are in Appendix~\ref{app:rl_ablation}.

\begin{table*}[!h]
  \centering
  \renewcommand{\arraystretch}{0.5}
  \caption{Ablation study of the RL module. All metrics are averaged across four LLM backbones. Subscripts F and G denote forecasting and generation respectively. Best results per dataset are \textbf{bolded}. -- indicates metric not applicable.}
  \label{tab:rl_ablation}
  \vspace{-0.5em}
  \resizebox{\textwidth}{!}{%
    \begin{tabular}{cl cccc cccc ccccc cc}
    \toprule
    \rowcolor[HTML]{FFF2CC}
    & & \multicolumn{4}{c}{\textbf{Forecasting}} & \multicolumn{4}{c}{\textbf{Generation}} & \multicolumn{5}{c}{\textbf{Financial (Crypto only)}} & \multicolumn{2}{c}{\textbf{Steps}} \\
    \cmidrule(lr){3-6}\cmidrule(lr){7-10}\cmidrule(lr){11-15}\cmidrule(lr){16-17}
    \rowcolor[HTML]{FFF2CC}
    \multirow{-3.5}{*}{\textbf{Dataset}} & \multirow{-3.5}{*}{\textbf{Variant}} & RMSE$\downarrow$ & MAE$\downarrow$ & MAPE$\downarrow$ & sMAPE$\downarrow$ & Marg$\downarrow$ & Corr$\downarrow$ & ACorr$\downarrow$ & Cov$\downarrow$ & $\Delta$Sharpe$_\text{F}$$\downarrow$ & $\Delta$VaR$_\text{F}$$\downarrow$ & $\Delta$ES$_\text{F}$$\downarrow$ & $\Delta$VaR$_\text{G}$$\downarrow$ & $\Delta$ES$_\text{G}$$\downarrow$ & F$\downarrow$ & G$\downarrow$ \\
    \midrule
    \rowcolor[HTML]{D9EAD3}
     \cellcolor{white} \multirow{4}{*}{Stock}
     & IQL (MOSAIC)  & \textbf{8.062} & \textbf{4.918} & \textbf{2.036} & \textbf{2.036} & \textbf{0.464} & \textbf{0.355} & \textbf{0.087} & \textbf{0.877} & -- & -- & -- & -- & -- & \textbf{8.65} & \textbf{8.70} \\
     & w/o Invalid Action Masking & 8.111 & 4.955 & 2.059 & 2.057 & 0.481 & 0.371 & 0.095 & 0.892 & -- & -- & -- & -- & -- & 9.05 & 9.10 \\
     & w/o Soft Revert & 8.296 & 5.094 & 2.143 & 2.135 & 0.560 & 0.428 & 0.119 & 0.984 & -- & -- & -- & -- & -- & 10.40 & 10.55 \\
     & w/o Trajectory Branching & 8.211 & 5.028 & 2.104 & 2.102 & 0.519 & 0.397 & 0.105 & 0.943 & -- & -- & -- & -- & -- & 9.75 & 9.70 \\
    \midrule
    \rowcolor[HTML]{D9EAD3}
    \cellcolor{white} \multirow{4}{*}{LOB} 
    &  IQL (MOSAIC) & \textbf{0.149} & \textbf{0.113} & \textbf{0.051} & \textbf{0.051} & \textbf{0.674} & \textbf{0.257} & \textbf{0.250} & \textbf{0.209} & -- & -- & -- & -- & -- & \textbf{8.70} & \textbf{8.85} \\
     & w/o Invalid Action Masking & 0.151 & 0.114 & 0.052 & 0.052 & 0.709 & 0.265 & 0.256 & 0.216 & -- & -- & -- & -- & -- & 9.10 & 9.25 \\
     & w/o Soft Revert & 0.160 & 0.124 & 0.056 & 0.056 & 0.858 & 0.306 & 0.314 & 0.271 & -- & -- & -- & -- & -- & 10.45 & 10.65 \\
     & w/o Trajectory Branching & 0.155 & 0.119 & 0.054 & 0.054 & 0.777 & 0.283 & 0.286 & 0.243 & -- & -- & -- & -- & -- & 9.75 & 9.85 \\
    \midrule
    \rowcolor[HTML]{D9EAD3}
     \cellcolor{white} \multirow{4}{*}{Crypto} 
     & IQL (MOSAIC) & \textbf{0.213} & \textbf{0.053} & \textbf{1.562} & \textbf{1.569} & -- & -- & -- & -- & \textbf{11.773} & \textbf{0.016} & \textbf{0.019} & \textbf{0.064} & \textbf{0.091} & \textbf{8.60} & \textbf{8.55} \\
     & w/o Invalid Action Masking & 0.216 & 0.054 & 1.584 & 1.588 & -- & -- & -- & -- & 12.179 & 0.017 & 0.020 & 0.068 & 0.097 & 9.00 & 9.00 \\
     & w/o Soft Revert & 0.226 & 0.057 & 1.661 & 1.656 & -- & -- & -- & -- & 13.686 & 0.018 & 0.025 & 0.081 & 0.119 & 10.40 & 10.35 \\
     & w/o Trajectory Branching & 0.221 & 0.055 & 1.623 & 1.622 & -- & -- & -- & -- & 12.928 & 0.018 & 0.023 & 0.076 & 0.108 & 9.75 & 9.65 \\
    \bottomrule
    \end{tabular}
  }
\end{table*}

\subsection{Case Study}
We present case studies of \textbf{model generation example} and \textbf{task generation} in Appendix~\ref{app:case_study}. In particular, the model generation example illustrate how \textsc{MOSAIC} composes novel architectures: GPT-4o builds a hybrid model for cryptocurrency forecasting (RMSE 0.207 vs.\ 0.218 for TS-Agent), while Claude Opus 4 assembles a diffusion-based model for LOB generation (Marginal 0.554 vs.\ 0.767 for Diffusion-TS alone).
\section{Conclusion}
\label{sec:conclusion}
We introduced \textsc{MOSAIC}, a structured agentic framework for automated data-science modelling that bridges rigid AutoML pipelines and unconstrained LLM-based code generation. Rather than treating model building as direct program synthesis, \textsc{MOSAIC} formulates it as a structured search over admissible modelling workflows by introducing an explicit blueprint layer between task understanding and executable implementation. Experiments on financial time-series tasks show that this module-based workflow construction and optimisation approach yields robust, high-performing, and interpretable data-science solutions.

\paragraph{Limitations and Future Work.} 
The \textsc{MOSAIC} framework is currently subject to several limitations. Its performance ceiling depends on the quality of knowledge banks, and it assumes reliable module extraction from heterogeneous repositories, which may not always hold in practice. In addition, repeated execution-based verification can incur substantial computational cost. Finally, stronger guarantees are needed to characterize the dimensional and functional compatibility of synthesized modules. Future work could address these challenges through adaptive methods that enhance the knowledge base via RL-driven retrieval and refinement, together with more robust verification procedures. It would also be valuable to extend \textsc{MOSAIC} beyond financial time series to broader application domains.

\section*{Acknowledgments}
AT, YA, YB, XX, and XL are supported by the National Research Foundation, Singapore, under its CyberSG R\&D Programme (CRPO Award No: CRPO-GC5-NUS-006), the Ministry of Education, Singapore, under its MOE AcRF TIER 1 Grant (T1 251RES2517), and the National University of Singapore, School of Computing Seed Fund. 
LJ and LS acknowledge the support of the UKRI Prosperity Partnership Scheme (FAIR) under the EPSRC Grant EP/V056883/1 and The Alan Turing Institute. HN is supported by the EPSRC Program Grant [Grant No. UKRI1010] entitled ``High order mathematical and computational infrastructure for streamed data that enhance contemporary generative and large language models''. 
WG is supported by the EPSRC through the UCL EPSRC Landscape Award (UELA) under Grant EP/Z534882/1. 
KZ was supported by the EPSRC Centre for Doctoral Training in Mathematical Modelling, Analysis and Computation (MAC-MIGS) funded by the UK Engineering and Physical Sciences Research Council (grant EP/S023291/1), Heriot-Watt University and the University of Edinburgh.

\bibliographystyle{unsrtnat}
\bibliography{main}

\begin{thebibliography}{54}
\providecommand{\natexlab}[1]{#1}
\providecommand{\url}[1]{\texttt{#1}}
\expandafter\ifx\csname urlstyle\endcsname\relax
  \providecommand{\doi}[1]{doi: #1}\else
  \providecommand{\doi}{doi: \begingroup \urlstyle{rm}\Url}\fi

\bibitem[Domingos(2012)]{domingos}
Pedro~M. Domingos.
\newblock A few useful things to know about machine learning.
\newblock \emph{Commun. {ACM}}, pages 78--87, 2012.

\bibitem[Feurer et~al.(2022)Feurer, Eggensperger, Falkner, Lindauer, and Hutter]{feurer2022auto}
Matthias Feurer, Katharina Eggensperger, Stefan Falkner, Marius Lindauer, and Frank Hutter.
\newblock Auto-sklearn 2.0: Hands-free automl via meta-learning.
\newblock \emph{JMLR}, pages 1--61, 2022.

\bibitem[Thornton et~al.(2013)Thornton, Hutter, Hoos, and Leyton-Brown]{thornton2013auto}
Chris Thornton, Frank Hutter, Holger~H Hoos, and Kevin Leyton-Brown.
\newblock Auto-weka: Combined selection and hyperparameter optimization of classification algorithms.
\newblock In \emph{KDD}, pages 847--855, 2013.

\bibitem[Yao et~al.(2023)Yao, Zhao, Yu, Du, Shafran, Narasimhan, and Cao]{yao2022react}
Shunyu Yao, Jeffrey Zhao, Dian Yu, Nan Du, Izhak Shafran, Karthik Narasimhan, and Yuan Cao.
\newblock React: Synergizing reasoning and acting in language models.
\newblock In \emph{ICLR}, 2023.

\bibitem[Wang et~al.(2024{\natexlab{a}})Wang, Xie, Jiang, Mandlekar, Xiao, Zhu, Fan, and Anandkumar]{wang2023voyager}
Guanzhi Wang, Yuqi Xie, Yunfan Jiang, Ajay Mandlekar, Chaowei Xiao, Yuke Zhu, Linxi Fan, and Anima Anandkumar.
\newblock Voyager: An open-ended embodied agent with large language models.
\newblock \emph{Trans. Mach. Learn. Res.}, 2024{\natexlab{a}}.

\bibitem[Lv et~al.(2024)Lv, Xia, and Huang]{lv2024codeact}
Weijie Lv, Xuan Xia, and Sheng-Jun Huang.
\newblock Codeact: Code adaptive compute-efficient tuning framework for code llms.
\newblock \emph{arXiv preprint arXiv:2408.02193}, 2024.

\bibitem[Guo et~al.(2024)Guo, Deng, Wen, Chen, Chang, and Wang]{guo2024ds}
Siyuan Guo, Cheng Deng, Ying Wen, Hechang Chen, Yi~Chang, and Jun Wang.
\newblock Ds-agent: Automated data science by empowering large language models with case-based reasoning.
\newblock In \emph{ICML}, pages 16813--16848, 2024.

\bibitem[Baek et~al.(2025)Baek, Jauhar, Cucerzan, and Hwang]{baek2024researchagent}
Jinheon Baek, Sujay~Kumar Jauhar, Silviu Cucerzan, and Sung~Ju Hwang.
\newblock Researchagent: Iterative research idea generation over scientific literature with large language models.
\newblock In \emph{NAACL}, pages 6709--6738, 2025.

\bibitem[Ang et~al.(2025)Ang, Bao, Jiang, Tao, Tung, Szpruch, and Ni]{tsagent}
Yihao Ang, Yifan Bao, Lei Jiang, Jiajie Tao, Anthony K.~H. Tung, Lukasz Szpruch, and Hao Ni.
\newblock Structured agentic workflows for financial time-series modeling with llms and reflective feedback.
\newblock In \emph{ICAIF}, 2025.

\bibitem[Shinn et~al.(2023)Shinn, Cassano, Gopinath, Narasimhan, and Yao]{shinn2023reflexion}
Noah Shinn, Federico Cassano, Ashwin Gopinath, Karthik Narasimhan, and Shunyu Yao.
\newblock Reflexion: Language agents with verbal reinforcement learning.
\newblock In \emph{NeurIPS}, pages 8634--8652, 2023.

\bibitem[Novikov et~al.(2025)Novikov, V{\~u}, Eisenberger, Dupont, Huang, Wagner, Shirobokov, Kozlovskii, Ruiz, Mehrabian, Kumar, See, Chaudhuri, Holland, Davies, Nowozin, Kohli, and Balog]{novikov2025alphaevolve}
Alexander Novikov, Ng{\^a}n V{\~u}, Marvin Eisenberger, Emilien Dupont, Po-Sen Huang, Adam~Zsolt Wagner, Sergey Shirobokov, Borislav Kozlovskii, Francisco~JR Ruiz, Abbas Mehrabian, M.~Pawan Kumar, Abigail See, Swarat Chaudhuri, George Holland, Alex Davies, Sebastian Nowozin, Pushmeet Kohli, and Matej Balog.
\newblock Alphaevolve: A coding agent for scientific and algorithmic discovery.
\newblock \emph{arXiv preprint arXiv:2506.13131}, 2025.

\bibitem[Ellis et~al.(2023)Ellis, Wong, Nye, Sable-Meyer, Cary, Anaya~Pozo, Hewitt, Solar-Lezama, and Tenenbaum]{ellis2023dreamcoder}
Kevin Ellis, Lionel Wong, Maxwell Nye, Mathias Sable-Meyer, Luc Cary, Lore Anaya~Pozo, Luke Hewitt, Armando Solar-Lezama, and Joshua~B Tenenbaum.
\newblock Dreamcoder: growing generalizable, interpretable knowledge with wake--sleep bayesian program learning.
\newblock \emph{Philosophical Transactions of the Royal Society A: Mathematical, Physical and Engineering Sciences}, 2023.

\bibitem[Shchur et~al.(2023)Shchur, Turkmen, Erickson, Shen, Shirkov, Hu, and Wang]{agtimeseries}
Oleksandr Shchur, Ali~Caner Turkmen, Nick Erickson, Huibin Shen, Alexander Shirkov, Tony Hu, and Bernie Wang.
\newblock Autogluon-timeseries: Automl for probabilistic time series forecasting.
\newblock In \emph{ICLR}, pages 9--1, 2023.

\bibitem[Akiba et~al.(2019)Akiba, Sano, Yanase, Ohta, and Koyama]{akiba2019optuna}
Takuya Akiba, Shotaro Sano, Toshihiko Yanase, Takeru Ohta, and Masanori Koyama.
\newblock {O}ptuna: A next-generation hyperparameter optimization framework.
\newblock In \emph{KDD}, pages 2623--2631, 2019.

\bibitem[Zoph and Le(2017)]{zoph2016neural}
Barret Zoph and Quoc~V Le.
\newblock Neural architecture search with reinforcement learning.
\newblock In \emph{ICLR}, 2017.

\bibitem[Elsken et~al.(2019)Elsken, Metzen, and Hutter]{elsken2019neural}
Thomas Elsken, Jan~Hendrik Metzen, and Frank Hutter.
\newblock Neural architecture search: A survey.
\newblock \emph{Journal of Machine Learning Research}, pages 1--21, 2019.

\bibitem[Liu et~al.(2018)Liu, Simonyan, and Yang]{liu2018darts}
Hanxiao Liu, Karen Simonyan, and Yiming Yang.
\newblock Darts: Differentiable architecture search.
\newblock In \emph{ICLR}, 2018.

\bibitem[Lu et~al.(2024)Lu, Lu, Lange, Foerster, Clune, and Ha]{lu2024ai}
Chris Lu, Cong Lu, Robert~Tjarko Lange, Jakob Foerster, Jeff Clune, and David Ha.
\newblock The ai scientist: Towards fully automated open-ended scientific discovery.
\newblock \emph{arXiv preprint arXiv:2408.06292}, 2024.

\bibitem[Schmidgall et~al.(2025)Schmidgall, Su, Wang, Sun, Wu, Yu, Liu, Moor, Liu, and Barsoum]{schmidgall2025agent}
Samuel Schmidgall, Yusheng Su, Ze~Wang, Ximeng Sun, Jialian Wu, Xiaodong Yu, Jiang Liu, Michael Moor, Zicheng Liu, and Emad Barsoum.
\newblock Agent laboratory: Using llm agents as research assistants.
\newblock In \emph{EMNLP}, pages 5977--6043, 2025.

\bibitem[Yuan et~al.(2025)Yuan, Pahwa, Chang, Kaba, Jiang, Ma, Zhang, and Sunkara]{yuan2025automated}
Michelle Yuan, Khushbu Pahwa, Shuaichen Chang, Mustafa Kaba, Jiarong Jiang, Xiaofei Ma, Yi~Zhang, and Monica Sunkara.
\newblock Automated composition of agents: A knapsack approach for agentic component selection.
\newblock \emph{arXiv preprint arXiv:2510.16499}, 2025.

\bibitem[Jiang et~al.(2025)Jiang, Schmidt, Srikanth, Xu, Kaplan, Jacenko, and Wu]{jiang2025aide}
Zhengyao Jiang, Dominik Schmidt, Dhruv Srikanth, Dixing Xu, Ian Kaplan, Deniss Jacenko, and Yuxiang Wu.
\newblock Aide: Ai-driven exploration in the space of code.
\newblock \emph{arXiv preprint arXiv:2502.13138}, 2025.

\bibitem[Huang et~al.(2024)Huang, Dai, Weng, Wu, Qing, Cui, Guo, and Zhang]{huang2024effilearner}
Dong Huang, Jianbo Dai, Han Weng, Puzhen Wu, Yuhao Qing, Heming Cui, Zhijiang Guo, and Jie Zhang.
\newblock Effilearner: Enhancing efficiency of generated code via self-optimization.
\newblock In \emph{NeurIPS}, 2024.

\bibitem[Ye et~al.(2025)Ye, Huang, Zhang, Ma, Liu, Yin, and Wang]{ye2025llm4effi}
Tong Ye, Weigang Huang, Xuhong Zhang, Tengfei Ma, Peiyu Liu, Jianwei Yin, and Wenhai Wang.
\newblock {LLM4EFFI:} leveraging large language models to enhance code efficiency and correctness.
\newblock \emph{arXiv preprint arXiv:2502.18489}, 2025.

\bibitem[Ouyang et~al.(2022)Ouyang, Wu, Jiang, Almeida, Wainwright, Mishkin, Zhang, Agarwal, Slama, Ray, Schulman, Hilton, Kelton, Miller, Simens, Askell, Welinder, Christiano, Leike, and Lowe]{ouyang2022training}
Long Ouyang, Jeffrey Wu, Xu~Jiang, Diogo Almeida, Carroll~L. Wainwright, Pamela Mishkin, Chong Zhang, Sandhini Agarwal, Katarina Slama, Alex Ray, John Schulman, Jacob Hilton, Fraser Kelton, Luke Miller, Maddie Simens, Amanda Askell, Peter Welinder, Paul~F. Christiano, Jan Leike, and Ryan Lowe.
\newblock Training language models to follow instructions with human feedback.
\newblock In \emph{NeurIPS}, 2022.

\bibitem[Ma et~al.(2024)Ma, Liang, Wang, Huang, Bastani, Jayaraman, Zhu, Fan, and Anandkumar]{ma2023eureka}
Yecheng~Jason Ma, William Liang, Guanzhi Wang, De-An Huang, Osbert Bastani, Dinesh Jayaraman, Yuke Zhu, Linxi Fan, and Anima Anandkumar.
\newblock Eureka: Human-level reward design via coding large language model.
\newblock In \emph{ICLR}, 2024.

\bibitem[Kostrikov et~al.(2022)Kostrikov, Nair, and Levine]{kostrikov2021offline}
Ilya Kostrikov, Ashvin Nair, and Sergey Levine.
\newblock Offline reinforcement learning with implicit q-learning.
\newblock \emph{ICLR}, 2022.

\bibitem[Huang and Onta{\~n}{\'o}n(2022)]{huang2020closer}
Shengyi Huang and Santiago Onta{\~n}{\'o}n.
\newblock A closer look at invalid action masking in policy gradient algorithms.
\newblock In \emph{FLAIRS}, 2022.

\bibitem[Alshiekh et~al.(2018)Alshiekh, Bloem, Ehlers, K{\"o}nighofer, Niekum, and Topcu]{alshiekh2018safe}
Mohammed Alshiekh, Roderick Bloem, R{\"u}diger Ehlers, Bettina K{\"o}nighofer, Scott Niekum, and Ufuk Topcu.
\newblock Safe reinforcement learning via shielding.
\newblock In \emph{AAAI}, pages 2669--2678, 2018.

\bibitem[Ang et~al.(2026)Ang, Wang, Huang, Bao, Xi, Tung, Jin, and Huang]{ctbench}
Yihao Ang, Qiang Wang, Qiang Huang, Yifan Bao, Xinyu Xi, Anthony~KH Tung, Chen Jin, and Zhiyong Huang.
\newblock Ctbench: Cryptocurrency time series generation benchmark.
\newblock In \emph{ICLR}, 2026.

\bibitem[Pomerleau(1988)]{pomerleau1988alvinn}
Dean~A Pomerleau.
\newblock Alvinn: An autonomous land vehicle in a neural network.
\newblock In \emph{NeurIPS}, 1988.

\bibitem[Mnih et~al.(2015)Mnih, Kavukcuoglu, Silver, Rusu, Veness, Bellemare, Graves, Riedmiller, Fidjeland, Ostrovski, Petersen, Beattie, Sadik, Antonoglou, King, Kumaran, Wierstra, Legg, and Hassabis]{mnih2015human}
Volodymyr Mnih, Koray Kavukcuoglu, David Silver, Andrei~A. Rusu, Joel Veness, Marc~G. Bellemare, Alex Graves, Martin~A. Riedmiller, Andreas Fidjeland, Georg Ostrovski, Stig Petersen, Charles Beattie, Amir Sadik, Ioannis Antonoglou, Helen King, Dharshan Kumaran, Daan Wierstra, Shane Legg, and Demis Hassabis.
\newblock Human-level control through deep reinforcement learning.
\newblock \emph{nature}, pages 529--533, 2015.

\bibitem[OpenAI(2026)]{openai2025gpt5}
OpenAI.
\newblock Gpt-5 system card.
\newblock \url{https://arxiv.org/abs/2601.03267}, 2026.

\bibitem[OpenAI(2023)]{openai2023gpt4}
OpenAI.
\newblock Gpt-4 technical report.
\newblock \url{https://arxiv.org/abs/2303.08774}, 2023.

\bibitem[Anthropic(2025)]{anthropic2025claude}
Anthropic.
\newblock System card: Claude opus 4 \& claude sonnet 4.
\newblock \url{https://www.anthropic.com/}, 2025.

\bibitem[Intelligence(2025)]{aws2025nova}
Amazon Artificial~General Intelligence.
\newblock The amazon nova family of models: Technical report and model card.
\newblock \url{https://assets.amazon.science/}, 2025.

\bibitem[Ni et~al.(2021)Ni, Szpruch, Sabate-Vidales, Xiao, Wiese, and Liao]{sig-gan}
Hao Ni, Lukasz Szpruch, Marc Sabate-Vidales, Baoren Xiao, Magnus Wiese, and Shujian Liao.
\newblock Sig-wasserstein gans for time series generation.
\newblock In \emph{ICAIF}, pages 1--8, 2021.

\bibitem[Lou et~al.(2023)Lou, Li, and Ni]{lou2023pcf}
Hang Lou, Siran Li, and Hao Ni.
\newblock Pcf-gan: generating sequential data via the characteristic function of measures on the path space.
\newblock In \emph{NeurIPS}, pages 39755--39781, 2023.

\bibitem[Ni et~al.(2023)Ni, Szpruch, and Tao]{ni2023regime}
Hao Ni, Lukasz Szpruch, and Jiajie Tao.
\newblock {Regime-switching Financial Time-Series Generation}.
\newblock \url{https://github.com/tjj0502/hackathon_starting_kit}, 2023.
\newblock ICAIF 2023 Hackathon.

\bibitem[Ni et~al.(2024)Ni, Szpruch, Tao, and Long]{ni2024crypto}
Hao Ni, Lukasz Szpruch, Jiajie Tao, and Yang Long.
\newblock {Crypto Market Simulation for Risk Estimation}.
\newblock \url{https://hackathon.deepintomlf.ai/competitions/40}, 2024.
\newblock Antalpha ICAIF 2024 Hackathon.

\bibitem[Wu et~al.(2021)Wu, Xu, Wang, and Long]{wu2021autoformer}
Haixu Wu, Jiehui Xu, Jianmin Wang, and Mingsheng Long.
\newblock Autoformer: Decomposition transformers with auto-correlation for long-term series forecasting.
\newblock In \emph{NeurIPS}, pages 22419--22430, 2021.

\bibitem[Nie et~al.(2023)Nie, Nguyen, Sinthong, and Kalagnanam]{patchtst}
Yuqi Nie, Nam~H Nguyen, Phanwadee Sinthong, and Jayant Kalagnanam.
\newblock A time series is worth 64 words: Long-term forecasting with transformers.
\newblock In \emph{ICLR}, 2023.

\bibitem[Wu et~al.(2023)Wu, Hu, Liu, Zhou, Wang, and Long]{wutimesnet}
Haixu Wu, Tengge Hu, Yong Liu, Hang Zhou, Jianmin Wang, and Mingsheng Long.
\newblock Timesnet: Temporal 2d-variation modeling for general time series analysis.
\newblock In \emph{ICLR}, 2023.

\bibitem[Zeng et~al.(2023)Zeng, Chen, Zhang, and Xu]{dlinear}
Ailing Zeng, Muxi Chen, Lei Zhang, and Qiang Xu.
\newblock Are transformers effective for time series forecasting?
\newblock In \emph{AAAI}, pages 11121--11128, 2023.

\bibitem[Wang et~al.(2024{\natexlab{b}})Wang, Wu, Shi, Hu, Luo, Ma, Zhang, and Zhou]{wangtimemixer}
Shiyu Wang, Haixu Wu, Xiaoming Shi, Tengge Hu, Huakun Luo, Lintao Ma, James~Y Zhang, and Jun Zhou.
\newblock Timemixer: Decomposable multiscale mixing for time series forecasting.
\newblock In \emph{ICLR}, 2024{\natexlab{b}}.

\bibitem[Yoon et~al.(2019)Yoon, Jarrett, and Van~der Schaar]{yoon2019time}
Jinsung Yoon, Daniel Jarrett, and Mihaela Van~der Schaar.
\newblock Time-series generative adversarial networks.
\newblock In \emph{NeurIPS}, pages 5509--5519, 2019.

\bibitem[Esteban et~al.(2017)Esteban, Hyland, and R{\"a}tsch]{esteban2017real}
Crist{\'o}bal Esteban, Stephanie~L Hyland, and Gunnar R{\"a}tsch.
\newblock Real-valued (medical) time series generation with recurrent conditional gans.
\newblock \emph{arXiv preprint arXiv:1706.02633}, 2017.

\bibitem[Wiese et~al.(2020)Wiese, Knobloch, Korn, and Kretschmer]{quant-gan}
Magnus Wiese, Robert Knobloch, Ralf Korn, and Peter Kretschmer.
\newblock Quant gans: deep generation of financial time series.
\newblock \emph{Quantitative Finance}, pages 1419--1440, 2020.

\bibitem[Seyfi et~al.(2022)Seyfi, Rajotte, and Ng]{seyfi2022COSCO-GAN}
Ali Seyfi, Jean-Francois Rajotte, and Raymond Ng.
\newblock Generating multivariate time series with common source coordinated gan (cosci-gan).
\newblock In \emph{NeurIPS}, pages 32777--32788, 2022.

\bibitem[Desai et~al.(2021)Desai, Freeman, Wang, and Beaver]{desai2021timevae}
Abhyuday Desai, Cynthia Freeman, Zuhui Wang, and Ian Beaver.
\newblock Timevae: A variational auto-encoder for multivariate time series generation.
\newblock \emph{arXiv preprint arXiv:2111.08095}, 2021.

\bibitem[Yuan and Qiao(2024)]{yuan2024diffusion}
Xinyu Yuan and Yan Qiao.
\newblock Diffusion-ts: Interpretable diffusion for general time series generation.
\newblock \emph{arXiv preprint arXiv:2403.01742}, 2024.

\bibitem[Wang et~al.(2025)Wang, Pan, Wu, Zhang, and Wu]{wang2025fourierflow}
Haixin Wang, Jiashu Pan, Hao Wu, Fan Zhang, and Tailin Wu.
\newblock Fourierflow: Frequency-aware flow matching for generative turbulence modeling.
\newblock \emph{arXiv preprint arXiv:2506.00862}, 2025.

\bibitem[Zhou et~al.(2023)Zhou, Poli, Xu, Massaroli, and Ermon]{zhou2023deep}
Linqi Zhou, Michael Poli, Winnie Xu, Stefano Massaroli, and Stefano Ermon.
\newblock Deep latent state space models for time-series generation.
\newblock In \emph{ICML}, pages 42625--42643, 2023.

\bibitem[Galib et~al.(2024)Galib, Tan, and Luo]{galib2024fide}
Asadullah~Hill Galib, Pang-Ning Tan, and Lifeng Luo.
\newblock Fide: Frequency-inflated conditional diffusion model for extreme-aware time series generation.
\newblock In \emph{NeurIPS}, pages 114434--114457, 2024.

\bibitem[Said and Dickey(1984)]{said1984testing}
Said~E Said and David~A Dickey.
\newblock Testing for unit roots in autoregressive-moving average models of unknown order.
\newblock \emph{Biometrika}, pages 599--607, 1984.

\end{thebibliography}

\newpage
\appendix

\section{Task Generation Details}
\label{app:taskgen}
We employ an automated task-generation module to construct diverse, executable time-series forecasting and generation tasks, forming a scalable corpus of tasks that can be converted to trajectory data for offline RL of refinement policies. 

Given a high-level instruction specifying task-generation requirements, the pipeline generates tasks with controlled variation in task type, data source, asset universe, temporal range, window construction, data transformation, and evaluation protocol, inducing a heterogeneous task distribution over which the refinement policy can observe diverse modelling choices, failure modes, and refinement strategies. Concrete examples of such instructions are provided in the case study in Appendix \ref{app:case_study}.

Let $\mathcal{B}_{\mathrm{data}}$ denote the dataset bank and $\mathcal{B}_{\mathrm{eval}}$ the evaluation-template bank. Given the instruction $\mathcal{I}$, the task generator $\mathcal{G}$ produces
$$\{\mathcal{T}_i\}_{i=1}^{N}
    =
    \mathcal{G}(\mathcal{I};\mathcal{B}_{\mathrm{data}},\mathcal{B}_{\mathrm{eval}}),$$
where each 
$\mathcal{T}_i=(\mathcal{W}_i, \mathcal{D}_i, \mathcal{L}_i)$
is a financial time-series task consisting of a dataset $\mathcal{D}_i$, an evaluation protocol $\mathcal{L}_i$, and a structured task description $\mathcal{W}_i$. 

\subsection{Dataset Bank}
The dataset bank $\mathcal{B}_{\text{data}}$ serves as a controlled source of diversity for RL training, and is intentionally constructed to be distinct from the datasets in the main experiments (LOB, Crypto, Stock). In particular, we enforce disjointness at the asset-universe level to eliminate overlap between training tasks and evaluation tasks to prevent information leakage. The bank consists of representative financial time-series panels across equities, foreign exchange, commodities, cryptocurrencies, and interest-rate term structures. Each dataset is associated with metadata describing its asset universe, sampling frequency, date coverage, market context, and admissible transformations. All datasets are preprocessed into aligned multivariate panels with missing observations removed. Table~\ref{tab:data_bank} summarises the datasets currently included in $\mathcal{B}_{\mathrm{data}}$.

\begin{table}[h]
\centering
\small
\caption{Dataset bank $\mathcal{B}_{\text{data}}$.}
\begin{tabular}{p{3.3cm}p{1.4cm}p{2.0cm}p{5.5cm}}
\toprule
\rowcolor[HTML]{FFF2CC}
\textbf{Dataset} & \textbf{Frequency} & \textbf{Source} & \textbf{Coverage} \\
\midrule
60 U.S. large-cap equities & daily & Yahoo Finance &
U.S. large-cap equities across energy, healthcare, financials,
consumer, technology, communications, and industrial sectors;
2002-05-23 to 2026-04-08 \\
16 Foreign exchange pairs & daily & Yahoo Finance &
G10 and emerging-market currency pairs in a largely USD-centered daily
panel; 2005-01-03 to 2026-03-30 \\
17 commodity futures series & daily & Yahoo Finance &
Precious metals, energy, grains, and soft commodities represented by
daily futures series; 2000-11-02 to 2026-03-31 \\
22 Cryptocurrencies & hourly & Yahoo Finance &
USD-denominated crypto assets including majors, smart-contract
platforms, DeFi and infrastructure tokens, privacy coins, and meme
assets, non overlapping with tasks used for evaluation; 2024-04-09 to 2026-04-08 \\
10 U.S. Treasury maturities & daily & FRED &
U.S. par-yield maturities covering the short, intermediate, and long
ends of the yield curve; 1993-10-01 to 2023-12-29 \\
\bottomrule
\end{tabular}
\label{tab:data_bank}
\end{table}

\subsection{Generation Process}
The generation process consists of two main steps. First, the module plans multiple structured task specifications from the same instruction. Second, each specification is instantiated into an executable task through dataset construction, evaluation construction, validation, constrained refinement, and final description synthesis. 
See concrete examples of generated forecasting and generation tasks in Appendix \ref{app:case_study} on Case Study.

\section{Knowledge Bank Details}
\label{app:knowledge_banks}

The system maintains three read-only external resources that collectively provide domain knowledge, executable code, and refinement guidance.

\paragraph{Case Bank ($\mathcal{E}_{case}$).}
The case bank contains curated task-solution pairs spanning both forecasting and generation. Each entry records a task description, dataset characteristics, the recommended model, and a summary of the solution methodology. Cases are drawn from financial benchmarks, competition entries, and peer-reviewed studies on financial time-series analysis~\citep{ctbench,sig-gan,lou2023pcf,ni2023regime,ni2024crypto}. The case bank covers diverse asset classes including equities, cryptocurrencies, foreign exchange, and commodities, with tasks ranging from multi-step price forecasting to regime-switching generation and conditional market simulation. In addition, successful trajectories collected during offline RL training are summarized by the LLM and appended to the case bank, enriching it with empirically validated solutions.

\paragraph{Code Base ($\mathcal{E}_{code}$).}
The code base consists of a model bank and an evaluation measure bank, both implemented as standalone Python modules.

The \textit{model bank} includes five forecasting models spanning major architectural paradigms: Autoformer \citep{wu2021autoformer} and PatchTST \citep{patchtst}(Transformer-based), TimesNet \citep{wutimesnet}(temporal CNN), DLinear \citep{dlinear}(linear), and TimeMixer \citep{wangtimemixer}(token-mixing). For generation, it includes ten models: TimeGAN \citep{yoon2019time}, RCGAN \citep{esteban2017real}, PCFGAN \citep{lou2023pcf}, QuantGAN \citep{quant-gan}, COSCIGAN \citep{seyfi2022COSCO-GAN} (GAN-based), TimeVAE \citep{desai2021timevae} (VAE-based), DiffusionTS \citep{yuan2024diffusion} (diffusion-based), FourierFlow \citep{wang2025fourierflow} (flow-based), and LS4 \citep{zhou2023deep}, FIDE \citep{galib2024fide} .

The \textit{evaluation measure bank} \citep{tsagent} provides standardised metric implementations for both tasks. For forecasting: RMSE, MAE, MAPE, sMAPE, and downstream financial metrics including Sharpe Ratio, Value at Risk (VaR), and Expected Shortfall (ES). For generation: Marginal Distribution, Correlation, Autocorrelation, and Covariance scores, along with VaR and ES for tail-risk evaluation.

\paragraph{Refinement Knowledge Bank ($\mathcal{E}_{refine}$).}
The refinement knowledge bank encodes domain-specific best practices organized into three categories:

\textit{Data preparation}: standardization, outlier handling (winsorising, clipping), and temporal data augmentation where domain-appropriate.

\textit{Training optimisation}: early stopping based on validation metrics, learning rate scheduling (e.g., ReduceLROnPlateau), mini-batch strategies that preserve temporal ordering, and regularization via weight decay and dropout.

\textit{Tuning and evaluation}: systematic hyperparameter search over learning rate, batch size, and regularization coefficients; walk-forward validation for temporal generalisation; and overfitting diagnostics comparing training and validation trajectories.

Each entry provides actionable guidance that the refinement stage can directly apply as code modifications without altering the model architecture.

\section{Semantic-aware EDA with Multimodal Support}
\label{app:eda}
\paragraph{EDA \& Feature Engineering.}
To bridge the gap between raw numerical data and LLM reasoning, the \textsc{MOSAIC} pipeline begins with a semantic-aware EDA module that maps continuous statistical properties into discrete, text-based semantic anchors. Rather than merely computing basic statistics, the module extracts high-level meta-features, such as data scale, task type, volatility, feature interdependence via correlation, and Augmented Dickey-Fuller test results~\citep{said1984testing}. Alongside the data schema and target variable statistics, these structured tags embed essential statistical priors directly into the prompt context.

\paragraph{Multimodal Support by VLM.}
Beyond textual summaries, complex temporal dynamics, such as regime shifts, heavy-tailed distributions, and cross-variable dependencies, are more naturally captured visually. 
To accommodate this, the EDA module generates standardised analytical plots (e.g., temporal line charts, correlation heatmaps) for subsequent analysis by a VLM. 
The resulting visual report and textual EDA summary are concatenated with the task description to form an enriched, multimodal query. This unified query guides the case-based reasoning retrieval stage, helping to mitigate the semantic mismatch that arises when raw numerical data are matched against textual knowledge sources. Furthermore, injecting this context into the LLM prompt grounds model selection in explicit semantic and visual evidence. By leveraging these structured data priors, the system reduces the impact of pre-training bias to infer suitable architectural traits, prioritising models that intrinsically fit the target data distribution over generic baselines.

\section{Repository-Grounded Model Generation}
\label{app:codegen}

Beyond selecting from a fixed model bank, the model generation module 
balances retrieval and innovation: it identifies relevant existing models while using LLMs to recombine their components into new executable architectures, thereby expanding the search space and potential performance ceiling.

\paragraph{Architecture Family Analysis.}
Based on the top-$k$ retrieved models, the system classifies candidates into architectural families (e.g., decomposition-based methods, MLP, and Transformers).
If models share the same architectural family, extracted modules (e.g., encoder blocks, embeddings) can be directly reused. Otherwise, the top-ranked model provides the backbone, and modules from other models inform high-level design principles for blueprint construction.

\paragraph{Module Identification and Extraction.}
For these candidates, the system performs static analysis of source code via abstract syntax tree parsing to identify reusable neural network modules. For each identified module, it extracts the source definition, forward pass signature, import dependencies, key operators (e.g., attention, convolution, decomposition). 

Each extracted module is then enriched with deterministic shape analysis, deriving expected input and output tensor dimensions 
from the source code, and with semantic annotations from an LLM capturing its functional role, design assumptions, and operational characteristics. Together, these form unified module representations that support downstream composition.

\paragraph{Blueprint Construction.} Given the module representations and the composition strategy determined above, the system constructs a structured blueprint by jointly considering the task profile, upstream data analysis results, and retrieved prior cases. The blueprint specifies which modules to use, their composition order, and the dimensional constraints
for inter-module compatibility, elevating  model descriptions into executable composition plans and separating architectural reasoning from code synthesis.

\paragraph{Code Generation.} Given the blueprint, an LLM generates executable model code. To promote diversity, each generation call is conditioned on a distinct structural variant within the blueprint's architectural framework, such as alternative gating mechanisms or different residual connection patterns, producing multiple candidate implementations from the same composition plan.

\paragraph{Execution, Verification, and Refinement.} Each generated candidate undergoes execution based validation. The system dynamically loads the generated code, instantiates the model with  appropriate task dimensions, and performs a forward pass to verify output shape consistency
and numerical stability. When verification fails, the system classifies the error type, constructs a targeted revision prompt incorporating the failing code and full diagnostic trace, and resubmits to the LLM for correction. This cycle repeats for a bounded number of iterations, yielding verified, executable model implementations ready for downstream training and evaluation.

\section{Trajectory Branching Mechanism Details}
\label{app:rl}
\begin{figure}[t]
    \centering
    \includegraphics[width=0.9\linewidth]
    {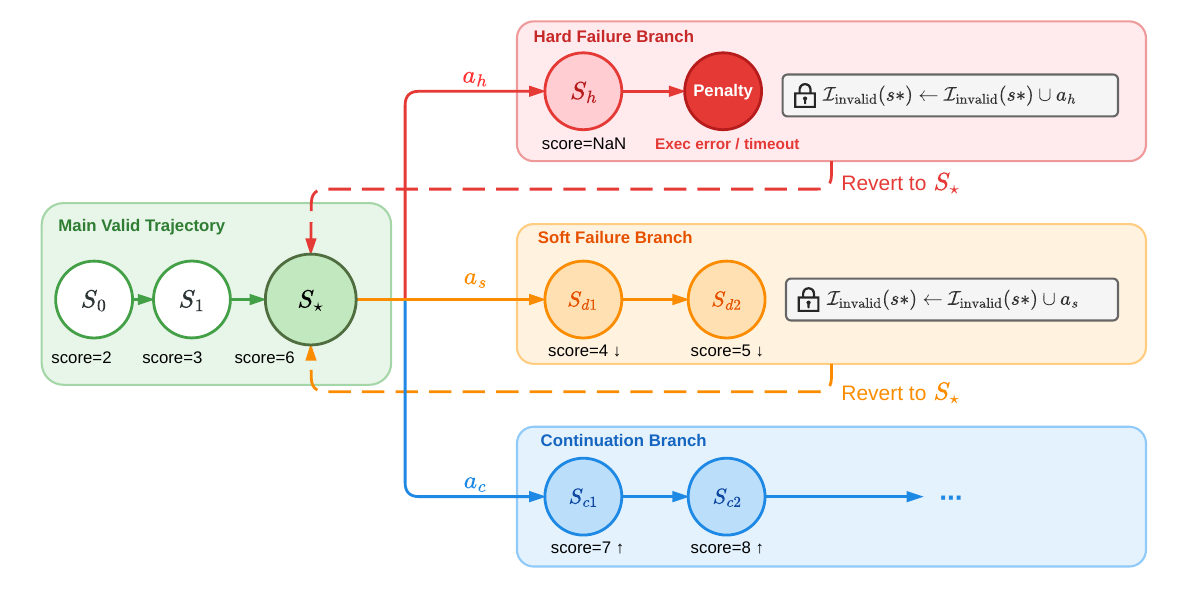}
    \vspace{-0.5em}
    \caption{Diagram of failure-aware trajectory branching and invalid-action masking}
    \label{fig:rl}
    \vspace{-1.0em}
\end{figure}

\textsc{MOSAIC} uses a trajectory branching mechanism to preserve both successful refinements and failed edits as useful supervision for offline RL. 
As shown in Figure~\ref{fig:rl}, all branches share the prefix trajectory up to the last valid checkpoint $s_\star$. At $s_\star$, the state-specific invalid-action mask $\mathcal{I}_{\mathrm{invalid}}(s_\star)$ already contains structurally infeasible or redundant actions before failure-induced updates, such as incompatible hyperparameter edits or no-op edits that set a hyperparameter to its current value.
If an action causes a hard failure, such as an execution error, timeout, or NaN, \textsc{MOSAIC} records an immediate terminal failure branch with a penalty. If an action causes a soft failure, where the code remains executable but the validation score degrades across multiple consecutive steps, \textsc{MOSAIC} records the degraded trajectory as a soft-failure branch. In both cases, the failed action is added to the invalid-action mask $\mathcal{I}_{\mathrm{invalid}}(s_\star)$, and the main rollout reverts to $s_\star$ before continuing along a valid branch. This converts failed refinements into negative training signals while preventing the policy from repeatedly selecting invalid actions.

\section{Dataset Details}
\label{app:datasets}

\begin{table}[h]
  \centering
  \caption{Dataset configurations for forecasting (left) and generation (right).}
  \label{tab:dataset_config}
  \resizebox{\columnwidth}{!}{%
    \begin{tabular}{lcc ccc c cc}
    \toprule
    \rowcolor[HTML]{FFF2CC}
    & & & \multicolumn{4}{c}{\textbf{Forecasting}} & \multicolumn{2}{c}{\textbf{Generation}} \\
    \cmidrule(lr){4-7}\cmidrule(lr){8-9}
    \rowcolor[HTML]{FFF2CC}
    \multirow{-2.5}{*}{\textbf{Dataset}} & \multirow{-2.5}{*}{\textbf{Freq.}} & \multirow{-2.5}{*}{\textbf{Dim.}}
      & \textbf{Seq. / Label / Pred.} & \textbf{Stride (Tr./Te.)} & \textbf{Eval. Targets} & \textbf{Split}
      & \textbf{Window} & \textbf{Split} \\
    \midrule
    Crypto & 1 h & 20 & 72 / 24 / 3 & 12 / 3 & All 20 & 70/15/15 & 60 & 70/15/15 \\
    LOB & 10 s & 14 & 72 / 24 / 3 & 12 / 3 & 5 (OHLCV) & 70/15/15 & 60 & 70/15/15 \\
    Stock & 1 B & 10 & 60 / 20 / 3 & 1 / 3 & All 10 & 70/15/15 & 60 & 70/15/15 \\
    \bottomrule
    \end{tabular}
  }
\end{table}

We evaluate on three financial datasets spanning different asset classes, frequencies, and data regimes. All datasets are split chronologically (70\%/15\%/15\%) to prevent look-ahead bias. Channel-wise z-score normalization is fitted on the training split and applied to all splits. Tables~\ref{tab:dataset_config} summarize task-specific configurations.

\begin{itemize}[nolistsep,leftmargin=10pt]
  \item \textbf{Crypto.} Hourly closing prices for 20 cryptocurrency trading pairs against USDT, covering a one-year period in 2024 (8,784 time steps, 20 channels).
  \item \textbf{LOB.} Limit order book data for a NASDAQ-listed equity, sourced from the LOBSTER platform. Raw event-level data is aggregated into 10-second bars with 14 microstructure features (OHLCV, buy/sell volume, trade counts, order flow statistics, and derived indicators). This dataset represents a challenging low-data regime common in proprietary financial settings. For forecasting, 5 price-related targets (open, high, low, close, VWAP) are evaluated; for generation, all 14 features are modeled.
  \item \textbf{Stock.} Daily closing prices for 10 major U.S. equities from 2020 to 2024 (1,257 business days, 10 channels). All channels serve as targets for both tasks.
\end{itemize}

Rolling windows are constructed with task-specific parameters. Training uses a denser stride to maximize sample utilization, while validation and test use non-overlapping strides to ensure independent evaluation.

\section{Implementation Details}
\label{app:implementation}

\paragraph{Hardware.}
All experiments were conducted on a server equipped with one NVIDIA H200 GPU, an AMD EPYC 9355 32-Core Processor CPU, and 128 GB of RAM.

\paragraph{LLM Configuration.}
The system uses four LLM clients with separate roles: a reasoning LLM for case retrieval, model selection, and EDA interpretation; a code-editing LLM for refinement and fine-tuning modifications; a code-generation LLM for blueprint-based model synthesise; and a VLM for multimodal analysis of EDA plots and training curves. By default, all four use GPT-4o with temperature 0.7 and a maximum output budget of 8,192 tokens. When evaluating with alternative backbones (GPT-5.4, Claude Opus 4, Nova Pro), all four clients are switched to the corresponding model simultaneously. Case retrieval embeddings are computed using all-MiniLM-L6-v2.

\paragraph{Pipeline Configuration.}
The system retrieves $k{=}3$ candidate models via case-based reasoning, with 3 retrieved cases informing model selection. Model generation module produces up to 3 candidate implementations per blueprint, with a maximum of 2 revision rounds for execution-based verification. The RL refinement stage uses 6 warm-up steps and 6 optimisation steps, with a debug retry limit of 3 and an execution timeout of 18,000 seconds per run.

\paragraph{Training and Model Hyperparameters.}
Training hyperparameters are initialized with shared defaults: learning rate $10^{-4}$, batch size 128, maximum 300 epochs with early stopping (patience 20). Model-specific architectural parameters (e.g., $d_{\text{model}}$, number of layers, attention heads) are initialized from default configurations provided with each model in the model bank. Both training and architectural hyperparameters are iteratively adjusted by the LLM during the refinement stage based on execution feedback and validation metrics.

\paragraph{Reinforcement Learning Training Details.}

We train two separate offline IQL agents, one for each task type. For time series forecasting, the agent is trained on 4,456 real refinement transitions collected from 653 episodes across 83 tasks, with a discrete action space of 39 actions (17 structural refinement operations and 22 hyperparameter-tuning bins across 4 slots: learning rate, dropout, layers, batch size). For time series generation, the agent is trained on 2,288 transitions from 509 episodes across 75 tasks, with a discrete action space of 49 actions (17 structural operations and 32 hyperparameter-tuning bins across 6 slots: learning rate, dropout, layers, batch size, hidden dim, and latent dim). Both agents share the same architecture and training configuration: states are encoded as 65-dimensional feature vectors summarizing training dynamics, refinement history, hyperparameter bins, and task context, processed by a shared MLP encoder followed by separate Q-, V-, and policy heads. All networks are optimised with Adam (lr $= 3 \times 10^{-4}$), with weight decay $10^{-4}$ applied to the Q- and V-networks. The IQL hyperparameters are set as follows: expectile level 0.7, AWR temperature 3.0, discount factor 0.95, EMA target update rate 0.005, maximum advantage weight 20, gradient clipping threshold 1.0, maximum advantage weight 20, and gradient clipping 1.0. Rewards are normalised before training, and each agent is trained for 30,000 gradient steps with batch size 256.

\section{Full Experimental Results}
\label{app:full_results}

\subsection{System-Level Forecasting Results}
\label{app:tsf_detailed}

\begin{table}[H]
  \centering
  \caption{System-level forecasting performance on three datasets. Each metric is averaged over five runs. Best result per LLM per dataset is \textbf{bolded}.}
  \label{tab:app_tsf_main}
  \resizebox{\textwidth}{!}{%
    \begin{tabular}{cl cccc cccc cccc cccc cccc}
    \toprule
    \rowcolor[HTML]{FFF2CC}
    & & \multicolumn{4}{c}{\textbf{RMSE $\downarrow$}} & \multicolumn{4}{c}{\textbf{MAE $\downarrow$}} & \multicolumn{4}{c}{\textbf{MAPE (\%) $\downarrow$}} & \multicolumn{4}{c}{\textbf{sMAPE (\%) $\downarrow$}} & \multicolumn{4}{c}{\textbf{Success (\%) $\uparrow$}} \\
    \cmidrule(lr){3-6}\cmidrule(lr){7-10}\cmidrule(lr){11-14}\cmidrule(lr){15-18}\cmidrule(lr){19-22}
    \rowcolor[HTML]{FFF2CC}
    \multirow{-2.5}{*}{\textbf{Dataset}} & \multirow{-2.5}{*}{\textbf{Model}} & GPT-4o & GPT-5.4 & Claude & Nova & GPT-4o & GPT-5.4 & Claude & Nova & GPT-4o & GPT-5.4 & Claude & Nova & GPT-4o & GPT-5.4 & Claude & Nova & GPT-4o & GPT-5.4 & Claude & Nova \\
    \midrule
    \rowcolor[HTML]{D9EAD3}
    \cellcolor{white} \multirow{5}{*}{Crypto}
      & MOSAIC          & \textbf{0.207} & \textbf{0.202} & \textbf{0.217} & \textbf{0.214} & \textbf{0.051} & \textbf{0.051} & \textbf{0.054} & \textbf{0.052} & \textbf{1.543} & \textbf{1.531} & \textbf{1.583} & \textbf{1.535} & \textbf{1.547} & \textbf{1.534} & \textbf{1.591} & \textbf{1.540} & \textbf{100} & \textbf{100} & \textbf{100} & \textbf{100} \\
      & TS-Agent        & 0.224 & 0.220 & 0.218 & 0.219 & 0.056 & 0.055 & \textbf{0.054} & 0.055 & 1.647 & 1.626 & 1.620 & 1.621 & 1.647 & 1.629 & 1.616 & 1.623 & \textbf{100} & \textbf{100} & \textbf{100} & \textbf{100} \\
      & DS-Agent        & 0.230 & 0.223 & 0.258 & 0.232 & 0.061 & 0.059 & 0.075 & 0.072 & 1.965 & 1.945 & 2.519 & 2.215 & 1.957 & 2.050 & 2.485 & 2.215 & 60 & 60 & 60 & 60 \\
      & ResearchAgent   & 0.239 & 0.232 & 0.301 & 0.222 & 0.063 & 0.064 & 0.081 & 0.061 & 1.895 & 1.911 & 2.480 & 2.083 & 1.903 & 1.856 & 2.518 & 2.082 & 60 & 60 & 60 & 60 \\
      & AutoGluon       & \multicolumn{4}{c}{0.223} & \multicolumn{4}{c}{0.055} & \multicolumn{4}{c}{1.661} & \multicolumn{4}{c}{1.668} & \multicolumn{4}{c}{\textbf{100}} \\
    \midrule
    \rowcolor[HTML]{D9EAD3}
    \cellcolor{white} \multirow{5}{*}{LOB}
      & MOSAIC          & \textbf{0.149} & \textbf{0.150} & \textbf{0.116} & \textbf{0.161} & \textbf{0.112} & \textbf{0.112} & \textbf{0.083} & \textbf{0.122} & \textbf{0.051} & \textbf{0.050} & \textbf{0.037} & \textbf{0.055} & \textbf{0.051} & \textbf{0.050} & \textbf{0.037} & \textbf{0.055} & \textbf{100} & \textbf{100} & \textbf{100} & \textbf{100} \\
      & TS-Agent        & 0.154 & 0.158 & 0.143 & 0.174 & 0.116 & 0.115 & 0.101 & 0.130 & 0.053 & 0.052 & 0.045 & 0.058 & 0.053 & 0.052 & 0.045 & 0.058 & \textbf{100} & \textbf{100} & \textbf{100} & \textbf{100} \\
      & DS-Agent        & 0.171 & 0.184 & 0.196 & 0.183 & 0.121 & 0.127 & 0.140 & 0.126 & 0.055 & 0.064 & 0.063 & 0.059 & 0.054 & 0.055 & 0.063 & 0.059 & \textbf{100} & \textbf{100} & 60 & 60 \\
      & ResearchAgent   & 0.205 & 0.204 & 0.203 & 0.192 & 0.147 & 0.145 & 0.145 & 0.142 & 0.066 & 0.068 & 0.065 & 0.061 & 0.066 & 0.068 & 0.065 & 0.061 & 40 & 80 & 80 & 60 \\
      & AutoGluon       & \multicolumn{4}{c}{0.215} & \multicolumn{4}{c}{0.123} & \multicolumn{4}{c}{0.066} & \multicolumn{4}{c}{0.066} & \multicolumn{4}{c}{\textbf{100}} \\
    \midrule
    \rowcolor[HTML]{D9EAD3}
    \cellcolor{white} \multirow{5}{*}{Stock}
      &  MOSAIC          & \textbf{7.847} & \textbf{7.815} & \textbf{7.904} & \textbf{8.064} & \textbf{4.804} & \textbf{4.804} & \textbf{4.843} & \textbf{4.963} & \textbf{1.982} & \textbf{1.981} & \textbf{2.005} & \textbf{2.075} & 1.991 & 1.982 & 1.956 & 2.086 & \textbf{100} & \textbf{100} & \textbf{100} & \textbf{100} \\
      & TS-Agent        & 8.017 & 8.079 & 7.982 & 8.590 & 4.912 & 4.964 & 4.905 & 5.047 & 2.046 & 2.060 & 2.076 & 2.123 & \textbf{1.770} & 2.062 & \textbf{1.765} & \textbf{1.850} & \textbf{100} & \textbf{100} & \textbf{100} & \textbf{100} \\
      & DS-Agent        & 8.557 & 8.445 & 8.559 & 8.732 & 5.193 & 5.169 & 5.150 & 5.207 & 2.137 & 2.288 & 2.177 & 2.244 & 1.969 & 1.986 & 2.055 & 2.099 & 80 & 80 & \textbf{100} & 60 \\
      & ResearchAgent   & 9.410 & 9.321 & 9.570 & 9.791 & 5.677 & 5.588 & 5.738 & 5.910 & 2.498 & 2.624 & 2.420 & 2.590 & 2.053 & 2.623 & 2.238 & 2.331 & 80 & 80 & 80 & 80 \\
      & AutoGluon       & \multicolumn{4}{c}{8.430} & \multicolumn{4}{c}{5.258} & \multicolumn{4}{c}{2.174} & \multicolumn{4}{c}{\textbf{1.890}} & \multicolumn{4}{c}{\textbf{100}} \\
    \bottomrule
    \end{tabular}
  }
\end{table}

\begin{table}[H]
  \centering
  \caption{Forecasting financial metrics on the Crypto dataset. Each metric is averaged over five runs. Best per LLM per dataset is \textbf{bolded}.}
  \label{tab:app_tsf_fin}
  \resizebox{0.85\textwidth}{!}{%
    \begin{tabular}{cl cccc cccc cccc}
    \toprule
    \rowcolor[HTML]{FFF2CC}
    & & \multicolumn{4}{c}{\textbf{$\Delta$Sharpe $\downarrow$}} & \multicolumn{4}{c}{\textbf{$\Delta$VaR $\downarrow$}} & \multicolumn{4}{c}{\textbf{$\Delta$ES $\downarrow$}} \\
    \cmidrule(lr){3-6}\cmidrule(lr){7-10}\cmidrule(lr){11-14}
    \rowcolor[HTML]{FFF2CC}
    \multirow{-2.5}{*}{\textbf{Dataset}} & \multirow{-2.5}{*}{\textbf{Model}} & GPT-4o & GPT-5.4 & Claude & Nova & GPT-4o & GPT-5.4 & Claude & Nova & GPT-4o & GPT-5.4 & Claude & Nova \\
    \midrule
    \rowcolor[HTML]{D9EAD3}
    \cellcolor{white} \multirow{5}{*}{Crypto}
      &  MOSAIC          & \textbf{6.654} & \textbf{4.329} & \textbf{10.687} & \textbf{10.211} & 0.012 & \textbf{0.012} & 0.015 & 0.014 & \textbf{0.014} & 0.015 & 0.016 & \textbf{0.016} \\
      & TS-Agent        & 12.431 & 4.530 & 11.297 & 12.416 & 0.015 & 0.015 & 0.017 & 0.015 & 0.019 & 0.022 & 0.024 & 0.023 \\
      & DS-Agent        & 30.572 & 29.352 & 39.897 & 30.868 & \textbf{0.010} & 0.014 & \textbf{0.013} & 0.015 & \textbf{0.014} & \textbf{0.014} & \textbf{0.013} & 0.017 \\
      & ResearchAgent   & 12.850 & 13.236 & 11.942 & 21.249 & \textbf{0.010} & 0.014 & 0.014 & \textbf{0.013} & 0.015 & 0.015 & 0.019 & 0.017 \\
      & AutoGluon       & \multicolumn{4}{c}{22.639} & \multicolumn{4}{c}{0.022} & \multicolumn{4}{c}{0.031} \\
    \bottomrule
    \end{tabular}
  }
\end{table}

Tables~\ref{tab:app_tsf_main} and~\ref{tab:app_tsf_fin} report the full per-LLM forecasting results. \textsc{MOSAIC} achieves the best or near-best performance across all four LLM backbones on every dataset. The improvement is most pronounced on LOB with Claude, where RMSE drops to 0.116, a 19\% reduction over TS-Agent (0.143) and 46\% over AutoGluon (0.215). Averaged across backbones, LOB RMSE is reduced by 8\%, Crypto by 5\%, and Stock by 3\%. For risk-aware metrics on Crypto, \textsc{MOSAIC} with GPT-4o achieves a $\Delta$Sharpe of 6.65, a 46\% reduction over TS-Agent (12.43); with GPT-5.4, $\Delta$Sharpe reaches 4.33, the lowest overall. DS-Agent and ResearchAgent exhibit $\Delta$Sharpe exceeding 10 across most backbones. \textsc{MOSAIC} maintains a 100\% success rate across all configurations, whereas DS-Agent drops to 60\% on multiple datasets. These improvements are attributable to the model generation module's ability to synthesize task-adapted architectures, the EDA module's grounding of model selection in data characteristics, and the RL policy's non-myopic refinement decisions.

\subsection{System-Level Generation Results}
\label{app:tsg_detailed}

\begin{table}[H]
  \centering
  \caption{System-level generation performance on LOB and Stock. Each metric is averaged over five runs. Best result per LLM per dataset is \textbf{bolded}.}
  \label{tab:app_tsg_main}
  \resizebox{\textwidth}{!}{%
    \begin{tabular}{cl cccc cccc cccc cccc cccc}
    \toprule
    \rowcolor[HTML]{FFF2CC}
    & & \multicolumn{4}{c}{\textbf{Marginal $\downarrow$}} & \multicolumn{4}{c}{\textbf{Correlation $\downarrow$}} & \multicolumn{4}{c}{\textbf{AutoCorrelation $\downarrow$}} & \multicolumn{4}{c}{\textbf{Covariance $\downarrow$}} & \multicolumn{4}{c}{\textbf{Success (\%) $\uparrow$}} \\
    \cmidrule(lr){3-6}\cmidrule(lr){7-10}\cmidrule(lr){11-14}\cmidrule(lr){15-18}\cmidrule(lr){19-22}
    \rowcolor[HTML]{FFF2CC}
    \multirow{-2.5}{*}{\textbf{Dataset}} & \multirow{-2.5}{*}{\textbf{Model}} & GPT-4o & GPT-5.4 & Claude & Nova & GPT-4o & GPT-5.4 & Claude & Nova & GPT-4o & GPT-5.4 & Claude & Nova & GPT-4o & GPT-5.4 & Claude & Nova & GPT-4o & GPT-5.4 & Claude & Nova \\
    \midrule
    \rowcolor[HTML]{D9EAD3}
    \cellcolor{white} \multirow{5}{*}{LOB}
      & MOSAIC          & \textbf{0.590} & \textbf{0.579} & \textbf{0.554} & 0.626 & \textbf{0.216} & \textbf{0.182} & 0.225 & 0.239 & \textbf{0.199} & \textbf{0.247} & \textbf{0.193} & \textbf{0.228} & \textbf{0.181} & \textbf{0.177} & 0.188 & \textbf{0.192} & \textbf{100} & \textbf{100} & \textbf{100} & \textbf{100} \\
      & TS-Agent        & 0.713 & 1.180 & 0.767 & 0.822 & 0.243 & 0.472 & 0.235 & \textbf{0.233} & 0.234 & 0.477 & 0.215 & 0.254 & 0.234 & 0.332 & 0.234 & 0.221 & \textbf{100} & \textbf{100} & \textbf{100} & \textbf{100} \\
      & DS-Agent        & 1.167 & 1.187 & 1.242 & 1.200 & 0.308 & 0.646 & 0.352 & 0.305 & 0.317 & 0.492 & 0.273 & 0.271 & 0.328 & 0.333 & 0.321 & 0.326 & 40 & 60 & \textbf{100} & 20 \\
      & ResearchAgent   & 0.619 & 0.802 & 0.814 & \textbf{0.617} & 0.241 & 0.632 & \textbf{0.138} & 0.243 & 0.240 & 0.611 & 0.244 & 0.243 & 0.239 & 0.340 & \textbf{0.136} & 0.341 & 80 & 80 & \textbf{100} & 20 \\
      & Optuna          & \multicolumn{4}{c}{0.824} & \multicolumn{4}{c}{0.711} & \multicolumn{4}{c}{0.527} & \multicolumn{4}{c}{0.210} & \multicolumn{4}{c}{\textbf{100}} \\
    \midrule
    \rowcolor[HTML]{D9EAD3} 
    \cellcolor{white} \multirow{5}{*}{Stock}
      & MOSAIC          & \textbf{0.408} & \textbf{0.371} & \textbf{0.383} & \textbf{0.385} & \textbf{0.357} & \textbf{0.343} & 0.345 & \textbf{0.378} & \textbf{0.091} & 0.087 & \textbf{0.081} & \textbf{0.081} & \textbf{0.881} & \textbf{0.902} & \textbf{0.880} & \textbf{0.903} & \textbf{100} & \textbf{100} & \textbf{100} & \textbf{100} \\
      & TS-Agent        & 0.455 & 0.681 & 0.421 & 0.473 & 0.397 & 0.437 & 0.361 & 0.401 & 0.129 & \textbf{0.075} & 0.193 & 0.859 & 0.910 & 0.939 & 0.909 & 0.930 & \textbf{100} & \textbf{100} & \textbf{100} & \textbf{100} \\
      & DS-Agent        & 1.371 & 0.774 & 0.427 & 0.482 & 0.397 & 0.594 & 0.472 & 0.422 & 0.208 & 0.442 & 0.106 & 0.568 & 0.944 & 0.942 & 0.913 & 0.912 & \textbf{100} & \textbf{100} & \textbf{100} & 20 \\
      & ResearchAgent   & 0.481 & 0.464 & 0.463 & 0.461 & 0.394 & 0.402 & \textbf{0.305} & 0.404 & 0.093 & 0.805 & 0.263 & 0.664 & 0.916 & 0.917 & 0.915 & 0.926 & \textbf{100} & \textbf{100} & \textbf{100} & 60 \\
      & Optuna          & \multicolumn{4}{c}{0.422} & \multicolumn{4}{c}{0.528} & \multicolumn{4}{c}{0.865} & \multicolumn{4}{c}{0.920} & \multicolumn{4}{c}{\textbf{100}} \\
    \bottomrule
    \end{tabular}
  }
\end{table}

\begin{table}[H]
  \centering
  \caption{Generation financial metrics and success rate on the Crypto dataset. Each metric is averaged over five runs. Best per LLM per dataset is \textbf{bolded}.}
  \label{tab:app_tsg_fin}
  \resizebox{0.85\textwidth}{!}{%
    \begin{tabular}{cl cccc cccc cccc}
    \toprule
    \rowcolor[HTML]{FFF2CC}
    & & \multicolumn{4}{c}{\textbf{$\Delta$VaR $\downarrow$}} & \multicolumn{4}{c}{\textbf{$\Delta$ES $\downarrow$}} & \multicolumn{4}{c}{\textbf{Success (\%) $\uparrow$}} \\
    \cmidrule(lr){3-6}\cmidrule(lr){7-10}\cmidrule(lr){11-14}
    \rowcolor[HTML]{FFF2CC}
    \multirow{-2.5}{*}{\textbf{Dataset}} & \multirow{-2.5}{*}{\textbf{Model}} & GPT-4o & GPT-5.4 & Claude & Nova & GPT-4o & GPT-5.4 & Claude & Nova & GPT-4o & GPT-5.4 & Claude & Nova \\
    \midrule
    \rowcolor[HTML]{D9EAD3} 
    \cellcolor{white} \multirow{5}{*}{Crypto}
      & MOSAIC          & 0.072 & 0.073 & \textbf{0.075} & \textbf{0.078} & \textbf{0.087} & 0.096 & \textbf{0.104} & \textbf{0.095} & \textbf{100} & \textbf{100} & \textbf{100} & \textbf{100} \\
      & TS-Agent        & \textbf{0.068} & \textbf{0.044} & 0.084 & 0.091 & 0.103 & \textbf{0.054} & 0.111 & 0.104 & \textbf{100} & \textbf{100} & \textbf{100} & \textbf{100} \\
      & DS-Agent        & 0.103 & 0.096 & 0.119 & 0.096 & 0.148 & 0.147 & 0.152 & 0.130 & 80 & 80 & \textbf{100} & 40 \\
      & ResearchAgent   & 0.093 & 0.094 & 0.099 & 0.103 & 0.105 & 0.105 & 0.134 & 0.105 & 80 & 80 & \textbf{100} & 40 \\
      & Optuna          & \multicolumn{4}{c}{0.091} & \multicolumn{4}{c}{0.126} & \multicolumn{4}{c}{\textbf{100}} \\
    \bottomrule
    \end{tabular}
  }
\end{table}

Tables~\ref{tab:app_tsg_main} and~\ref{tab:app_tsg_fin} present the full generation results. \textsc{MOSAIC} achieves the lowest Marginal and Correlation distances on both LOB and Stock across most backbones. On LOB with GPT-5.4, Marginal distance drops to 0.579 compared to 1.180 for TS-Agent and 1.187 for DS-Agent, a reduction exceeding 50\%. Averaged across backbones, LOB Marginal is reduced by 32\% and Correlation by 27\% relative to TS-Agent. On Stock, \textsc{MOSAIC} with GPT-5.4 achieves the lowest Marginal (0.371) and Correlation (0.343) distances, with average Marginal reduced by 24\%. For Crypto tail-risk metrics, \textsc{MOSAIC} achieves competitive $\Delta$VaR and the lowest $\Delta$ES on three of four backbones, though TS-Agent obtains lower $\Delta$VaR with GPT-4o (0.068 vs.\ 0.072) and GPT-5.4 (0.044 vs.\ 0.073). Throughout, \textsc{MOSAIC} maintains a 100\% success rate, while DS-Agent and ResearchAgent drop as low as 20\% on LOB and Stock with weaker backbones. The consistent advantage across distributional metrics reflects the model generation module's capacity to compose generative architectures tailored to each dataset's statistical properties.

\subsection{Model Generation Module Comparison}
\label{app:cg_detailed}

\begin{table}[H]
  \centering
  \caption{Model generation module comparison: forecasting metrics. Each metric is averaged over five runs. Best per LLM per dataset is \textbf{bolded}.}
  \label{tab:app_cg_tsf}
  \resizebox{\textwidth}{!}{%
    \begin{tabular}{cl cccc cccc cccc cccc cccc}
    \toprule
    \rowcolor[HTML]{FFF2CC}
    & & \multicolumn{4}{c}{\textbf{RMSE $\downarrow$}} & \multicolumn{4}{c}{\textbf{MAE $\downarrow$}} & \multicolumn{4}{c}{\textbf{MAPE (\%) $\downarrow$}} & \multicolumn{4}{c}{\textbf{sMAPE (\%) $\downarrow$}} & \multicolumn{4}{c}{\textbf{Win (\%) $\uparrow$}} \\
    \cmidrule(lr){3-6}\cmidrule(lr){7-10}\cmidrule(lr){11-14}\cmidrule(lr){15-18}\cmidrule(lr){19-22}
    \rowcolor[HTML]{FFF2CC}
    \multirow{-2.5}{*}{\textbf{Dataset}} & \multirow{-2.5}{*}{\textbf{Model}} & GPT-4o & GPT-5.4 & Claude & Nova & GPT-4o & GPT-5.4 & Claude & Nova & GPT-4o & GPT-5.4 & Claude & Nova & GPT-4o & GPT-5.4 & Claude & Nova & GPT-4o & GPT-5.4 & Claude & Nova\\
    \midrule
    \rowcolor[HTML]{D9EAD3}
    \cellcolor{white} \multirow{5}{*}{Crypto}
      &  ModelGen (MOSAIC)          & \textbf{0.205} & \textbf{0.202} & 0.219 & \textbf{0.205} & \textbf{0.051} & \textbf{0.051} & 0.055 & \textbf{0.051} & \textbf{1.530} & \textbf{1.514} & \textbf{1.615} & \textbf{1.523} & \textbf{1.531} & \textbf{1.519} & 1.621 & \textbf{1.525} & \textbf{60} & \textbf{80} & \textbf{60} & \textbf{60} \\
      & AlphaEvolve     & 0.218 & 0.218 & 0.215 & 0.218 & 0.055 & 0.055 & \textbf{0.054} & 0.054 & 1.627 & 1.610 & 1.619 & 1.609 & 1.629 & 1.611 & 1.621 & 1.610 & \textbf{60} & 60 & 40 & 40 \\
      & AIDE            & 0.218 & 0.217 & 0.218 & 0.218 & 0.055 & 0.054 & \textbf{0.054} & 0.055 & 1.611 & 1.614 & \textbf{1.615} & 1.621 & 1.611 & 1.612 & \textbf{1.615} & 1.623 & 40 & 60 & 40 & 20 \\
      & EffiLearner     & 0.218 & 0.218 & \textbf{0.213} & 0.217 & 0.054 & 0.054 & \textbf{0.054} & 0.055 & 1.617 & 1.616 & 1.625 & 1.614 & 1.617 & 1.616 & 1.625 & 1.614 & 40 & 60 & 40 & 20 \\
      & LLM4EFFI        & 0.217 & 0.218 & 0.219 & 0.217 & 0.054 & 0.055 & 0.055 & 0.055 & 1.619 & 1.620 & 1.621 & 1.613 & 1.620 & 1.620 & 1.620 & 1.615 & 40 & 60 & 40 & 40 \\
    \midrule
    \rowcolor[HTML]{D9EAD3}
    \cellcolor{white} \multirow{5}{*}{LOB}
      &  ModelGen (MOSAIC)          & \textbf{0.153} & \textbf{0.155} & \textbf{0.095} & \textbf{0.160} & \textbf{0.115} & \textbf{0.115} & \textbf{0.067} & 0.119 & \textbf{0.052} & \textbf{0.051} & \textbf{0.030} & 0.054 & \textbf{0.052} & \textbf{0.052} & \textbf{0.030} & 0.054 & \textbf{60} & \textbf{60} & \textbf{60} & \textbf{60} \\
      & AlphaEvolve     & 0.158 & 0.162 & 0.167 & 0.164 & 0.116 & 0.120 & 0.119 & 0.121 & \textbf{0.052} & 0.054 & 0.054 & 0.053 & \textbf{0.052} & 0.054 & 0.054 & 0.053 & \textbf{60} & 40 & \textbf{60} & 40 \\
      & AIDE            & 0.171 & 0.162 & 0.211 & 0.166 & 0.122 & 0.120 & 0.153 & 0.121 & 0.055 & 0.054 & 0.069 & 0.060 & 0.055 & 0.054 & 0.069 & 0.060 & 40 & 40 & 40 & 40 \\
      & EffiLearner     & 0.190 & 0.158 & 0.200 & 0.162 & 0.136 & 0.117 & 0.145 & \textbf{0.115} & 0.061 & 0.053 & 0.065 & \textbf{0.052} & 0.061 & 0.053 & 0.065 & \textbf{0.052} & 20 & \textbf{60} & \textbf{60} & 20 \\
      & LLM4EFFI        & 0.168 & 0.157 & 0.197 & 0.162 & 0.121 & 0.116 & 0.146 & 0.120 & 0.054 & 0.052 & 0.066 & 0.054 & 0.054 & 0.053 & 0.065 & 0.054 & 40 & 40 & \textbf{60} & 20 \\
    \midrule
    \rowcolor[HTML]{D9EAD3}
    \cellcolor{white} \multirow{5}{*}{Stock}
      &  ModelGen (MOSAIC)        & \textbf{7.805} & \textbf{7.769} & \textbf{7.867} & \textbf{7.924} & \textbf{4.774} & \textbf{4.790} & \textbf{4.840} & \textbf{4.910} & \textbf{1.967} & \textbf{1.977} & \textbf{2.017} & \textbf{2.045} & \textbf{1.977} & \textbf{1.979} & \textbf{1.939} & 2.064 & \textbf{80} & \textbf{80} & \textbf{60} & \textbf{60} \\
      & AlphaEvolve     & 8.554 & 8.254 & 8.225 & 8.194 & 5.273 & 5.072 & 5.074 & 5.047 & 2.156 & 2.098 & 2.091 & 2.091 & 2.168 & 2.108 & 2.101 & 2.096 & 40 & 40 & 40 & 40 \\
      & AIDE            & 7.920 & 7.929 & 8.035 & 8.190 & 4.865 & 4.885 & 4.982 & 5.020 & 2.018 & 2.009 & 2.038 & 2.073 & 2.022 & 2.020 & 2.053 & 2.083 & 40 & 40 & 40 & 20 \\
      & EffiLearner     & 8.092 & 8.076 & 7.936 & 8.208 & 4.986 & 4.989 & 4.858 & 5.043 & 2.064 & 2.043 & 2.018 & 2.079 & 2.069 & 2.057 & 2.015 & 2.088 & 40 & 60 & 40 & 20 \\
      & LLM4EFFI        & 7.864 & 7.876 & 8.058 & 8.012 & 4.818 & 4.825 & 4.973 & 4.936 & 2.000 & 1.990 & 2.053 & 2.050 & 2.004 & 1.999 & 2.060 & \textbf{2.057} & 60 & 60 & \textbf{60} & \textbf{60} \\
    \bottomrule
    \end{tabular}
  }
\end{table}
 
\begin{table}[H]
  \centering
  \caption{Model generation comparison: Crypto financial metrics (forecasting). Each metric is averaged over five runs. Best per LLM per dataset is \textbf{bolded}.}
  \label{tab:app_cg_tsf_fin}
  \resizebox{0.85\textwidth}{!}{%
    \begin{tabular}{cl cccc cccc cccc}
    \toprule
    \rowcolor[HTML]{FFF2CC}
    & & \multicolumn{4}{c}{\textbf{$\Delta$Sharpe $\downarrow$}} & \multicolumn{4}{c}{\textbf{$\Delta$VaR $\downarrow$}} & \multicolumn{4}{c}{\textbf{$\Delta$ES $\downarrow$}} \\
    \cmidrule(lr){3-6}\cmidrule(lr){7-10}\cmidrule(lr){11-14}
    \rowcolor[HTML]{FFF2CC}
    \multirow{-2.5}{*}{\textbf{Dataset}} & \multirow{-2.5}{*}{\textbf{Model}} & GPT-4o & GPT-5.4 & Claude & Nova & GPT-4o & GPT-5.4 & Claude & Nova & GPT-4o & GPT-5.4 & Claude & Nova \\
    \midrule
    \rowcolor[HTML]{D9EAD3}
    \cellcolor{white} \multirow{5}{*}{Crypto}
      &  ModelGen (MOSAIC)          & \textbf{2.074} & \textbf{2.431} & 10.924 & 9.477 & \textbf{0.010} & \textbf{0.011} & \textbf{0.014} & \textbf{0.012} & \textbf{0.011} & \textbf{0.013} & \textbf{0.014} & \textbf{0.014} \\
      & AlphaEvolve     & 6.949 & 7.370 & 27.326 & 7.219 & 0.014 & 0.014 & 0.020 & 0.016 & 0.020 & 0.020 & 0.028 & 0.024 \\
      & AIDE            & 5.898 & 5.890 & \textbf{4.009} & 18.636 & 0.015 & 0.017 & 0.017 & 0.016 & 0.022 & 0.025 & 0.025 & 0.024 \\
      & EffiLearner     & 9.715 & 6.947 & 24.718 & \textbf{2.571} & 0.016 & 0.016 & 0.020 & 0.016 & 0.024 & 0.024 & 0.028 & 0.023 \\
      & LLM4EFFI        & 8.211 & 3.552 & 8.842 & 5.962 & 0.015 & 0.016 & 0.017 & 0.016 & 0.023 & 0.024 & 0.024 & 0.023 \\
    \bottomrule
    \end{tabular}
  }
\end{table}
 
\begin{table}[H]
  \centering
  \caption{Model generation comparison: generation metrics. Each metric is averaged over five runs. Best per LLM per dataset is \textbf{bolded}.}
  \label{tab:app_cg_tsg}
  \resizebox{\textwidth}{!}{%
    \begin{tabular}{cl cccc cccc cccc cccc cccc}
    \toprule
    \rowcolor[HTML]{FFF2CC}
    & & \multicolumn{4}{c}{\textbf{Marginal $\downarrow$}} & \multicolumn{4}{c}{\textbf{Correlation $\downarrow$}} & \multicolumn{4}{c}{\textbf{AutoCorrelation $\downarrow$}} & \multicolumn{4}{c}{\textbf{Covariance $\downarrow$}} & \multicolumn{4}{c}{\textbf{Win(\%) $\uparrow$}} \\
    \cmidrule(lr){3-6}\cmidrule(lr){7-10}\cmidrule(lr){11-14}\cmidrule(lr){15-18}\cmidrule(lr){19-22}
    \rowcolor[HTML]{FFF2CC}
    \multirow{-2.5}{*}{\textbf{Dataset}} & \multirow{-2.5}{*}{\textbf{Model}} & GPT-4o & GPT-5.4 & Claude & Nova & GPT-4o & GPT-5.4 & Claude & Nova & GPT-4o & GPT-5.4 & Claude & Nova & GPT-4o & GPT-5.4 & Claude & Nova & GPT-4o & GPT-5.4 & Claude & Nova\\
    \midrule
    \rowcolor[HTML]{D9EAD3}
    \cellcolor{white} \multirow{5}{*}{LOB}
      & ModelGen (MOSAIC)          & \textbf{0.556} & \textbf{0.536} & \textbf{0.513} & \textbf{0.582} & \textbf{0.195} & \textbf{0.168} & \textbf{0.214} & \textbf{0.206} & \textbf{0.194} & 0.212 & \textbf{0.191} & \textbf{0.233} & \textbf{0.169} & \textbf{0.162} & \textbf{0.187} & \textbf{0.186} & \textbf{60} & \textbf{80} & \textbf{60} & \textbf{60} \\
      & AlphaEvolve     & 1.142 & 0.668 & 0.785 & 0.734 & 0.375 & 0.256 & 0.265 & 0.252 & 0.579 & 0.175 & 0.201 & 0.497 & 0.298 & 0.229 & 0.228 & 0.229 & 40 & 40 & \textbf{60} & 40 \\
      & AIDE            & 1.534 & 0.709 & 0.725 & 0.588 & 1.059 & 0.188 & 0.237 & 0.209 & 0.614 & 0.099 & 0.250 & 0.253 & 0.336 & 0.179 & 0.197 & 0.195 & 20 & 40 & 40 & 40 \\
      & EffiLearner     & 0.570 & 0.680 & 1.504 & 1.380 & 0.220 & 0.185 & 0.286 & 0.371 & 0.215 & 0.102 & 0.198 & 0.442 & 0.198 & 0.171 & 0.335 & 0.329 & 20 & 60 & \textbf{60} & 40 \\
      & LLM4EFFI        & 0.728 & 0.646 & 1.558 & 1.403 & 0.216 & 0.185 & 0.286 & 0.369 & 0.197 & \textbf{0.083} & 0.195 & 0.362 & 0.195 & 0.174 & 0.333 & 0.328 & 40 & 60 & \textbf{60} & 20 \\
    \midrule
    \rowcolor[HTML]{D9EAD3}
    \cellcolor{white} \multirow{5}{*}{Stock}
      & ModelGen (MOSAIC)          & 0.362 & \textbf{0.326} & \textbf{0.382} & \textbf{0.350} & \textbf{0.364} & \textbf{0.339} & \textbf{0.360} & \textbf{0.374} & \textbf{0.090} & \textbf{0.091} & \textbf{0.078} & \textbf{0.073} & 0.894 & 0.893 & \textbf{0.911} & 0.923 & \textbf{60} & \textbf{80} & \textbf{60} & \textbf{60} \\
      & AlphaEvolve     & \textbf{0.354} & 0.356 & 0.406 & 0.411 & 0.413 & 0.382 & 0.361 & 0.446 & 0.093 & 0.094 & 0.079 & 0.822 & \textbf{0.870} & \textbf{0.873} & 0.918 & 0.875 & 40 & 40 & 40 & 40 \\
      & AIDE            & 0.380 & 0.363 & 0.646 & 0.421 & 0.482 & 0.401 & 0.514 & 0.444 & 0.093 & 0.094 & 0.097 & 0.876 & 0.901 & 0.894 & 0.939 & 0.883 & 20 & 40 & 40 & 20 \\
      & EffiLearner     & 0.364 & 0.356 & 0.411 & 0.652 & 0.402 & 0.382 & 0.381 & 0.573 & 0.094 & 0.094 & 0.273 & 0.081 & 0.894 & \textbf{0.873} & \textbf{0.911} & 0.886 & 40 & 60 & 40 & 20 \\
      & LLM4EFFI        & 0.362 & 0.452 & 0.444 & 0.826 & 0.393 & 0.370 & 0.526 & 0.655 & 0.094 & 0.237 & 0.092 & 0.080 & 0.894 & 0.929 & 0.916 & \textbf{0.328} & 40 & 40 & 40 & 40 \\
    \bottomrule
    \end{tabular}
  }
\end{table}
 
\begin{table}[H]
  \centering
  \caption{Model generation comparison: Crypto financial metrics (generation). Each metric is averaged over five runs. Best per LLM per dataset is \textbf{bolded}.}
  \label{tab:app_cg_tsg_fin}
  \resizebox{0.85\textwidth}{!}{%
    \begin{tabular}{cl cccc cccc cccc}
    \toprule
    \rowcolor[HTML]{FFF2CC}
    & & \multicolumn{4}{c}{\textbf{$\Delta$VaR $\downarrow$}} & \multicolumn{4}{c}{\textbf{$\Delta$ES $\downarrow$}} & \multicolumn{4}{c}{\textbf{Win (\%) $\uparrow$}} \\
    \cmidrule(lr){3-6}\cmidrule(lr){7-10}\cmidrule(lr){11-14}
    \rowcolor[HTML]{FFF2CC}
    \multirow{-2.5}{*}{\textbf{Dataset}} & \multirow{-2.5}{*}{\textbf{Model}} & GPT-4o & GPT-5.4 & Claude & Nova & GPT-4o & GPT-5.4 & Claude & Nova & GPT-4o & GPT-5.4 & Claude & Nova \\
    \midrule
    \rowcolor[HTML]{D9EAD3}
    \cellcolor{white} \multirow{5}{*}{Crypto}
      & ModelGen (MOSAIC)        & \textbf{0.081} & \textbf{0.082} & \textbf{0.081} & \textbf{0.082} & \textbf{0.082} & \textbf{0.103} & \textbf{0.105} & \textbf{0.102} & \textbf{60} & \textbf{60} & \textbf{60} & \textbf{60} \\
      & AlphaEvolve     & 0.111 & 0.110 & 0.089 & 0.213 & 0.125 & 0.125 & 0.110 & 0.213 & 40 & 40 & 40 & 20 \\
      & AIDE            & 0.233 & 0.108 & 0.112 & 0.105 & 0.235 & 0.117 & 0.125 & 0.131 & 40 & \textbf{60} & 40 & 20 \\
      & EffiLearner     & 0.112 & 0.109 & 0.120 & 0.114 & 0.115 & 0.117 & 0.122 & 0.152 & 40 & \textbf{60} & 40 & 20 \\
      & LLM4EFFI        & 0.106 & 0.110 & 0.086 & 0.159 & 0.109 & 0.123 & 0.109 & 0.178 & 40 & \textbf{60} & 40 & 20 \\
    \bottomrule
    \end{tabular}
  }
\end{table}

Tables~\ref{tab:app_cg_tsf}--~\ref{tab:app_cg_tsg_fin} present the full per-LLM code generation module comparison. On forecasting, \textsc{MOSAIC}’s model generation module (ModelGen) achieves the lowest RMSE on LOB with Claude (0.095), outperforming the next best alternative by over 39\%. On Crypto, it achieves the lowest Sharpe Ratio Difference (2.07 with GPT-4o) while all competitors exceed 3.5. On generation, ModelGen leads on LOB across all four metrics and all backbones, with Marginal distances 20–55\% lower than alternatives. Win Rate further confirms these advantages: ModelGen achieves 60–80\% Win Rate across datasets and backbones, while alternatives rarely exceed 40\%. The structured blueprint with shape contracts enables principled module recombination that outperforms unconstrained evolutionary, tree-based, and feedback-driven search strategies.

\subsection{Reinforcement Learning Comparison}
\label{app:rl_comparison}

\begin{table}[H]
  \centering
  \caption{RL module comparison: forecasting metrics. Each metric is averaged over five runs. Best per LLM per dataset is \textbf{bolded}.}
  \label{tab:app_rl_tsf}
  \resizebox{\textwidth}{!}{%
    \begin{tabular}{cl cccc cccc cccc cccc cccc}
    \toprule
    \rowcolor[HTML]{FFF2CC}
    & & \multicolumn{4}{c}{\textbf{RMSE $\downarrow$}} & \multicolumn{4}{c}{\textbf{MAE $\downarrow$}} & \multicolumn{4}{c}{\textbf{MAPE $\downarrow$}} & \multicolumn{4}{c}{\textbf{sMAPE $\downarrow$}} & \multicolumn{4}{c}{\textbf{Steps to Best Incumbent $\downarrow$}} \\
    \cmidrule(lr){3-6}\cmidrule(lr){7-10}\cmidrule(lr){11-14}\cmidrule(lr){15-18}\cmidrule(lr){19-22}
    \rowcolor[HTML]{FFF2CC}
    \multirow{-2.5}{*}{\textbf{Dataset}} & \multirow{-2.5}{*}{\textbf{Model}} & GPT-4o & GPT-5.4 & Claude & Nova & GPT-4o & GPT-5.4 & Claude & Nova & GPT-4o & GPT-5.4 & Claude & Nova & GPT-4o & GPT-5.4 & Claude & Nova & GPT-4o & GPT-5.4 & Claude & Nova\\
    \midrule
    \rowcolor[HTML]{D9EAD3}
    \cellcolor{white} \multirow{4}{*}{Crypto}
      & IQL (MOSAIC) & \textbf{0.209} & \textbf{0.203} & \textbf{0.214} & 0.227 & \textbf{0.053} & \textbf{0.053} & \textbf{0.052} & \textbf{0.053} & \textbf{1.563} & \textbf{1.596} & \textbf{1.536} & \textbf{1.552} & \textbf{1.571} & 1.596 & \textbf{1.545} & \textbf{1.563} & \textbf{8.8} & 8.4 & 8.6 & \textbf{8.6} \\
      & BC & 0.220 & 0.215 & 0.235 & 0.236 & 0.060 & 0.054 & 0.056 & 0.056 & 1.629 & 1.611 & 1.685 & 1.612 & 1.638 & 1.612 & 1.677 & 1.627 & \textbf{8.8} & \textbf{7.8} & 9.6 & 9.4 \\
      & DQN & 0.216 & 0.215 & 0.218 & 0.233 & 0.056 & \textbf{0.053} & 0.055 & 0.055 & 1.610 & 1.599 & 1.582 & 1.591 & 1.622 & \textbf{1.586} & 1.597 & 1.606 & 9.0 & 9.0 & \textbf{8.4} & 9.0 \\
      & LLM & 0.221 & 0.215 & 0.219 & \textbf{0.220} & 0.058 & 0.054 & 0.054 & 0.057 & 1.687 & 1.609 & 1.630 & 1.630 & 1.671 & 1.609 & 1.628 & 1.624 & 10.0 & 9.6 & 9.8 & 9.4 \\
    \midrule
    \rowcolor[HTML]{D9EAD3}
    \cellcolor{white} \multirow{4}{*}{LOB}
      &  IQL (MOSAIC) & \textbf{0.144} & \textbf{0.143} & 0.147 & \textbf{0.162} & \textbf{0.108} & \textbf{0.108} & \textbf{0.107} & \textbf{0.127} & \textbf{0.050} & \textbf{0.049} & \textbf{0.048} & 0.058 & \textbf{0.050} & \textbf{0.049} & \textbf{0.048} & 0.058 & \textbf{8.8} & \textbf{8.6} & \textbf{8.4} & \textbf{9.0} \\
      & BC & 0.159 & 0.155 & \textbf{0.146} & 0.176 & 0.111 & 0.109 & 0.109 & 0.140 & 0.057 & \textbf{0.049} & 0.051 & 0.058 & 0.057 & \textbf{0.049} & 0.051 & 0.058 & 9.6 & 9.8 & 9.2 & 10.2 \\
      & DQN & 0.160 & 0.151 & 0.148 & 0.172 & 0.113 & 0.110 & 0.108 & 0.135 & 0.056 & \textbf{0.049} & 0.050 & \textbf{0.057} & 0.056 & \textbf{0.049} & 0.050 & \textbf{0.057} & 9.4 & 9.4 & 9.0 & 9.8 \\
      & LLM & 0.154 & 0.145 & 0.150 & 0.173 & 0.117 & 0.110 & 0.109 & 0.145 & 0.054 & \textbf{0.049} & 0.053 & 0.060 & 0.054 & \textbf{0.049} & 0.053 & 0.060 & 10.0 & 10.6 & 9.6 & 10.4 \\
    \midrule
    \rowcolor[HTML]{D9EAD3}
    \cellcolor{white} \multirow{4}{*}{Stock}
      & IQL (MOSAIC) & \textbf{8.013} & \textbf{8.001} & \textbf{7.958} & \textbf{8.275} & \textbf{4.922} & \textbf{4.859} & \textbf{4.847} & \textbf{5.042} & \textbf{2.042} & \textbf{1.997} & \textbf{1.986} & \textbf{2.120} & \textbf{2.049} & \textbf{1.995} & \textbf{1.982} & 2.118 & \textbf{8.8} & \textbf{8.4} & \textbf{8.2} & \textbf{9.2} \\
      & BC & 8.139 & 8.136 & 8.072 & 8.389 & 5.085 & 5.052 & 4.944 & 5.107 & 2.094 & 2.079 & 2.040 & 2.260 & 2.116 & 2.093 & 2.053 & 2.273 & 9.6 & 9.6 & 9.4 & 10.0 \\
      & DQN & 8.104 & 8.067 & 8.114 & 8.467 & 5.019 & 5.001 & 4.981 & 5.165 & 2.069 & 2.078 & 2.062 & 2.205 & 2.080 & 2.086 & 2.097 & 2.187 & 9.4 & 10.2 & 9.6 & 9.8 \\
      & LLM & 8.142 & 8.173 & 8.095 & 8.446 & 5.079 & 5.082 & 4.957 & 5.132 & 2.083 & 2.091 & 2.058 & 2.241 & 2.104 & 2.111 & 2.071 & \textbf{2.113} & 9.6 & 9.8 & 10.0 & 10.6 \\
    \bottomrule
    \end{tabular}
  }
\end{table}

\begin{table}[H]
  \centering
  \caption{RL module comparison: Crypto financial metrics (forecasting). Each metric is averaged over five runs. Best per LLM per dataset is \textbf{bolded}.}
  \label{tab:app_rl_tsf_fin}
  \resizebox{0.85\textwidth}{!}{%
    \begin{tabular}{cl cccc cccc cccc}
    \toprule
    \rowcolor[HTML]{FFF2CC}
    & & \multicolumn{4}{c}{\textbf{$\Delta$Sharpe $\downarrow$}} & \multicolumn{4}{c}{\textbf{$\Delta$VaR $\downarrow$}} & \multicolumn{4}{c}{\textbf{$\Delta$ES $\downarrow$}} \\
    \cmidrule(lr){3-6}\cmidrule(lr){7-10}\cmidrule(lr){11-14}
    \rowcolor[HTML]{FFF2CC}
    \multirow{-2.5}{*}{\textbf{Dataset}} & \multirow{-2.5}{*}{\textbf{Model}} & GPT-4o & GPT-5.4 & Claude & Nova & GPT-4o & GPT-5.4 & Claude & Nova & GPT-4o & GPT-5.4 & Claude & Nova\\
    \midrule
    \rowcolor[HTML]{D9EAD3}
    \cellcolor{white} \multirow{4}{*}{Crypto}
      & IQL (MOSAIC) & 13.524 & \textbf{11.924} & \textbf{10.331} & \textbf{11.313} & \textbf{0.015} & 0.018 & 0.017 & 0.016 & \textbf{0.019} & \textbf{0.020} & \textbf{0.019} & \textbf{0.019} \\
      & BC & 15.935 & 16.437 & 12.683 & 15.678 & 0.016 & 0.017 & 0.017 & 0.017 & 0.024 & 0.025 & 0.026 & 0.025 \\
      & DQN & \textbf{12.488} & 12.797 & 13.992 & 17.989 & 0.017 & 0.018 & \textbf{0.016} & 0.016 & 0.023 & 0.025 & 0.024 & 0.024 \\
      & LLM & 14.948 & 13.934 & 11.846 & 12.482 & 0.017 & \textbf{0.016} & 0.017 & \textbf{0.015} & 0.022 & 0.024 & 0.027 & 0.025 \\
    \bottomrule
    \end{tabular}
  }
\end{table}

\begin{table}[H]
  \centering
  \caption{RL module comparison: generation metrics. Each metric is averaged over five runs. Best per LLM per dataset is \textbf{bolded}.}
  \label{tab:app_rl_tsg}
  \resizebox{\textwidth}{!}{%
    \begin{tabular}{cl cccc cccc cccc cccc cccc}
    \toprule
    \rowcolor[HTML]{FFF2CC}
    & & \multicolumn{4}{c}{\textbf{Marginal $\downarrow$}} & \multicolumn{4}{c}{\textbf{Correlation $\downarrow$}} & \multicolumn{4}{c}{\textbf{AutoCorrelation $\downarrow$}} & \multicolumn{4}{c}{\textbf{Covariance $\downarrow$}} & \multicolumn{4}{c}{\textbf{Steps to Best Incumbent $\downarrow$}} \\
    \cmidrule(lr){3-6}\cmidrule(lr){7-10}\cmidrule(lr){11-14}\cmidrule(lr){15-18}\cmidrule(lr){19-22}
    \rowcolor[HTML]{FFF2CC}
    \multirow{-2.5}{*}{\textbf{Dataset}} & \multirow{-2.5}{*}{\textbf{Model}} & GPT-4o & GPT-5.4 & Claude & Nova & GPT-4o & GPT-5.4 & Claude & Nova & GPT-4o & GPT-5.4 & Claude & Nova & GPT-4o & GPT-5.4 & Claude & Nova & GPT-4o & GPT-5.4 & Claude & Nova\\
    \midrule
    \rowcolor[HTML]{D9EAD3}
    \cellcolor{white} \multirow{4}{*}{LOB}
      & IQL (MOSAIC) & \textbf{0.645} & \textbf{0.754} & \textbf{0.621} & \textbf{0.692} & \textbf{0.248} & \textbf{0.238} & \textbf{0.241} & 0.285 & \textbf{0.206} & 0.384 & \textbf{0.195} & \textbf{0.216} & \textbf{0.201} & \textbf{0.231} & \textbf{0.192} & \textbf{0.204} & \textbf{8.8} & \textbf{9.0} & 8.6 & \textbf{9.0} \\
      & BC & 0.757 & 0.902 & 0.736 & 0.814 & 0.298 & 0.280 & 0.291 & 0.286 & 0.248 & 0.470 & 0.240 & 0.260 & 0.236 & 0.336 & 0.230 & 0.242 & 9.8 & 10.2 & \textbf{8.4} & 10.0 \\
      & DQN & 0.719 & 0.957 & 0.708 & 0.774 & 0.282 & 0.295 & 0.322 & \textbf{0.272} & 0.282 & 0.450 & 0.226 & 0.247 & 0.267 & 0.325 & 0.217 & 0.228 & 9.4 & 9.8 & 9.2 & 9.6 \\
      & LLM & 0.798 & 0.868 & 0.781 & 0.847 & 0.256 & 0.294 & 0.303 & 0.280 & 0.261 & \textbf{0.274} & 0.251 & 0.264 & 0.249 & 0.267 & 0.247 & 0.248 & 10.2 & 10.4 & 10.0 & 10.6 \\
    \midrule
    \rowcolor[HTML]{D9EAD3}
    \cellcolor{white} \multirow{4}{*}{Stock}
      & IQL (MOSAIC) & 0.480 & \textbf{0.553} & \textbf{0.388} & \textbf{0.440} & \textbf{0.349} & \textbf{0.365} & \textbf{0.326} & 0.386 & \textbf{0.088} & \textbf{0.070} & \textbf{0.082} & \textbf{0.092} & \textbf{0.864} & 0.944 & \textbf{0.838} & \textbf{0.868} & \textbf{8.8} & \textbf{8.6} & \textbf{8.4} & \textbf{9.0} \\
      & BC & 0.492 & 0.646 & 0.455 & 0.560 & 0.437 & 0.420 & 0.381 & 0.398 & 0.114 & 0.104 & 0.104 & 0.138 & 0.918 & 0.915 & 0.902 & 0.957 & 9.8 & 9.6 & 9.4 & 10.0 \\
      & DQN & \textbf{0.473} & 0.626 & 0.487 & 0.499 & 0.390 & 0.400 & 0.373 & \textbf{0.374} & 0.111 & 0.088 & 0.098 & 0.127 & 0.905 & \textbf{0.899} & 0.884 & 0.930 & 9.4 & 9.2 & 9.0 & 9.6 \\
      & LLM & 0.512 & 0.596 & 0.472 & 0.548 & 0.429 & 0.441 & 0.373 & 0.411 & 0.123 & 0.100 & 0.112 & 0.984 & 0.960 & 0.950 & 0.934 & 0.984 & 10.4 & 10.0 & 9.8 & 10.4 \\
    \bottomrule
    \end{tabular}
  }
\end{table}

\begin{table}[H]
  \centering
  \caption{RL module comparison: Crypto financial metrics (generation). Each metric is averaged over five runs. Best per LLM per dataset is \textbf{bolded}.}
  \label{tab:app_rl_tsg_fin}
  \resizebox{0.85\textwidth}{!}{%
    \begin{tabular}{cl cccc cccc cccc}
    \toprule
    \rowcolor[HTML]{FFF2CC}
    & & \multicolumn{4}{c}{\textbf{$\Delta$VaR $\downarrow$}} & \multicolumn{4}{c}{\textbf{$\Delta$ES $\downarrow$}} & \multicolumn{4}{c}{\textbf{Steps to Best Incumbent $\downarrow$}} \\
    \cmidrule(lr){3-6}\cmidrule(lr){7-10}\cmidrule(lr){11-14}
    \rowcolor[HTML]{FFF2CC}
    \multirow{-2.5}{*}{\textbf{Dataset}} & \multirow{-2.5}{*}{\textbf{Model}} & GPT-4o & GPT-5.4 & Claude & Nova & GPT-4o & GPT-5.4 & Claude & Nova & GPT-4o & GPT-5.4 & Claude & Nova\\
    \midrule
    \rowcolor[HTML]{D9EAD3}
    \cellcolor{white} \multirow{4}{*}{Crypto}
      & IQL (MOSAIC) & 0.058 & 0.061 & \textbf{0.066} & \textbf{0.071} & \textbf{0.094} & \textbf{0.086} & \textbf{0.101} & \textbf{0.083} & \textbf{8.8} & \textbf{8.6} & \textbf{8.2} & \textbf{8.6} \\
      & BC & 0.060 & 0.063 & 0.079 & 0.080 & 0.116 & 0.101 & 0.121 & 0.101 & 9.4 & 9.6 & 9.0 & 8.8 \\
      & DQN & \textbf{0.057} & \textbf{0.059} & 0.075 & 0.083 & 0.126 & 0.097 & 0.116 & 0.096 & 9.2 & 9.2 & 8.8 & 9.2 \\
      & LLM & 0.064 & 0.068 & 0.085 & 0.092 & 0.122 & 0.109 & 0.120 & 0.105 & 10.0 & 10.2 & 9.2 & 9.8 \\
    \bottomrule
    \end{tabular}
  }
\end{table}

Tables~\ref{tab:app_rl_tsf}--~\ref{tab:app_rl_tsg_fin} present the full RL module comparison results. 
IQL-based \textsc{MOSAIC} shows the most consistent overall advantage across forecasting, generation, financial metrics, and refinement efficiency. 
For forecasting, IQL achieves the best or near-best error metrics on most dataset–backbone combinations, while also reaching the best incumbent in fewer refinement steps than BC, DQN, and LLM-only refinement. On Crypto financial metrics, IQL is strongest on most $\Delta$Sharpe and $\Delta$ES results, although DQN or LLM occasionally achieves slightly lower $\Delta$VaR for individual backbones. For generation, IQL again dominates most Marginal, Correlation, Autocorrelation, and Covariance distances on LOB and Stock, and obtains the best $\Delta$ES and fastest refinement on Crypto, while DQN is competitive on some $\Delta$VaR entries. Overall, these results suggest that the IQL policy provides more stable long-horizon refinement decisions than other RL algorithms, or prompt-only LLM refinement, improving both final model quality and the speed of reaching the best candidate.

\subsection{EDA and Multimodal Ablation}
\label{app:eda_ablation}

\begin{table}[H]
  \centering
  \caption{EDA and multimodal ablation: forecasting metrics. Each metric is averaged over five runs. Best per LLM is \textbf{bolded}.}
  \resizebox{\textwidth}{!}{%
    \begin{tabular}{cl cccc cccc cccc cccc}
    \toprule
    \rowcolor[HTML]{FFF2CC}
    & & \multicolumn{4}{c}{\textbf{RMSE $\downarrow$}} & \multicolumn{4}{c}{\textbf{MAE $\downarrow$}} & \multicolumn{4}{c}{\textbf{MAPE (\%) $\downarrow$}} & \multicolumn{4}{c}{\textbf{sMAPE (\%) $\downarrow$}}\\
    \cmidrule(lr){3-6}\cmidrule(lr){7-10}\cmidrule(lr){11-14}\cmidrule(lr){15-18}
    \rowcolor[HTML]{FFF2CC}
    \multirow{-2.5}{*}{\textbf{Dataset}} & \multirow{-2.5}{*}{\textbf{{Variant}}} & GPT-4o & GPT-5.4 & Claude & Nova & GPT-4o & GPT-5.4 & Claude & Nova & GPT-4o & GPT-5.4 & Claude & Nova & GPT-4o & GPT-5.4 & Claude & Nova\\
    \midrule
    \rowcolor[HTML]{D9EAD3}
    \cellcolor{white} \multirow{3}{*}{Crypto}
      & MOSAIC & \textbf{0.207} & \textbf{0.202} & \textbf{0.217} & \textbf{0.214} & \textbf{0.051} & \textbf{0.051} & \textbf{0.054} & \textbf{0.052} & \textbf{1.543} & \textbf{1.531} & \textbf{1.583} & \textbf{1.535} & \textbf{1.547} & \textbf{1.534} & \textbf{1.591} & \textbf{1.540} \\
      & EDA (MOSAIC) & 0.211 & 0.216 & 0.218 & 0.219 & 0.052 & 0.054 & 0.055 & 0.055 & 1.568 & 1.616 & 1.614 & 1.630 & 1.567 & 1.617 & 1.616 & 1.631 \\
      & TS-Agent & 0.224 & 0.220 & 0.218 & 0.219 & 0.056 & 0.055 & \textbf{0.054} & 0.055 & 1.647 & 1.626 & 1.620 & 1.621 & 1.647 & 1.629 & 1.616 & 1.623 \\
    \midrule
    \rowcolor[HTML]{D9EAD3}
    \cellcolor{white} \multirow{3}{*}{LOB}
      & MOSAIC & \textbf{0.149} & \textbf{0.150} & \textbf{0.116} & \textbf{0.161} & \textbf{0.112} & \textbf{0.112} & \textbf{0.083} & 0.122 & \textbf{0.051} & \textbf{0.050} & \textbf{0.037} & 0.055 & \textbf{0.051} & \textbf{0.050} & \textbf{0.037} & 0.055 \\
      & EDA (MOSAIC) & 0.167 & 0.156 & 0.143 & 0.164 & 0.121 & 0.113 & 0.096 & \textbf{0.120} & 0.055 & 0.051 & 0.042 & \textbf{0.054} & 0.055 & 0.051 & 0.042 & \textbf{0.054} \\
      & TS-Agent & 0.154 & 0.158 & 0.143 & 0.174 & 0.116 & 0.115 & 0.101 & 0.130 & 0.053 & 0.052 & 0.045 & 0.058 & 0.053 & 0.052 & 0.045 & 0.058 \\
    \midrule
    \rowcolor[HTML]{D9EAD3}
    \cellcolor{white} \multirow{3}{*}{Stock}
      & MOSAIC & \textbf{7.847} & \textbf{7.815} & \textbf{7.904} & 8.064 & \textbf{4.804} & \textbf{4.804} & \textbf{4.843} & 4.963 & \textbf{1.982} & \textbf{1.981} & \textbf{2.005} & 2.075 & 1.991 & \textbf{1.982} & 1.956 & 2.086 \\
      & EDA (MOSAIC) & 7.892 & 7.958 & 8.047 & \textbf{7.932} & 4.845 & 4.857 & 5.005 & \textbf{4.942} & 1.997 & 2.010 & 2.064 & \textbf{2.053} & 2.007 & 2.016 & 2.073 & 2.057 \\
      & TS-Agent & 8.017 & 8.079 & 7.982 & 8.590 & 4.912 & 4.964 & 4.905 & 5.047 & 2.046 & 2.060 & 2.076 & 2.123 & \textbf{1.770} & 2.062 & \textbf{1.765} & \textbf{1.850} \\
    \bottomrule
    \end{tabular}%
  }
  \label{tab:eda_f_ab}%
\end{table}

\begin{table}[H]
  \centering
  \caption{EDA and multimodal ablation: Crypto financial metrics (forecasting). Each metric is averaged over five runs. Best per LLM is \textbf{bolded}.}
  \resizebox{0.85\textwidth}{!}{%
    \begin{tabular}{cl cccc cccc cccc}
    \toprule
    \rowcolor[HTML]{FFF2CC}
    & & \multicolumn{4}{c}{\textbf{$\Delta$Sharpe $\downarrow$}} & \multicolumn{4}{c}{\textbf{$\Delta$VaR $\downarrow$}} & \multicolumn{4}{c}{\textbf{$\Delta$ES $\downarrow$}} \\
    \cmidrule(lr){3-6}\cmidrule(lr){7-10}\cmidrule(lr){11-14}
    \rowcolor[HTML]{FFF2CC}
    \multirow{-2.5}{*}{\textbf{Dataset}} & \multirow{-2.5}{*}{\textbf{Variant}} & GPT-4o & GPT-5.4 & Claude & Nova & GPT-4o & GPT-5.4 & Claude & Nova & GPT-4o & GPT-5.4 & Claude & Nova \\
    \midrule
    \rowcolor[HTML]{D9EAD3} 
    \cellcolor{white} \multirow{3}[1]{*}{Crypto}
      & MOSAIC & \textbf{6.654} & \textbf{4.329} & 10.687 & 10.211 & \textbf{0.012} & \textbf{0.012} & \textbf{0.015} & \textbf{0.014} & \textbf{0.014} & \textbf{0.015} & \textbf{0.016} & \textbf{0.016} \\
      & EDA (MOSAIC) & 20.039 & 9.401 & \textbf{7.887} & \textbf{7.427} & 0.078 & 0.016 & 0.093 & 0.082 & 0.102 & 0.024 & 0.142 & 0.108 \\
      & TS-Agent & 12.431 & 4.530 & 11.297 & 12.416 & 0.015 & 0.015 & 0.017 & 0.015 & 0.019 & 0.022 & 0.024 & 0.023 \\
    \bottomrule
    \end{tabular}%
  }
  \label{tab:eda_crypto_f_ab}%
\end{table}

\begin{table}[H]
  \centering
  \caption{EDA and multimodal ablation: generation metrics. Each metric is averaged over five runs. Best per LLM is \textbf{bolded}.}
  \resizebox{\textwidth}{!}{%
    \begin{tabular}{clcccccccccccccccc}
    \toprule
    \rowcolor[HTML]{FFF2CC}
    & & \multicolumn{4}{c}{\textbf{Marginal} $\downarrow$} & \multicolumn{4}{c}{\textbf{Correlation} $\downarrow$} & \multicolumn{4}{c}{\textbf{Autocorrelation} $\downarrow$} & \multicolumn{4}{c}{\textbf{Covariance} $\downarrow$} \\
    \cmidrule(lr){3-6}\cmidrule(lr){7-10}\cmidrule(lr){11-14}\cmidrule(lr){15-18}
    \rowcolor[HTML]{FFF2CC}
    \multirow{-2.5}{*}{\textbf{Dataset}} & \multirow{-2.5}{*}{\textbf{{Variant}}} & GPT-4o & GPT-5.4 & Claude & Nova & GPT-4o & GPT-5.4 & Claude & Nova & GPT-4o & GPT-5.4 & Claude & Nova & GPT-4o & GPT-5.4 & Claude & Nova \\
    \midrule
    \rowcolor[HTML]{D9EAD3}
    \cellcolor{white} \multirow{3}{*}{LOB}
      & MOSAIC & \textbf{0.590} & \textbf{0.579} & \textbf{0.554} & \textbf{0.626} & \textbf{0.216} & \textbf{0.182} & \textbf{0.225} & 0.239 & \textbf{0.199} & \textbf{0.247} & \textbf{0.193} & \textbf{0.228} & \textbf{0.181} & \textbf{0.177} & \textbf{0.188} & \textbf{0.192} \\
      & EDA (MOSAIC) & 0.946 & 0.844 & 0.969 & 0.884 & 0.440 & 0.381 & 0.340 & 0.668 & 0.734 & 0.582 & 0.400 & 0.791 & 0.290 & 0.274 & 0.311 & 0.280 \\
      & TS-Agent & 0.713 & 1.180 & 0.767 & 0.822 & 0.243 & 0.472 & 0.235 & \textbf{0.233} & 0.234 & 0.477 & 0.215 & 0.254 & 0.234 & 0.332 & 0.234 & 0.221 \\
    \midrule
    \rowcolor[HTML]{D9EAD3}
    \cellcolor{white} \multirow{3}{*}{Stock}
      & MOSAIC & \textbf{0.408} & \textbf{0.371} & \textbf{0.383} & \textbf{0.385} & \textbf{0.357} & \textbf{0.343} & \textbf{0.345} & \textbf{0.378} & \textbf{0.091} & 0.087 & \textbf{0.081} & 0.081 & \textbf{0.881} & \textbf{0.902} & 0.880 & \textbf{0.903} \\
      & EDA (MOSAIC) & 0.450 & 0.396 & 0.433 & 0.631 & 0.374 & 0.393 & 0.461 & 0.407 & 0.750 & 0.101 & 0.361 & \textbf{0.074} & 0.912 & 0.913 & \textbf{0.850} & 0.938 \\
      & TS-Agent & 0.455 & 0.681 & 0.421 & 0.473 & 0.397 & 0.437 & 0.361 & 0.401 & 0.129 & \textbf{0.075} & 0.193 & 0.859 & 0.910 & 0.939 & 0.909 & 0.930 \\
    \bottomrule
    \end{tabular}%
  }
  \label{tab:eda_g_ab}%
\end{table}

\begin{table}[H]
  \centering
  \caption{EDA and multimodal ablation: Crypto financial metrics (generation). Each metric is averaged over five runs. Best per LLM is \textbf{bolded}.}
  \resizebox{0.85\textwidth}{!}{%
    \begin{tabular}{cl cccc cccc cccc}
    \toprule
    \rowcolor[HTML]{FFF2CC}
    & & \multicolumn{4}{c}{\textbf{$\Delta$VaR $\downarrow$}} & \multicolumn{4}{c}{\textbf{$\Delta$ES $\downarrow$}}\\
    \cmidrule(lr){3-6}\cmidrule(lr){7-10}
    \rowcolor[HTML]{FFF2CC}
    \multirow{-2.5}{*}{\textbf{Dataset}} & \multirow{-2.5}{*}{\textbf{Variant}} & GPT-4o & GPT-5.4 & Claude & Nova & GPT-4o & GPT-5.4 & Claude & Nova \\
    \midrule
    \rowcolor[HTML]{D9EAD3}
    \cellcolor{white} \multirow{3}{*}{Crypto}
      & MOSAIC  & 0.072 & 0.073 & \textbf{0.075} & \textbf{0.078} & \textbf{0.087} & 0.096 & \textbf{0.104} & \textbf{0.095} \\
      & EDA (MOSAIC) & 0.078 & 0.098 & 0.093 & 0.082 & 0.102 & 0.135 & 0.142 & 0.108 \\
      & TS-Agent & \textbf{0.068} & \textbf{0.044} & 0.084 & 0.091 & 0.103 & \textbf{0.054} & 0.111 & 0.104 \\
    \bottomrule
    \end{tabular}%
  }
  \label{tab:eda_crypto_g_ab}%
\end{table}

Tables~\ref{tab:eda_f_ab}--\ref{tab:eda_crypto_g_ab} present the full per-LLM ablation results for the EDA and multimodal module. Removing these components severely degrades generation fidelity; for instance, LOB Marginal distance spikes by up to 75\% on Claude (0.554 $\to$ 0.969). In downstream financial tasks, their absence causes the Crypto $\Delta$Sharpe to deteriorate drastically (e.g., 6.654 $\to$ 20.039 on GPT-4o). For forecasting, semantic EDA consistently outperforms the baseline TS-Agent, improving Stock RMSE across all backbones (e.g., 8.079 $\to$ 7.815 on GPT-5.4). These breakdowns confirm that while advanced backbones can occasionally compensate for missing modalities, explicitly injecting statistical and visual priors is strictly necessary to guarantee robust generation quality and financial safety across varying model scale.

\subsection{Model Generation Ablation}
\label{app:cg_ablation}

\begin{table}[H]
  \centering
  \caption{Model generation ablation: forecasting metrics. Each metric is averaged over five runs. Best per LLM per dataset is \textbf{bolded}.}
  \label{tab:app_cg_ab_tsf}
  \resizebox{\textwidth}{!}{%
    \begin{tabular}{cl cccc cccc cccc cccc cccc}
    \toprule
    \rowcolor[HTML]{FFF2CC}
    & & \multicolumn{4}{c}{\textbf{RMSE $\downarrow$}} & \multicolumn{4}{c}{\textbf{MAE $\downarrow$}} & \multicolumn{4}{c}{\textbf{MAPE (\%) $\downarrow$}} & \multicolumn{4}{c}{\textbf{sMAPE (\%) $\downarrow$}} & \multicolumn{4}{c}{\textbf{Win (\%) $\uparrow$}} \\
    \cmidrule(lr){3-6}\cmidrule(lr){7-10}\cmidrule(lr){11-14}\cmidrule(lr){15-18}\cmidrule(lr){19-22}
    \rowcolor[HTML]{FFF2CC}
    \multirow{-2.5}{*}{\textbf{Dataset}} & \multirow{-2.5}{*}{\textbf{Variant}} & GPT-4o & GPT-5.4 & Claude & Nova & GPT-4o & GPT-5.4 & Claude & Nova & GPT-4o & GPT-5.4 & Claude & Nova & GPT-4o & GPT-5.4 & Claude & Nova & GPT-4o & GPT-5.4 & Claude & Nova\\
    \midrule
    \rowcolor[HTML]{D9EAD3}
    \cellcolor{white} \multirow{5}{*}{Crypto}
      & ModelGen (MOSAIC)              & \textbf{0.205} & \textbf{0.202} & \textbf{0.219} & \textbf{0.205} & \textbf{0.051} & \textbf{0.051} & 0.055 & \textbf{0.051} & \textbf{1.530} & \textbf{1.514} & \textbf{1.615} & \textbf{1.523} & \textbf{1.531} & \textbf{1.519} & 1.621 & \textbf{1.525} & \textbf{60} & \textbf{80} & \textbf{60} & \textbf{60} \\
      & w/o Mod.\ Und.       & 0.222 & 0.216 & 0.223 & 0.220 & 0.056 & 0.055 & 0.055 & 0.054 & 1.655 & 1.624 & 1.681 & 1.623 & 1.622 & 1.644 & 1.656 & 1.636 & 20 & 20 & 20 & 20 \\
      & w/o Blueprint        & 0.228 & 0.222 & 0.221 & 0.238 & 0.056 & 0.056 & \textbf{0.053} & 0.056 & 1.679 & 1.636 & \textbf{1.615} & 1.656 & 1.676 & 1.644 & 1.633 & 1.628 & 20 & 20 & 0 & 0 \\
      & w/o Verify           & 0.223 & 0.220 & 0.226 & 0.224 & 0.056 & 0.055 & 0.056 & 0.055 & 1.662 & 1.645 & 1.718 & 1.631 & 1.676 & 1.670 & 1.646 & 1.651 & 20 & 20 & 20 & 20 \\
      & Naive LLM            & 0.225 & 0.225 & 0.237 & 0.247 & 0.055 & 0.057 & 0.058 & 0.055 & 1.659 & 1.675 & 1.627 & 1.630 & 1.657 & 1.687 & \textbf{1.620} & 1.626 & 0 & 20 & 0 & 0 \\
    \midrule
    \rowcolor[HTML]{D9EAD3}
    \cellcolor{white} \multirow{5}{*}{LOB}
      & ModelGen (MOSAIC)              & \textbf{0.153} & \textbf{0.155} & \textbf{0.095} & \textbf{0.160} & \textbf{0.115} & \textbf{0.115} & \textbf{0.067} & \textbf{0.119} & \textbf{0.052} & \textbf{0.051} & \textbf{0.030} & \textbf{0.054} & \textbf{0.052} & \textbf{0.052} & \textbf{0.030} & \textbf{0.054} & \textbf{60} & \textbf{60} & \textbf{60} & \textbf{60} \\
      & w/o Mod.\ Und.       & 0.159 & 0.162 & 0.123 & 0.172 & 0.120 & 0.117 & 0.086 & 0.129 & 0.055 & 0.054 & 0.039 & 0.058 & 0.054 & 0.053 & 0.038 & 0.058 & 20 & 20 & 20 & 20 \\
      & w/o Blueprint        & 0.156 & 0.166 & 0.145 & 0.172 & 0.117 & 0.119 & 0.100 & 0.132 & 0.053 & 0.055 & 0.046 & 0.058 & \textbf{0.052} & 0.053 & 0.045 & 0.058 & 0 & 20 & 0 & 0 \\
      & w/o Verify           & 0.162 & 0.160 & 0.142 & 0.173 & 0.121 & \textbf{0.115} & 0.101 & 0.129 & 0.055 & 0.052 & 0.045 & 0.060 & 0.054 & \textbf{0.052} & 0.045 & 0.058 & 20 & 0 & 0 & 20 \\
      & Naive LLM            & 0.169 & 0.170 & 0.143 & 0.173 & \textbf{0.115} & 0.120 & 0.102 & 0.130 & 0.053 & 0.056 & 0.045 & 0.058 & 0.053 & 0.054 & 0.045 & 0.059 & 0 & 20 & 0 & 0 \\
    \midrule
    \rowcolor[HTML]{D9EAD3}
    \cellcolor{white} \multirow{5}{*}{Stock}
      & ModelGen (MOSAIC)              & \textbf{7.805} & \textbf{7.769} & \textbf{7.867} & \textbf{7.924} & \textbf{4.774} & \textbf{4.790} & \textbf{4.840} & \textbf{4.910} & \textbf{1.967} & \textbf{1.977} & \textbf{2.017} & \textbf{2.045} & \textbf{1.977} & \textbf{1.979} & \textbf{1.939} & 2.064 & \textbf{80} & \textbf{80} & \textbf{60} & \textbf{60} \\
      & w/o Mod.\ Und.       & 8.219 & 8.220 & 8.266 & 8.555 & 4.995 & 5.113 & 5.070 & 5.217 & 2.105 & 2.077 & 2.137 & 2.158 & 1.981 & 2.111 & 2.113 & 2.121 & 20 & 20 & 20 & 20 \\
      & w/o Blueprint        & 8.374 & 8.372 & 8.369 & 8.622 & 5.126 & 5.186 & 5.193 & 5.025 & 2.116 & 2.143 & 2.173 & 2.128 & 1.992 & 2.165 & 2.121 & 2.123 & 20 & 20 & 20 & 0 \\
      & w/o Verify           & 8.291 & 8.249 & 7.888 & 8.496 & 5.071 & 5.098 & 4.916 & 5.212 & 2.117 & 2.135 & 2.065 & 2.176 & 2.101 & 2.104 & 2.201 & \textbf{2.040} & 20 & 20 & 0 & 20 \\
      & Naive LLM            & 8.511 & 8.089 & 8.010 & 8.508 & 5.309 & 4.882 & 4.905 & 5.034 & 2.197 & 2.062 & 2.053 & 2.133 & 2.120 & 2.084 & 2.213 & 2.203 & 20 & 0 & 0 & 0 \\
    \bottomrule
    \end{tabular}
  }
\end{table}

\begin{table}[H]
  \centering
  \caption{Model generation ablation: Crypto financial metrics (forecasting). Each metric is averaged over five runs. Best per LLM is \textbf{bolded}.}
  \label{tab:app_cg_ab_tsf_fin}
  \resizebox{0.85\textwidth}{!}{%
    \begin{tabular}{cl cccc cccc cccc}
    \toprule
    \rowcolor[HTML]{FFF2CC}
    & & \multicolumn{4}{c}{\textbf{$\Delta$Sharpe $\downarrow$}} & \multicolumn{4}{c}{\textbf{$\Delta$VaR $\downarrow$}} & \multicolumn{4}{c}{\textbf{$\Delta$ES $\downarrow$}} \\
    \cmidrule(lr){3-6}\cmidrule(lr){7-10}\cmidrule(lr){11-14}
    \rowcolor[HTML]{FFF2CC}
    \multirow{-2.5}{*}{\textbf{Dataset}} & \multirow{-2.5}{*}{\textbf{Variant}} & GPT-4o & GPT-5.4 & Claude & Nova & GPT-4o & GPT-5.4 & Claude & Nova & GPT-4o & GPT-5.4 & Claude & Nova \\
    \midrule
    \rowcolor[HTML]{D9EAD3}
    \cellcolor{white} \multirow{5}{*}{Crypto}
      & ModelGen (MOSAIC)              & \textbf{2.074} & \textbf{2.431} & \textbf{10.924} & \textbf{9.477} & \textbf{0.010} & \textbf{0.011} & \textbf{0.014} & \textbf{0.012} & \textbf{0.011} & \textbf{0.013} & \textbf{0.014} & \textbf{0.014} \\
      & w/o Mod.\ Und.       & 7.191 & 3.518 & 11.458 & 11.211 & 0.013 & 0.014 & 0.016 & 0.014 & 0.016 & 0.018 & 0.020 & 0.019 \\
      & w/o Blueprint        & 7.150 & 3.581 & 11.325 & 12.212 & 0.013 & 0.013 & 0.017 & 0.016 & 0.016 & 0.018 & 0.024 & 0.023 \\
      & w/o Verify           & 7.072 & 3.544 & 11.528 & 11.547 & 0.013 & 0.013 & 0.016 & 0.014 & 0.016 & 0.018 & 0.020 & 0.019 \\
      & Naive LLM            & 12.500 & 3.627 & 11.382 & 12.272 & 0.015 & 0.014 & 0.017 & 0.015 & 0.020 & 0.019 & 0.024 & 0.023 \\
    \bottomrule
    \end{tabular}
  }
\end{table}

\begin{table}[H]
  \centering
  \caption{Model generation ablation: generation metrics. Each metric is averaged over five runs. Best per LLM per dataset is \textbf{bolded}.}
  \label{tab:app_cg_ab_tsg}
  \resizebox{\textwidth}{!}{%
    \begin{tabular}{cl cccc cccc cccc cccc cccc}
    \toprule
    \rowcolor[HTML]{FFF2CC}
    & & \multicolumn{4}{c}{\textbf{Marginal $\downarrow$}} & \multicolumn{4}{c}{\textbf{Correlation $\downarrow$}} & \multicolumn{4}{c}{\textbf{AutoCorrelation $\downarrow$}} & \multicolumn{4}{c}{\textbf{Covariance $\downarrow$}} & \multicolumn{4}{c}{\textbf{Win(\%) $\uparrow$}} \\
    \cmidrule(lr){3-6}\cmidrule(lr){7-10}\cmidrule(lr){11-14}\cmidrule(lr){15-18}\cmidrule(lr){19-22}
    \rowcolor[HTML]{FFF2CC}
    \multirow{-2.5}{*}{\textbf{Dataset}} & \multirow{-2.5}{*}{\textbf{Variant}} & GPT-4o & GPT-5.4 & Claude & Nova & GPT-4o & GPT-5.4 & Claude & Nova & GPT-4o & GPT-5.4 & Claude & Nova & GPT-4o & GPT-5.4 & Claude & Nova & GPT-4o & GPT-5.4 & Claude & Nova\\
    \midrule
    \rowcolor[HTML]{D9EAD3}
    \cellcolor{white} \multirow{5}{*}{LOB}
      & ModelGen (MOSAIC)              & \textbf{0.556} & \textbf{0.536} & \textbf{0.513} & \textbf{0.582} & \textbf{0.195} & \textbf{0.168} & \textbf{0.214} & \textbf{0.206} & \textbf{0.194} & \textbf{0.212} & \textbf{0.191} & \textbf{0.233} & \textbf{0.169} & \textbf{0.162} & \textbf{0.187} & \textbf{0.186} & \textbf{60} & \textbf{80} & \textbf{60} & \textbf{60} \\
      & w/o Mod.\ Und.       & 0.655 & 0.874 & 0.770 & 0.715 & 0.224 & 0.321 & 0.237 & 0.224 & 0.220 & 0.351 & 0.215 & 0.256 & 0.208 & 0.252 & 0.232 & 0.214 & 20 & 20 & 0 & 20 \\
      & w/o Blueprint        & 0.718 & 0.880 & 0.772 & 0.824 & 0.246 & 0.323 & 0.234 & 0.232 & 0.234 & 0.353 & 0.212 & 0.256 & 0.232 & 0.252 & 0.233 & 0.223 & 0 & 20 & 0 & 0 \\
      & w/o Verify           & 0.701 & 1.167 & 0.674 & 0.831 & 0.247 & 0.471 & 0.237 & 0.232 & 0.233 & 0.470 & 0.212 & 0.253 & 0.230 & 0.329 & 0.218 & 0.219 & 0 & 0 & 20 & 0 \\
      & Naive LLM            & 0.718 & 0.886 & 0.769 & 0.830 & 0.244 & 0.323 & 0.236 & 0.231 & 0.231 & 0.356 & 0.216 & 0.255 & 0.231 & 0.256 & 0.235 & 0.220 & 0 & 20 & 0 & 0 \\
    \midrule
    \rowcolor[HTML]{D9EAD3} 
    \cellcolor{white} \multirow{5}{*}{Stock}
      & ModelGen (MOSAIC)              & \textbf{0.362} & \textbf{0.326} & \textbf{0.382} & \textbf{0.350} & \textbf{0.364} & \textbf{0.339} & 0.360 & \textbf{0.374} & \textbf{0.090} & 0.091 & \textbf{0.078} & \textbf{0.073} & \textbf{0.894} & \textbf{0.893} & 0.911 & \textbf{0.923} & \textbf{60} & \textbf{80} & \textbf{60} & \textbf{60} \\
      & w/o Mod.\ Und.       & 0.421 & 0.682 & 0.421 & 0.427 & 0.395 & 0.432 & \textbf{0.358} & 0.403 & 0.110 & 0.075 & 0.193 & 0.455 & 0.935 & 0.940 & \textbf{0.904} & 0.958 & 20 & 0 & 0 & 20 \\
      & w/o Blueprint        & 0.432 & 0.678 & 0.422 & 0.477 & 0.407 & 0.442 & 0.359 & 0.396 & 0.114 & \textbf{0.074} & 0.193 & 0.857 & 0.955 & 0.926 & 0.911 & 0.944 & 20 & 0 & 0 & 0 \\
      & w/o Verify           & 0.431 & 0.685 & 0.422 & 0.424 & 0.397 & 0.445 & 0.380 & 0.410 & 0.113 & 0.075 & 0.138 & 0.457 & 0.939 & 0.938 & 0.949 & 0.973 & 20 & 0 & 20 & 20 \\
      & Naive LLM            & 0.459 & 0.681 & 0.427 & 0.470 & 0.392 & 0.430 & 0.388 & 0.399 & 0.127 & \textbf{0.074} & 0.139 & 0.862 & 0.914 & 0.940 & 0.953 & 0.930 & 0 & 0 & 20 & 0 \\
    \bottomrule
    \end{tabular}
  }
\end{table}

\begin{table}[H]
  \centering
  \caption{Model generation ablation: Crypto financial metrics (generation). Each metric is averaged over five runs. Best per LLM is \textbf{bolded}.}
  \label{tab:app_cg_ab_tsg_fin}
  \resizebox{0.85\textwidth}{!}{%
    \begin{tabular}{cl cccc cccc cccc}
    \toprule
    \rowcolor[HTML]{FFF2CC}
    & & \multicolumn{4}{c}{\textbf{$\Delta$VaR $\downarrow$}} & \multicolumn{4}{c}{\textbf{$\Delta$ES $\downarrow$}} & \multicolumn{4}{c}{\textbf{Win (\%) $\uparrow$}} \\
    \cmidrule(lr){3-6}\cmidrule(lr){7-10}\cmidrule(lr){11-14}
    \rowcolor[HTML]{FFF2CC}
    \multirow{-2.5}{*}{\textbf{Dataset}} & \multirow{-2.5}{*}{\textbf{Variant}} & GPT-4o & GPT-5.4 & Claude & Nova & GPT-4o & GPT-5.4 & Claude & Nova & GPT-4o & GPT-5.4 & Claude & Nova \\
    \midrule
    \rowcolor[HTML]{D9EAD3}
    \cellcolor{white} \multirow{5}{*}{Crypto}
      & ModelGen (MOSAIC)              & 0.081 & 0.082 & \textbf{0.081} & \textbf{0.082} & \textbf{0.082} & 0.103 & \textbf{0.105} & \textbf{0.102} & \textbf{60} & \textbf{60} & \textbf{60} & \textbf{60} \\
      & w/o Mod.\ Und.       & 0.073 & 0.054 & 0.085 & 0.090 & 0.096 & 0.069 & 0.113 & 0.108 & 20 & 20 & 20 & 20 \\
      & w/o Blueprint        & 0.073 & \textbf{0.044} & 0.085 & 0.091 & 0.098 & \textbf{0.053} & 0.111 & 0.105 & 20 & 0 & 0 & 0 \\
      & w/o Verify           & 0.074 & \textbf{0.044} & 0.083 & 0.090 & 0.098 & 0.054 & 0.110 & 0.107 & 20 & 0 & 0 & 20 \\
      & Naive LLM            & \textbf{0.068} & \textbf{0.044} & 0.084 & 0.091 & 0.104 & 0.054 & 0.111 & 0.104 & 0 & 0 & 0 & 0 \\
    \bottomrule
    \end{tabular}
  }
\end{table}

Tables~\ref{tab:app_cg_ab_tsf}--\ref{tab:app_cg_ab_tsg_fin} report the full per-LLM ablation results, complementing the averaged analysis in the main text. The patterns hold consistently across all four backbones. Blueprint planning is the most critical component: removing it causes Win Rate to drop to 0\% on LOB and Stock with most backbones, and degrades RMSE by up to 52\% on LOB (Claude: 0.095 $\to$ 0.145). Module understanding contributes the second largest effect, with its removal increasing LOB Marginal distance by 28--63\% depending on backbone. The verify/revise loop primarily affects reliability rather than peak performance: without it, Win Rate drops to 0--20\% on most configurations even when metric values remain competitive, confirming that unverified implementations frequently contain runtime errors. Naive LLM achieves near-zero Win Rate across all backbones and datasets, with the sole exception of GPT-5.4 on LOB (20\%), demonstrating that unstructured generation cannot reliably produce models competitive with repository-grounded synthesise regardless of backbone capability.

\subsection{Reinforcement Learning Ablation}
\label{app:rl_ablation}

\begin{table}[H]
  \centering
  \caption{RL ablation: forecasting metrics. Each metric is averaged over five runs. Best per LLM per dataset is \textbf{bolded}.}
  \label{tab:app_rl_ab_tsf}
  \resizebox{\textwidth}{!}{%
    \begin{tabular}{cl cccc cccc cccc cccc cccc}
    \toprule
    \rowcolor[HTML]{FFF2CC}
    & & \multicolumn{4}{c}{\textbf{RMSE $\downarrow$}} & \multicolumn{4}{c}{\textbf{MAE $\downarrow$}} & \multicolumn{4}{c}{\textbf{MAPE (\%) $\downarrow$}} & \multicolumn{4}{c}{\textbf{sMAPE (\%) $\downarrow$}} & \multicolumn{4}{c}{\textbf{Steps$_\text{F}$$\downarrow$}} \\
    \cmidrule(lr){3-6}\cmidrule(lr){7-10}\cmidrule(lr){11-14}\cmidrule(lr){15-18}\cmidrule(lr){19-22}
    \rowcolor[HTML]{FFF2CC}
    \multirow{-2.5}{*}{\textbf{Dataset}} & \multirow{-2.5}{*}{\textbf{Variant}} & GPT-4o & GPT-5.4 & Claude & Nova & GPT-4o & GPT-5.4 & Claude & Nova & GPT-4o & GPT-5.4 & Claude & Nova & GPT-4o & GPT-5.4 & Claude & Nova & GPT-4o & GPT-5.4 & Claude & Nova \\
    \midrule
    \rowcolor[HTML]{D9EAD3} 
    \cellcolor{white} \multirow{4}{*}{Crypto} & IQL (MOSAIC) & \textbf{0.209} & \textbf{0.203} & \textbf{0.214} & \textbf{0.227} & \textbf{0.053} & \textbf{0.053} & \textbf{0.052} & \textbf{0.053} & \textbf{1.563} & \textbf{1.596} & \textbf{1.536} & \textbf{1.552} & \textbf{1.571} & \textbf{1.596} & \textbf{1.545} & \textbf{1.563} & \textbf{8.80} & \textbf{8.40} & \textbf{8.60} & \textbf{8.60} \\
     & w/o Invalid Action Masking & 0.212 & 0.207 & 0.216 & 0.229 & \textbf{0.053} & 0.054 & 0.053 & 0.054 & 1.590 & 1.609 & 1.559 & 1.579 & 1.593 & 1.609 & 1.564 & 1.586 & 9.20 & 8.80 & 9.00 & 9.00 \\
     & w/o Soft Revert & 0.225 & 0.218 & 0.224 & 0.238 & 0.059 & 0.055 & 0.056 & 0.057 & 1.696 & 1.656 & 1.642 & 1.648 & 1.683 & 1.657 & 1.637 & 1.649 & 10.60 & 10.40 & 10.40 & 10.20 \\
     & w/o Trajectory Branching & 0.219 & 0.213 & 0.221 & 0.233 & 0.056 & 0.055 & 0.054 & 0.055 & 1.644 & 1.632 & 1.603 & 1.615 & 1.636 & 1.634 & 1.599 & 1.621 & 10.00 & 9.60 & 9.80 & 9.60 \\
    \midrule
    \rowcolor[HTML]{D9EAD3} 
    \cellcolor{white} \multirow{4}{*}{LOB} & IQL (MOSAIC) & \textbf{0.144} & \textbf{0.143} & \textbf{0.147} & \textbf{0.162} & \textbf{0.108} & \textbf{0.108} & \textbf{0.107} & \textbf{0.127} & \textbf{0.050} & \textbf{0.049} & \textbf{0.048} & \textbf{0.058} & \textbf{0.050} & \textbf{0.049} & \textbf{0.048} & \textbf{0.058} & \textbf{8.80} & \textbf{8.60} & \textbf{8.40} & \textbf{9.00} \\
     & w/o Invalid Action Masking & 0.146 & 0.144 & 0.148 & 0.165 & 0.110 & \textbf{0.108} & 0.108 & 0.130 & 0.051 & \textbf{0.049} & 0.049 & 0.059 & 0.051 & \textbf{0.049} & 0.049 & 0.059 & 9.20 & 9.00 & 8.80 & 9.40 \\
     & w/o Soft Revert & 0.159 & 0.151 & 0.154 & 0.178 & 0.120 & 0.114 & 0.113 & 0.148 & 0.056 & 0.052 & 0.054 & 0.061 & 0.055 & 0.052 & 0.054 & 0.061 & 10.60 & 10.40 & 10.00 & 10.80 \\
     & w/o Trajectory Branching & 0.152 & 0.147 & 0.151 & 0.172 & 0.115 & 0.111 & 0.111 & 0.140 & 0.053 & 0.051 & 0.052 & 0.060 & 0.053 & 0.051 & 0.051 & 0.060 & 9.80 & 9.80 & 9.40 & 10.00 \\
    \midrule
    \rowcolor[HTML]{D9EAD3} 
    \cellcolor{white} \multirow{4}{*}{Stock} & IQL (MOSAIC) & \textbf{8.013} & \textbf{8.001} & \textbf{7.958} & \textbf{8.275} & \textbf{4.922} & \textbf{4.859} & \textbf{4.847} & \textbf{5.042} & \textbf{2.042} & \textbf{1.997} & \textbf{1.986} & \textbf{2.120} & \textbf{2.049} & \textbf{1.995} & \textbf{1.982} & \textbf{2.118} & \textbf{8.80} & \textbf{8.40} & \textbf{8.20} & \textbf{9.20} \\
     & w/o Invalid Action Masking & 8.058 & 8.047 & 8.013 & 8.325 & 4.959 & 4.904 & 4.885 & 5.074 & 2.057 & 2.019 & 2.009 & 2.154 & 2.065 & 2.024 & 2.012 & 2.129 & 9.20 & 8.80 & 8.60 & 9.60 \\
     & w/o Soft Revert & 8.252 & 8.239 & 8.176 & 8.518 & 5.103 & 5.093 & 5.013 & 5.168 & 2.119 & 2.105 & 2.087 & 2.261 & 2.137 & 2.125 & 2.103 & 2.177 & 10.20 & 10.40 & 10.20 & 10.80 \\
     & w/o Trajectory Branching & 8.165 & 8.152 & 8.087 & 8.438 & 5.036 & 5.013 & 4.944 & 5.121 & 2.087 & 2.067 & 2.052 & 2.211 & 2.106 & 2.083 & 2.067 & 2.152 & 9.60 & 9.80 & 9.40 & 10.20 \\
    \bottomrule
    \end{tabular}
  }
\end{table}

\begin{table}[H]
  \centering
  \caption{RL ablation: Crypto financial metrics (forecasting). Each metric is averaged over five runs. Best per LLM is \textbf{bolded}.}
  \label{tab:app_rl_ab_tsf_fin}
  \resizebox{0.85\textwidth}{!}{%
    \begin{tabular}{cl cccc cccc cccc}
    \toprule
    \rowcolor[HTML]{FFF2CC}
    & & \multicolumn{4}{c}{\textbf{$\Delta$Sharpe $\downarrow$}} & \multicolumn{4}{c}{\textbf{$\Delta$VaR $\downarrow$}} & \multicolumn{4}{c}{\textbf{$\Delta$ES $\downarrow$}} \\
    \cmidrule(lr){3-6}\cmidrule(lr){7-10}\cmidrule(lr){11-14}
    \rowcolor[HTML]{FFF2CC}
    \multirow{-2.5}{*}{\textbf{Dataset}} & \multirow{-2.5}{*}{\textbf{Variant}} & GPT-4o & GPT-5.4 & Claude & Nova & GPT-4o & GPT-5.4 & Claude & Nova & GPT-4o & GPT-5.4 & Claude & Nova \\
    \midrule
    \rowcolor[HTML]{D9EAD3} 
    \cellcolor{white} \multirow{4}{*}{Crypto} & IQL (MOSAIC) & \textbf{13.524} & \textbf{11.924} & \textbf{10.331} & \textbf{11.313} & \textbf{0.015} & \textbf{0.018} & \textbf{0.017} & \textbf{0.016} & \textbf{0.019} & \textbf{0.020} & \textbf{0.019} & \textbf{0.019} \\
     & w/o Invalid Action Masking & 13.947 & 12.319 & 10.695 & 11.756 & 0.016 & \textbf{0.018} & \textbf{0.017} & \textbf{0.016} & 0.020 & 0.021 & 0.020 & 0.021 \\
     & w/o Soft Revert & 15.385 & 14.274 & 12.246 & 12.836 & 0.018 & 0.019 & 0.019 & 0.017 & 0.023 & 0.025 & 0.026 & 0.025 \\
     & w/o Trajectory Branching & 14.722 & 13.187 & 11.483 & 12.319 & 0.017 & 0.019 & 0.018 & 0.017 & 0.021 & 0.022 & 0.024 & 0.023 \\
    \bottomrule
    \end{tabular}
  }
\end{table}

\begin{table}[H]
  \centering
  \caption{RL ablation: generation metrics. Each metric is averaged over five runs. Best per LLM per dataset is \textbf{bolded}.}
  \label{tab:app_rl_ab_tsg}
  \resizebox{\textwidth}{!}{%
    \begin{tabular}{cl cccc cccc cccc cccc cccc}
    \toprule
    \rowcolor[HTML]{FFF2CC}
    & & \multicolumn{4}{c}{\textbf{Marginal $\downarrow$}} & \multicolumn{4}{c}{\textbf{Correlation $\downarrow$}} & \multicolumn{4}{c}{\textbf{AutoCorrelation $\downarrow$}} & \multicolumn{4}{c}{\textbf{Covariance $\downarrow$}} & \multicolumn{4}{c}{\textbf{Steps$_\text{G}$$\downarrow$}} \\
    \cmidrule(lr){3-6}\cmidrule(lr){7-10}\cmidrule(lr){11-14}\cmidrule(lr){15-18}\cmidrule(lr){19-22}
    \rowcolor[HTML]{FFF2CC}
    \multirow{-2.5}{*}{\textbf{Dataset}} & \multirow{-2.5}{*}{\textbf{Variant}} & GPT-4o & GPT-5.4 & Claude & Nova & GPT-4o & GPT-5.4 & Claude & Nova & GPT-4o & GPT-5.4 & Claude & Nova & GPT-4o & GPT-5.4 & Claude & Nova & GPT-4o & GPT-5.4 & Claude & Nova \\
    \midrule
    \rowcolor[HTML]{D9EAD3} 
    \cellcolor{white} \multirow{4}{*}{LOB} & IQL (MOSAIC) & \textbf{0.642} & \textbf{0.748} & \textbf{0.616} & \textbf{0.691} & \textbf{0.250} & \textbf{0.242} & \textbf{0.242} & 0.294 & \textbf{0.202} & 0.386 & \textbf{0.193} & \textbf{0.219} & \textbf{0.201} & \textbf{0.238} & \textbf{0.193} & \textbf{0.206} & \textbf{8.80} & \textbf{9.00} & \textbf{8.60} & \textbf{9.00} \\
     & w/o Invalid Action Masking & 0.675 & 0.779 & 0.651 & 0.732 & 0.260 & 0.253 & 0.255 & \textbf{0.293} & 0.220 & \textbf{0.367} & 0.210 & 0.226 & 0.212 & 0.246 & 0.196 & 0.210 & 9.20 & 9.40 & 9.00 & 9.40 \\
     & w/o Soft Revert & 0.825 & 0.907 & 0.815 & 0.886 & 0.286 & 0.306 & 0.320 & 0.314 & 0.277 & 0.425 & 0.269 & 0.286 & 0.263 & 0.285 & 0.262 & 0.272 & 10.60 & 10.80 & 10.40 & 10.80 \\
     & w/o Trajectory Branching & 0.747 & 0.837 & 0.727 & 0.797 & 0.273 & 0.273 & 0.283 & 0.303 & 0.245 & 0.410 & 0.228 & 0.261 & 0.234 & 0.265 & 0.232 & 0.240 & 9.80 & 10.00 & 9.60 & 10.00 \\
    \midrule
    \rowcolor[HTML]{D9EAD3} 
    \cellcolor{white} \multirow{4}{*}{Stock} & IQL (MOSAIC) & \textbf{0.481} & \textbf{0.551} & \textbf{0.385} & \textbf{0.440} & \textbf{0.350} & \textbf{0.357} & \textbf{0.326} & \textbf{0.387} & \textbf{0.094} & \textbf{0.072} & \textbf{0.088} & \textbf{0.096} & \textbf{0.861} & \textbf{0.941} & \textbf{0.832} & \textbf{0.873} & \textbf{8.80} & \textbf{8.60} & \textbf{8.40} & \textbf{9.00} \\
     & w/o Invalid Action Masking & 0.495 & 0.568 & 0.400 & 0.461 & 0.367 & 0.375 & 0.341 & 0.400 & 0.099 & 0.080 & 0.098 & 0.101 & 0.876 & 0.952 & 0.846 & 0.894 & 9.20 & 9.00 & 8.80 & 9.40 \\
     & w/o Soft Revert & 0.546 & 0.626 & 0.495 & 0.571 & 0.442 & 0.454 & 0.386 & 0.430 & 0.140 & 0.112 & 0.126 & 0.099 & 0.971 & 0.987 & 0.956 & 1.023 & 10.80 & 10.40 & 10.20 & 10.80 \\
     & w/o Trajectory Branching & 0.526 & 0.594 & 0.449 & 0.507 & 0.400 & 0.415 & 0.363 & 0.410 & 0.121 & 0.092 & 0.110 & 0.098 & 0.925 & 0.962 & 0.909 & 0.976 & 9.80 & 9.60 & 9.40 & 10.00 \\
    \bottomrule
    \end{tabular}
  }
\end{table}

\begin{table}[H]
  \centering
  \caption{RL ablation: Crypto financial metrics (generation). Each metric is averaged over five runs. Best per LLM is \textbf{bolded}.}
  \label{tab:app_rl_ab_tsg_fin}
  \resizebox{0.85\textwidth}{!}{%
    \begin{tabular}{cl cccc cccc cccc}
    \toprule
    \rowcolor[HTML]{FFF2CC}
    & & \multicolumn{4}{c}{\textbf{$\Delta$VaR $\downarrow$}} & \multicolumn{4}{c}{\textbf{$\Delta$ES $\downarrow$}} & \multicolumn{4}{c}{\textbf{Steps$_\text{G}$$\downarrow$}} \\
    \cmidrule(lr){3-6}\cmidrule(lr){7-10}\cmidrule(lr){11-14}
    \rowcolor[HTML]{FFF2CC}
    \multirow{-2.5}{*}{\textbf{Dataset}} & \multirow{-2.5}{*}{\textbf{Variant}} & GPT-4o & GPT-5.4 & Claude & Nova & GPT-4o & GPT-5.4 & Claude & Nova & GPT-4o & GPT-5.4 & Claude & Nova \\
    \midrule
    \rowcolor[HTML]{D9EAD3} 
    \cellcolor{white} \multirow{4}{*}{Crypto} & IQL (MOSAIC) & \textbf{0.058} & \textbf{0.061} & \textbf{0.066} & \textbf{0.071} & \textbf{0.094} & \textbf{0.086} & \textbf{0.101} & \textbf{0.083} & \textbf{8.80} & \textbf{8.60} & \textbf{8.20} & \textbf{8.60} \\
     & w/o Invalid Action Masking & 0.061 & 0.064 & 0.070 & 0.076 & 0.101 & 0.091 & 0.106 & 0.089 & 9.20 & 9.00 & 8.80 & 9.00 \\
     & w/o Soft Revert & 0.069 & 0.073 & 0.088 & 0.095 & 0.127 & 0.113 & 0.128 & 0.110 & 10.40 & 10.60 & 10.00 & 10.40 \\
     & w/o Trajectory Branching & 0.066 & 0.069 & 0.080 & 0.087 & 0.116 & 0.102 & 0.116 & 0.099 & 9.80 & 9.80 & 9.40 & 9.60 \\
    \bottomrule
    \end{tabular}
  }
\end{table}

Tables~\ref{tab:app_rl_ab_tsf}--\ref{tab:app_rl_ab_tsg_fin} present the full RL ablation results. 
The full IQL-based \textsc{MOSAIC} variant consistently achieves the best results across nearly all datasets, LLM backbones, and evaluation metrics, confirming the importance of the proposed failure-aware refinement design. Removing invalid action masking leads to mild but consistent degradation, suggesting that preventing repeated or incompatible actions improves refinement stability. In contrast, removing soft revert causes the largest performance drop and requires more steps to reach the best incumbent, indicating that retaining executable but degrading edits can push the refinement trajectory into poorer program states. Removing trajectory branching also harms performance, especially on generation metrics, showing that preserving failed or reverted segments as negative supervision is useful for offline policy learning. Overall, the ablations demonstrate that invalid-action masking, soft rollback, and trajectory branching are complementary components for stable long-horizon refinement.

\subsection{Case Study}
\label{app:case_study}

\paragraph{Model Generation Examples.}
We present two representative examples from the model generation module, illustrating the two composition modes: component-level recombination within a shared architectural family, and idea-level transfer across families.

 \begin{figure}[h]
  \centering
  \includegraphics[width=\textwidth]{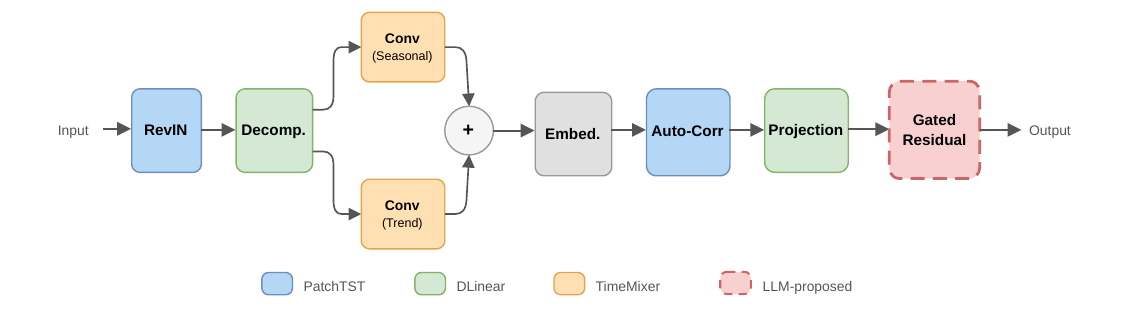} 
  \vspace{-0.5em}
  \caption{Generated model architecture for hourly cryptocurrency forecasting by GPT‑4o.}%
  \label{fig:crypto_arch}%
  \vspace{-1.0em}
\end{figure}

\CaseHeading{\ding{182}}{Case 1: Forecasting on Crypto.} 

Using the LLM GPT‑4o, we generate task-specific architectures for hourly cryptocurrency price forecasting (Figure~\ref{fig:crypto_arch}). The top-$k$ shortlist contains PatchTST, DLinear, and TimeMixer, all sharing a decomposition-and-projection paradigm. The family classifier selects direct component recombination. The generated architecture combines DLinear's series decomposition with TimeMixer's multi-scale convolution branches and PatchTST's auto-correlation attention, connected through a temporal embedding layer. The blueprint stage adds a gated residual head, automatically designed by the LLM to suit the task, blending the network output with the most recent observation via a learned gate. During the execute-verify-refine loop, a shape mismatch at the projection head is detected and automatically corrected by inserting separate temporal and channel linear layers. The resulting hybrid achieves RMSE 0.205 versus 0.217 for the best single-model baseline (PatchTST), with Sharpe Ratio Difference reduced from 12.4 to 2.07.

 \begin{figure}[H]
  \centering
  \includegraphics[width=\textwidth]{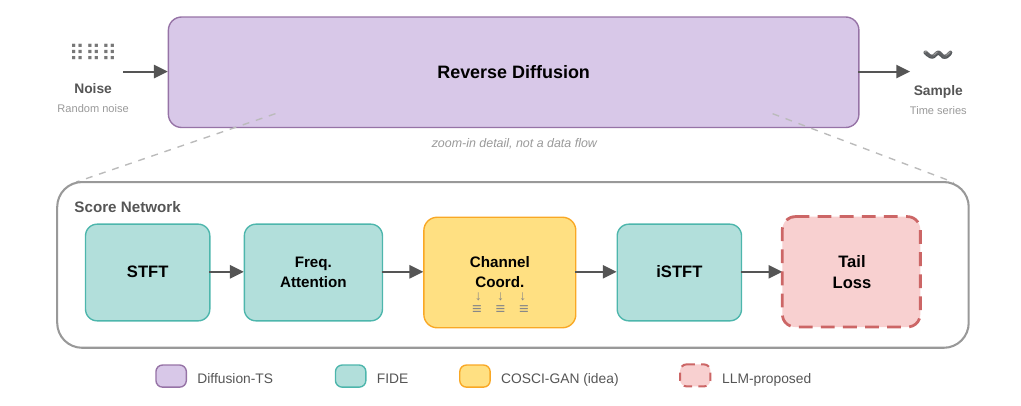} 
  \vspace{-0.5em}
  \caption{Generated model architecture for high-frequency LOB generation by Claude Opus 4.}%
  \label{fig:lob_arch}%
  \vspace{-1.0em}
\end{figure}

\CaseHeading{\ding{183}}{Case 2: Generation on LOB.} 

For unconditional generation of high-frequency order-flow sequences, the LLM Claude Opus 4 produces a blueprint shown in Figure~\ref{fig:lob_arch}. The shortlist contains Diffusion-TS, FIDE, and COSCI-GAN, spanning diffusion-based and GAN-based families. The family classifier designates Diffusion-TS as the primary backbone. FIDE, being in the same family, contributes its frequency-inflated score block as a direct replacement for the original time-domain encoder. COSCI-GAN contributes at the idea level only: its per-channel coordination principle is reinterpreted as a channel-factorised conditioning module inside the diffusion score network, avoiding the instability that would arise from directly grafting adversarial components into a score-matching objective. The blueprint also introduces a tail-aware loss weighting targeting the heavy-tailed distributions flagged by EDA. Guided by insights from the retrieved prior cases, the model blueprint incorporates per-channel residual connections to mitigate mode collapse observed in the initial sample visualisation. The final model achieves Marginal distance 0.513 versus 0.767 for Diffusion-TS alone, and Autocorrelation distance 0.191 versus 0.215.

These two cases demonstrate that the module's blueprint-constrained composition, whether through direct component substitution or cross-family idea transfer, produces novel architectures that outperform their constituent candidates while remaining executable and stable.

\paragraph{Generated Task Examples.}
\label{app:generated_task_examples}
The following cases show how a human-written instruction is transformed into a structured task specification and then into a final task description generated by the pipeline.


\CaseHeading{\ding{182}}{Case 1: Foreign-exchange generation.}
This case illustrates an unsupervised generation task for developed-market foreign-exchange returns, where generated windows are evaluated by distributional, temporal, and cross-asset dependence metrics.
\par\nobreak

\Needspace{12\baselineskip}
\begin{casebox}{Case 1(a): Instruction}
\begin{casecontent}

Create five executable financial time-series generation tasks using the foreign-exchange dataset only.

\CaseField{Task type}
Each task must be generative and unsupervised. A usable sample is a full fixed-length window. There is no forecasting horizon, label, supervised input/output pair, or internal train-target split within a window.

\CaseField{Data constraints}
The data source must be foreign exchange. Each task must select an explicit, meaningful, non-empty subset of assets. The task should not use all available assets. The preferred number of target assets is between 1 and 12.

\CaseField{Statistical requirements}
Each task should generate realistic multi-asset foreign-exchange trajectories and should emphasize heavy tails, volatility clustering, temporal dependence, and cross-asset dependence.

\CaseField{Diversity requirements}
Since the source dataset is fixed, tasks should vary in window length, stride, asset subset, liquidity profile, and time range. The generated set should include narrow, moderate, and broad asset configurations.

\CaseField{Feasibility requirements}
The date range and window parameters should ensure at least 100 validation windows and 100 test windows.

\CaseField{Evaluation requirements}
If a shared generative evaluation package is provided, use it as the default evaluation path. Otherwise, select appropriate metrics from the evaluation-template library. The evaluation should include correlation, marginal distribution, autocorrelation, and covariance scores.

\end{casecontent}
\end{casebox}

\Needspace{12\baselineskip}
\begin{casebox}{Case 1(b): Generated Task Specification}
\begin{casecontent}

\CaseField{Task brief}
Generate 60-day windows of developed-market foreign-exchange pairs to capture stable developed-market dynamics.

\CaseField{Task type}
Generation

\CaseField{Dataset plan}
\begin{casecode}
dataset_id = foreign_exchange,
selection_locked = True.
\end{casecode}

\CaseField{Parameter plan}
\begin{casecode}
window_size = 60,
log_return = True,
stride = 5,
start_time = 2005-01-03,
end_time = 2026-03-30,
target_assets = {eur_usd, gbp_usd, usd_jpy, usd_chf, usd_cad, nzd_usd},
train/val/test = (0.70, 0.15, 0.15).
\end{casecode}

\CaseField{Evaluation plan}
\begin{casecode}
mode = library_metrics,
focus = realism and dependence structure.
\end{casecode}

The evaluation protocol includes four library metrics:

\begin{casecode}
correlation_score: L2 difference between cross-feature correlation matrices.
marginal_distribution_score: L2 distance between per-feature flattened marginal distributions.
autocorrelation_score: L2 difference of per-feature autocorrelation up to the maximum lag.
covariance_score: L2 difference between cross-feature covariance matrices.
\end{casecode}

\CaseField{Constraints}
The task must remain generative, with no forecasting horizon or supervised input/output split. The source dataset must remain foreign exchange. The target assets must be an explicit non-empty subset and must not include all available assets. The task must contain sufficient train, validation, and test windows after chronological splitting.

\CaseField{Executability rationale}
The period from 2005-01-03 to 2026-03-30 provides over 4,000 daily observations. With a 60-day window and stride 5, this yields over 700 rolling windows, sufficient for the required train, validation, and test splits. Restricting the task to six developed-market foreign-exchange pairs gives a moderate-dimensional system with meaningful cross-asset dependence and manageable generative-modelling complexity.

\end{casecontent}
\end{casebox}

\Needspace{12\baselineskip}
\begin{casebox}{Case 1(c): Final Task Description}
\begin{casecontent}

\CaseField{Task overview}
This task involves generative modelling of 60-day multivariate sequences of developed-market currency pairs. The objective is to synthesize realistic financial time-series windows that preserve cross-asset dependence and marginal distributional properties of the underlying foreign-exchange data.

\CaseField{Problem statement}
The model must generate synthetic 60-day log-return windows for six currency pairs:

\begin{casecode}
{eur_usd, gbp_usd, usd_jpy, usd_chf, usd_cad, nzd_usd}
\end{casecode}

The task is strictly generative and does not involve forecasting, prediction horizons, labels, or supervised input--target decomposition. The solver must learn the joint distribution of these six return series and generate sequences matching historical statistical properties. The realised dataset contains 760 training samples, 163 validation samples, and 164 test samples, each with a fixed window length of 60 days.

\CaseField{Dataset}
The dataset is derived from the foreign-exchange source and processed into a clean, synchronised 16-dimensional panel. The data span 2005-01-03 to 2026-03-30. Close prices are transformed into log-returns, after which rolling windows are constructed with window size 60 and stride 5. The realised tensor shapes are:

\begin{casecode}
train = [760, 60, 6],
val = [163, 60, 6],
test = [164, 60, 6].
\end{casecode}

The resulting artifact is stored as a processed dataset package.

\CaseField{Evaluation}
The generated samples are evaluated using:

\begin{casecode}
{correlation_score, marginal_distribution_score, autocorrelation_score, covariance_score, avg_sharpe_diff, avg_var_diff, avg_es_diff}
\end{casecode}

\CaseField{Quick reference}
\begin{casecode}
task_type = generation,
dataset = foreign_exchange,
window_size = 60,
stride = 5,
assets = {eur_usd, gbp_usd, usd_jpy, usd_chf, usd_cad, nzd_usd}.
\end{casecode}

\end{casecontent}
\end{casebox}


\CaseHeading{\ding{183}}{Case 2: Equity log-return forecasting.}
This case illustrates a supervised multi-asset forecasting task for retail and leisure equities, where a historical input window is used to predict a future return horizon for the same target assets.
\par\nobreak

\Needspace{12\baselineskip}
\begin{casebox}{Case 2(a): Instruction}
\begin{casecontent}

Create multiple, distinct executable, sensible, implementable equity forecasting tasks using the equity source only.

\CaseField{Task requirements}
Each task type must be forecasting. Forecasting here means a supervised time-series forecasting task. Each usable sample has a clear input/output structure: a historical input window over target assets is used to predict a future target horizon for those same target assets. Keep forecasting semantics explicit throughout; do not describe the task as full-window generation or as an unsupervised modelling problem.

Each data source must be equity time-series and should remain a standard source throughout planning. If a dataset is pre-selected externally for the run, every child task must use the same dataset and diversify through parameter choices rather than dataset choice. Multiple assets should be used as input context to capture cross-asset dependence in a volatile market setting.

Target assets must be an explicit, meaningful, manually chosen, non-empty list. Use the same target asset list for both forecasting input context and forecast outputs. Use at least 6 target assets, preferably 7--25 when supported by the dataset. The forecasting target should be operationally meaningful and reflect changing correlation structure. Asset selection should account for asset-to-asset correlation and redundancy.

\CaseField{Diversity requirements}
Since the data source is fixed as equity time-series, tasks should diversify in forecasting horizon, stride, asset selection, time range, input window, and regime framing while preserving empirical and operational meaning. The date range should not be too narrow, and validation and test splits should contain sufficient samples.

\CaseField{Evaluation requirements}
A fixed forecasting evaluation package will be provided. The task description must state exactly that metrics include \texttt{mae}, \texttt{mse}, \texttt{rmse}, \texttt{mape}, \texttt{smape}, \texttt{sharpe ratio}, \texttt{VaR}, and \texttt{ES}. Do not invent other information.

\CaseField{Implementation expectations}
The task should read like a sensible, implementable equity forecasting task. Avoid synthetic replacements or generic dataset choices. Keep all instructions precise, implementable, and suitable for automatic task generation.

\end{casecontent}
\end{casebox}

\Needspace{12\baselineskip}
\begin{casebox}{Case 2(b): Generated Task Specification}
\begin{casecontent}

\CaseField{Task brief}
Forecast 10-day returns for a diversified basket of retail and leisure stocks.

\CaseField{Task type}
Forecasting

\CaseField{Dataset plan}
\begin{casecode}
dataset_id = equity,
selection_locked = True.
\end{casecode}

\CaseField{Parameter plan}
\begin{casecode}
train/val/test = (0.70, 0.15, 0.15),
log_return = True,
horizon = 10,
stride = 2,
input_window = 50,
target_assets = {low, sbux, tjx, hd, nike, dis},
forecast_shuffle_train_after_split = True.
\end{casecode}

\CaseField{Evaluation plan}
\begin{casecode}
mode = library_metrics,
focus = consumer sentiment tracking.
\end{casecode}

The evaluation protocol uses standard forecasting error and risk metrics:

\begin{casecode}
metrics = {mae, mse, rmse, mape, smape, sharpe ratio, VaR, ES}.
\end{casecode}

\CaseField{Constraints}
The task must use \texttt{standard\_equity} only. The target asset list must contain at least 6 assets. The same target assets must be used for both the historical input window and the forecasting output horizon. The task must remain supervised forecasting, with explicit input/output structure.

\CaseField{Executability rationale}
Retail and leisure stocks are sensitive to consumer spending and seasonal demand. A 50-day input window captures short-term market and sector trends, while a 10-day horizon defines a meaningful forecasting target. The selected 6-asset basket supports cross-asset dependence modelling while remaining computationally manageable.

\end{casecontent}
\end{casebox}

\Needspace{12\baselineskip}
\begin{casebox}{Case 2(c): Final Task Description}
\begin{casecontent}

\CaseField{Task overview}
This task involves multi-asset forecasting of 10-day log returns for a basket of retail and leisure equities. The objective is to capture consumer sentiment trends and sector-specific volatility in retail-sensitive markets.

\CaseField{Problem statement}
The model must predict the 10-day log-return horizon for 6 retail and leisure assets:

\begin{casecode}
{low, sbux, tjx, hd, nike, dis}
\end{casecode}

Each sample contains a 50-day historical input window and a subsequent 10-day forecasting target for the same assets. The solver operates on 2,073 training samples, 446 validation samples, and 447 test samples. The model must account for the 2-day stride used in window construction. This task is meaningful for tracking consumer spending sensitivity and seasonal retail trends.

\CaseField{Dataset}
The dataset is derived from the \texttt{equity} source, covering daily close prices from 2002-05-23 to 2026-04-08. Prices are transformed into log-returns and split chronologically using a 70/15/15 ratio before windowing. The realised tensor shapes are:

\begin{casecode}
train_X = [2073, 50, 6], train_Y = [2073, 10, 6],
val_X = [446, 50, 6], val_Y = [446, 10, 6],
test_X = [447, 50, 6], test_Y = [447, 10, 6].
\end{casecode}

The dataset is provided as a processed archive containing complete multivariate observations for the selected assets.

\CaseField{Evaluation}
The forecasts are evaluated using:

\begin{casecode}
{mae, mse, rmse, mape, smape, sharpe ratio, VaR, ES}
\end{casecode}

\CaseField{Quick reference}
\begin{casecode}
task_type = forecasting,
dataset = equity,
input_window = 50,
horizon = 10,
stride = 2,
assets = {low, sbux, tjx, hd, nike, dis},
metrics = {mae, mse, rmse, mape, smape, sharpe ratio, VaR, ES}.
\end{casecode}

\end{casecontent}
\end{casebox}

\end{document}